\documentclass{article}
\usepackage[accsupp]{axessibility}  % Improves PDF readability for those with disabilities.

\usepackage{graphicx}
\usepackage{booktabs}
\graphicspath{ {./figs/} }
\usepackage{subcaption}
\usepackage{xspace}
\usepackage[dvipsnames]{xcolor}
\usepackage{comment}
\usepackage{adjustbox}
\usepackage{array}
\usepackage{makecell}
\usepackage{pifont}
\usepackage{bbm}
\usepackage{bm}
\usepackage{amsbsy}
\usepackage{stmaryrd}
\usepackage{wasysym}
\usepackage{multirow}
\usepackage{flushend}
\usepackage{colortbl}
\usepackage{enumitem}
\usepackage{comment}
\usepackage{wrapfig}
\usepackage{sidecap}
\newcolumntype{R}[2]{%
    >{\adjustbox{angle=#1,lap=1.3\width-(#2)}\bgroup}%
    l%
    <{\egroup}%
}
\usepackage[]{pgfplots,pgfplotstable}
\usepackage[]{tikz}
\pgfplotsset{compat=1.18}
\usepackage{soul}
\setuldepth{x}
\usepackage[normalem]{ulem}
\usepackage{setspace}
\makeatletter
\newcommand\myparagraph{\@startsection{paragraph}{4}{\z@}%
    {-6\p@ \@plus -3\p@ \@minus -3\p@}%
    {-0.5em \@plus -0.22em \@minus -0.1em}%
    {\normalfont\normalsize\itshape}}
\newcommand\cvprparagraph{\@startsection{paragraph}{4}{\z@}%
    {0.25ex \@plus 1ex \@minus .2ex}%
    {-1em}%
    {\normalfont \normalsize \bfseries}}
\makeatother
\newcommand\simpletextparagraph[1]{\textbf{#1}\ }

\newcommand{\ra}[1]{\renewcommand{\arraystretch}{#1}}

\definecolor{myblue}{rgb}{0.19, 0.55, 0.91}
\definecolor{myred}{rgb}{0.82, 0.1, 0.26}
\definecolor{MyGreen}{RGB}{0, 154, 55} %{0, 180, 0}
\definecolor{MyRed}{RGB}{248, 3, 7} %{180, 0, 0}
\definecolor{MyYellow}{RGB}{160, 160, 0}
\definecolor{MyBrown}{RGB}{191, 135, 93}
\newcommand{\cmark}{{\textcolor{MyGreen}{\ding{51}}}}%
\newcommand{\xmark}{{\textcolor{MyRed}{\ding{55}}}}%
\newcommand{\mmark}{{\textcolor{MyYellow}{(\ding{51})}}}%
\newcommand{\mmmark}{{\textcolor{MyBrown}{($\boldsymbol\sim$)}}}%
\renewcommand{\mmmark}{{\textcolor{MyBrown}{$\pmb\pm$}}}%

\newcommand\nocodeavail{{\textcolor{MyRed}{$\Circle$}}}
\newcommand\codeavailnoSK{{\textcolor{MyYellow}{($\CIRCLE$)}}}
\newcommand\codeavailissues{{\textcolor{MyBrown}{$\LEFTcircle$}}}
\newcommand\codeavailissuesnoSK{{\textcolor{MyBrown}{($\LEFTcircle$)}}}
\newcommand\codeavail{{\textcolor{MyGreen}{$\CIRCLE$}}}

\newcommand{\nermin}[1]{\textcolor{black}{#1}}

\newcommand{\bRM}[1]{\bgroup\color{black}{#1}}
\newcommand{\eRM}[1]{\egroup}
\newcommand{\renaud}[1]{{\color{black}#1}}

\def\sk{SemanticKITTI\xspace} 
\def\ns{nuScenes\xspace}

\def\Ours{IGLOSS} % Image Generation for Lidar Open-vocabulary Semantic Segmentation
\def\ours{\Ours\xspace}
\def\Oursclosedens{\Ours\textsubscript{mix,clo}}
\def\Oursclosed{\Ours\textsubscript{clo}}
\def\oursclosed{\Oursclosed\xspace}
\def\Oursens{\Ours\textsubscript{mix}}
\def\oursens{\Oursens\xspace}
\def\oursclosedens{\Oursclosedens\xspace}
\def\Scalrplus{ScaLR\textsuperscript{\smash+}\!}% +}%
\def\Scalrplus{ScaLR\raisebox{0.9ex}{\scriptsize+\!}}% +}%
\def\scalrplus{\Scalrplus\xspace}
\newcommand\conc{{$\varoplus$}}

\newcommand\na{\mbox{\scriptsize\color{gray} N/A}}

\usepackage{pgfplots}
\usepackage[nomessages]{fp}
\usepackage{pgfplots}
\usepackage{pgfkeys}
\def\addlegendimage{\csname pgfplots@addlegendimage\endcsname}
\newcommand\img{\text{\textcolor{blue}{\textbf{i}}}}
\newcommand\txt{\text{\textcolor{magenta}{\textbf{t}}}}
\newcommand\msk{\text{\textcolor{violet}{\textbf{m}}}}
\newcommand\bx{\text{\textcolor{cyan}{\textbf{b}}}}
\newcommand\feat{\text{\textcolor{black}{$f$}}}
\newcommand\pnt{\text{\textcolor{ForestGreen}{\textbf{p}}}}
\newcommand\ift{$\img\,{\Rightarrow}\,\feat\,{\Leftarrow}\,\txt$}
\newcommand\ifp{$\img\,{\Rightarrow}\,\feat\,{\Leftarrow}\,\pnt$}
\newcommand\tif{$\txt\,{\Rightarrow}\,\img\,{\Rightarrow}\,\feat$}
\newcommand\tgf{$\txt\,{\Rightarrow}\,\feat$}
\newcommand\igf{$\img\,{\Rightarrow}\,\feat$}
\newcommand\igt{$\img\,{\Rightarrow}\,\txt$}
\newcommand\pgf{$\pnt\,{\Rightarrow}\,\feat$}
\newcommand\tgi{$\txt\,{\Rightarrow}\,\img$}
\newcommand\igm{$\img\,{\Rightarrow}\,\msk$}
\newcommand\itgb{$(\img,\txt)\,{\Rightarrow}\,\bx$}
\newcommand\itgm{$(\img,\txt)\,{\Rightarrow}\,\msk$}

\newcommand{\quot}[1]{``\,#1\,''}

\newcommand\lrangle[1]{$\langle$#1$\rangle$}

\DeclareMathOperator{\argmax}{arg\,max}
\DeclareMathOperator{\LR}{LR}
\DeclareMathOperator{\NN}{NN}

\newcounter{benef}
\newcommand\strength[1]{#1}
\newcommand\intropt[1]{\emph{advantage \##1}}
\newcommand\intropts[2]{\emph{advantages \##1 and \##2}}

\def\App{App.}
\newcommand{\appref}[1]{\App~\ref{#1}}
\PassOptionsToPackage{numbers, sort&compress}{natbib}
 \usepackage[main, final]{neurips_2026}
\usepackage{hyperref}
\usepackage[capitalize,noabbrev]{cleveref}

\title{\ours: Image Generation for Lidar Open-vocabulary Semantic Segmentation}

\author{\hspace{-3mm}Nermin Samet$^1$,
Gilles Puy$^1$,
Renaud Marlet$^{1,2}$ \\[2mm]
\hspace{-5mm}\textsuperscript{1}Valeo.ai, Paris, France \\
\hspace{2.5mm}\textsuperscript{2}LIGM, Ecole des Ponts, Univ Gustave Eiffel, CNRS, Marne-la-Vall\'ee, France \\
}

\begin{document}

\maketitle

\begin{abstract}
This paper presents a new method for the zero-shot open-vocabulary semantic segmentation (OVSS) of 3D automotive lidar data. To circumvent the recognized image-text modality gap that is intrinsic to approaches based on Vision Language Models (VLMs) such as CLIP, our method relies instead on image generation from text, to create prototype images. Given a 3D network distilled from a 2D Vision Foundation Model (VFM), we then label a point cloud by matching 3D point features with 2D image features of these prototypes. Our method is state-of-the-art for OVSS on nuScenes and SemanticKITTI. Code, pre-trained models and generated images are available at \url{https://github.com/valeoai/IGLOSS}.
\end{abstract}
\section{Introduction}
\label{sec:intro}

% This paper presents a new approach for the \emph{zero-shot open-vocabulary semantic segmentation (OVSS) of 3D automotive lidar data}:
We introduce a new framework for \emph{zero-shot open-vocabulary semantic segmentation (OVSS) of 3D automotive LiDAR data}:
given free text prompts defining classes of interest, our method labels scan points with one of these classes.
Being \emph{zero-shot}, thus not requiring any labels nor any sample, brings efficiency in a driving setting where manually labeling 3D data is long and costly~\cite{semantickitti}. 
Besides, while autonomous driving can simply rely on a closed set of classes, being \emph{open-vocabulary}, not specific to fixed classes, brings flexibility. It effortlessly offers any segmentation specialization depending, e.g., on a target world region or on a specific driving scenario. Such an OVSS method is not intended to be embedded in a vehicle but to be used, e.g., for auto-labeling or retrieval.

 % \TODO{NN (or LR) missing: for feature comparison and label assignment, the 3D pretrained backbone should be frozen.}
OVSS has developed \cite{zhou2022maskclip, ding2023maskclip, dong2023maskclip} with % the rise of 
\emph{Vision Language Models (VLMs)}, %. These models are 
built on the image modality \cite{radford2021clip, jia2021align, zhai2023siglip} thanks to the availability of massive 2D data aligned with text \cite{cha2023tcl}. 
In contrast, there is much less 3D data available, even less for driving, and even less with associated text.

% --- Image cell size (edit these) ---
% \newlength{\cellw}
% \newlength{\cellh}

% --- Image cell size (edit these) ---
\newlength{\cellw}
\newlength{\cellh}
\setlength{\cellw}{1.4cm} % width of each image box
\setlength{\cellh}{0.7cm} % height of each image box

% Helper macro: fits image into a fixed box without distortion
\newcommand{\cellimg}[1]{%
\adjincludegraphics[width=\cellw,height=\cellh]{#1}%
}

\newcommand{\tinyafterresize}[1]{\scalebox{0.35}{\strut #1}} 

\begin{figure}
\setlength{\tabcolsep}{0.5pt}
\renewcommand{\arraystretch}{0.5}
\resizebox{0.9\linewidth}{!}{
  \centering
  \begin{tabular}{*{4}{>{\centering\arraybackslash}m{\cellw}}}
    % Row 1
     \cellimg{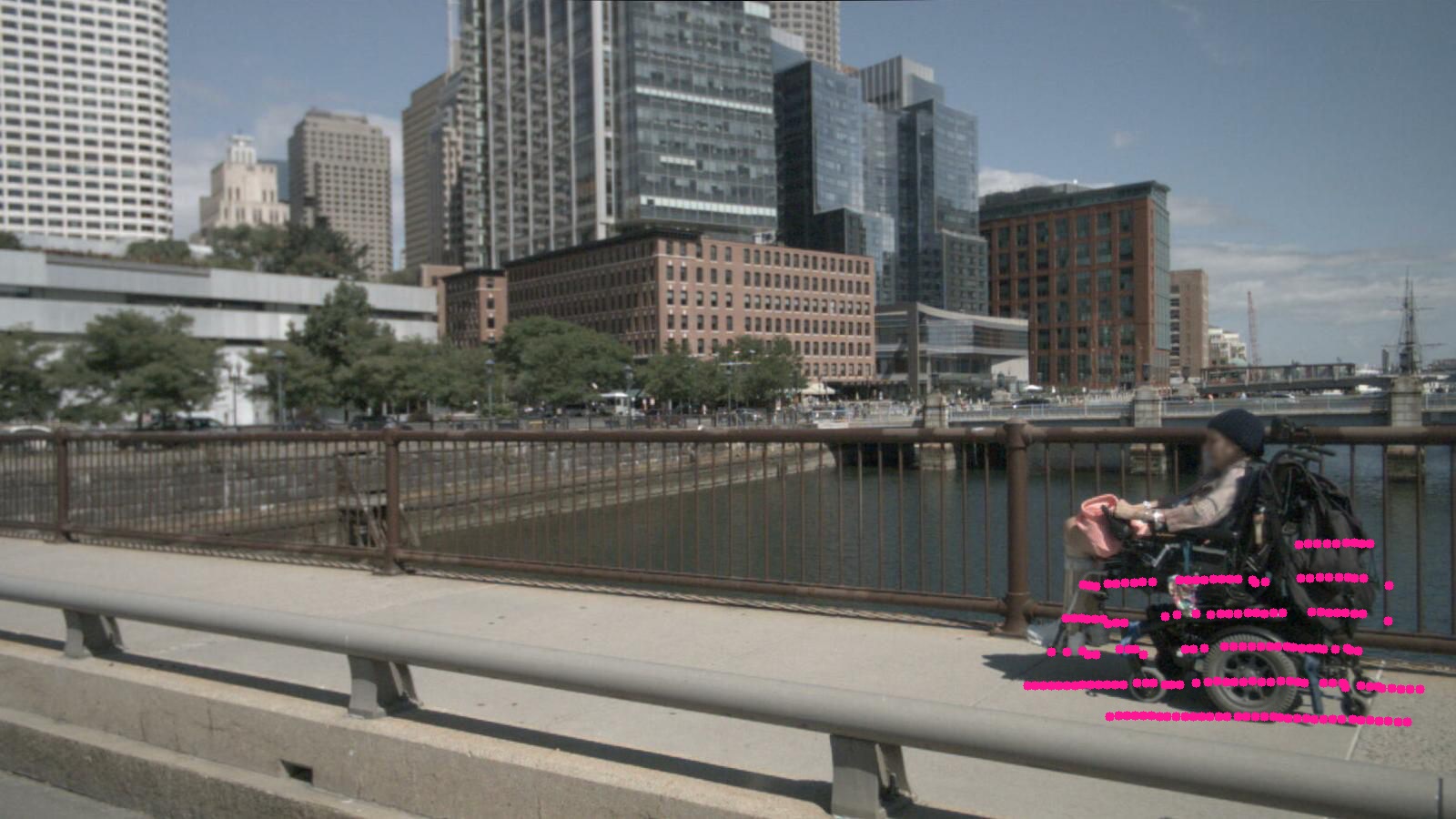} 
     & \cellimg{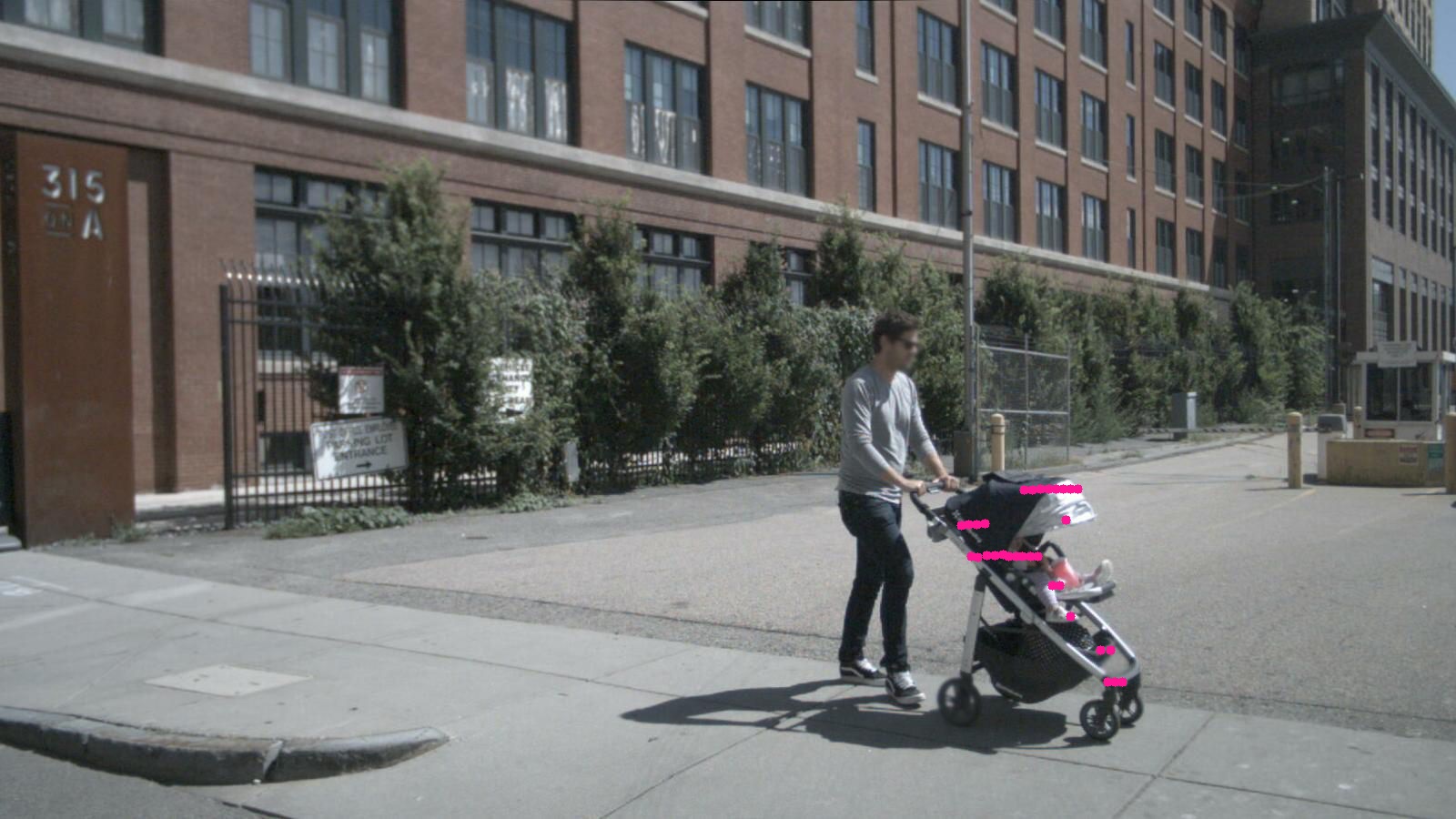} 
     &  \cellimg{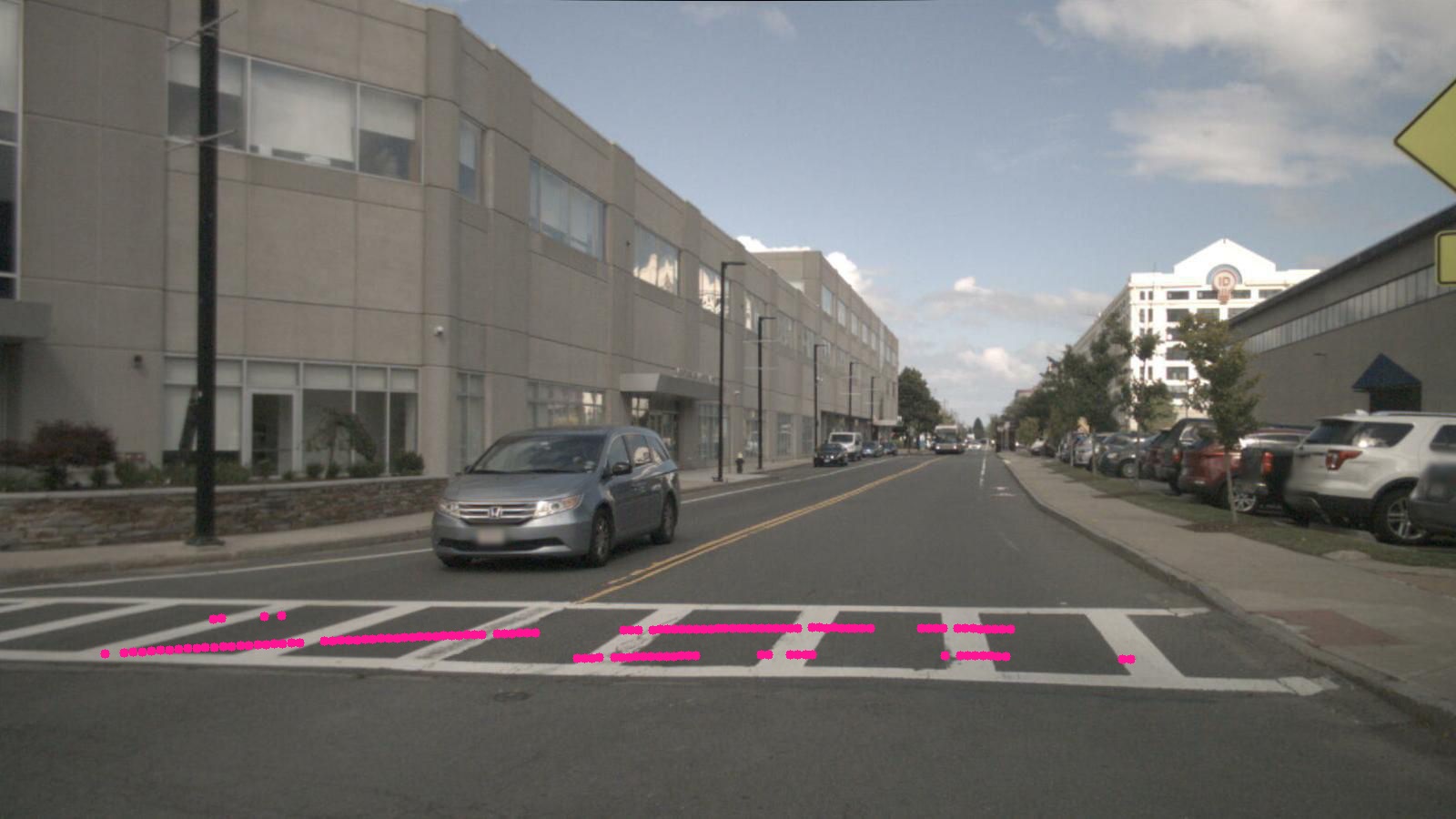} \\      
     
    \cellimg{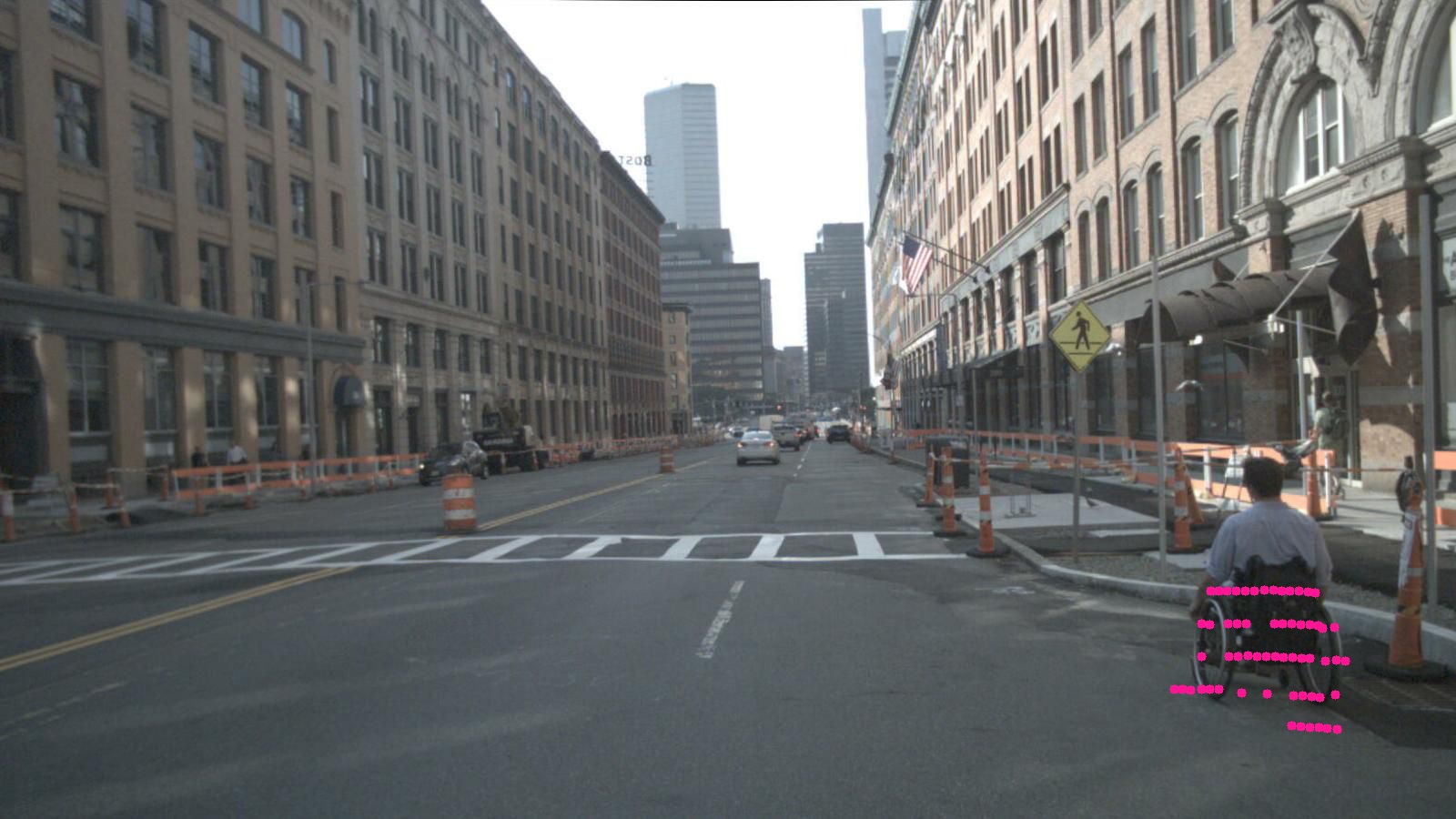} 
    & \cellimg{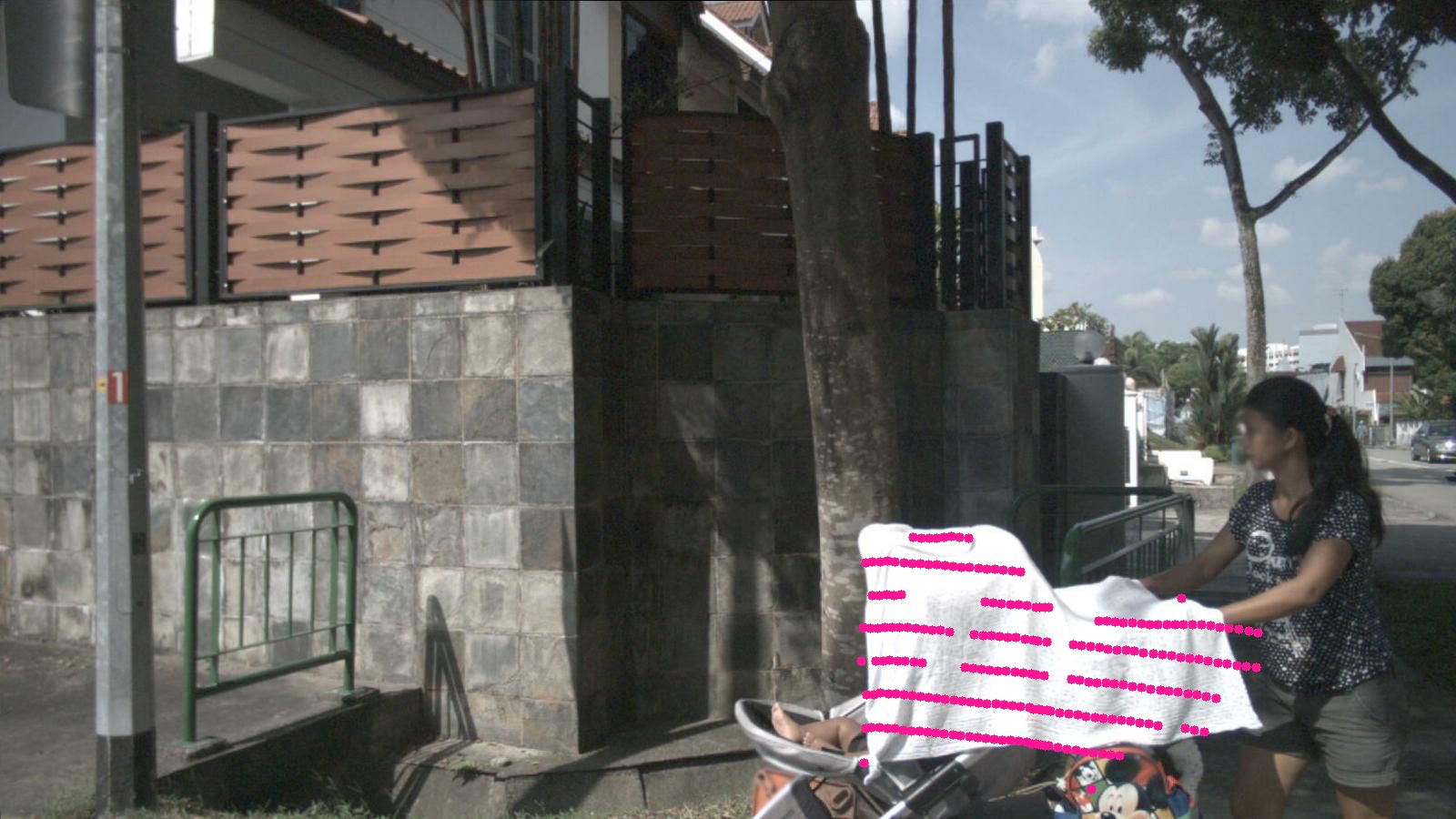} 
    & \cellimg{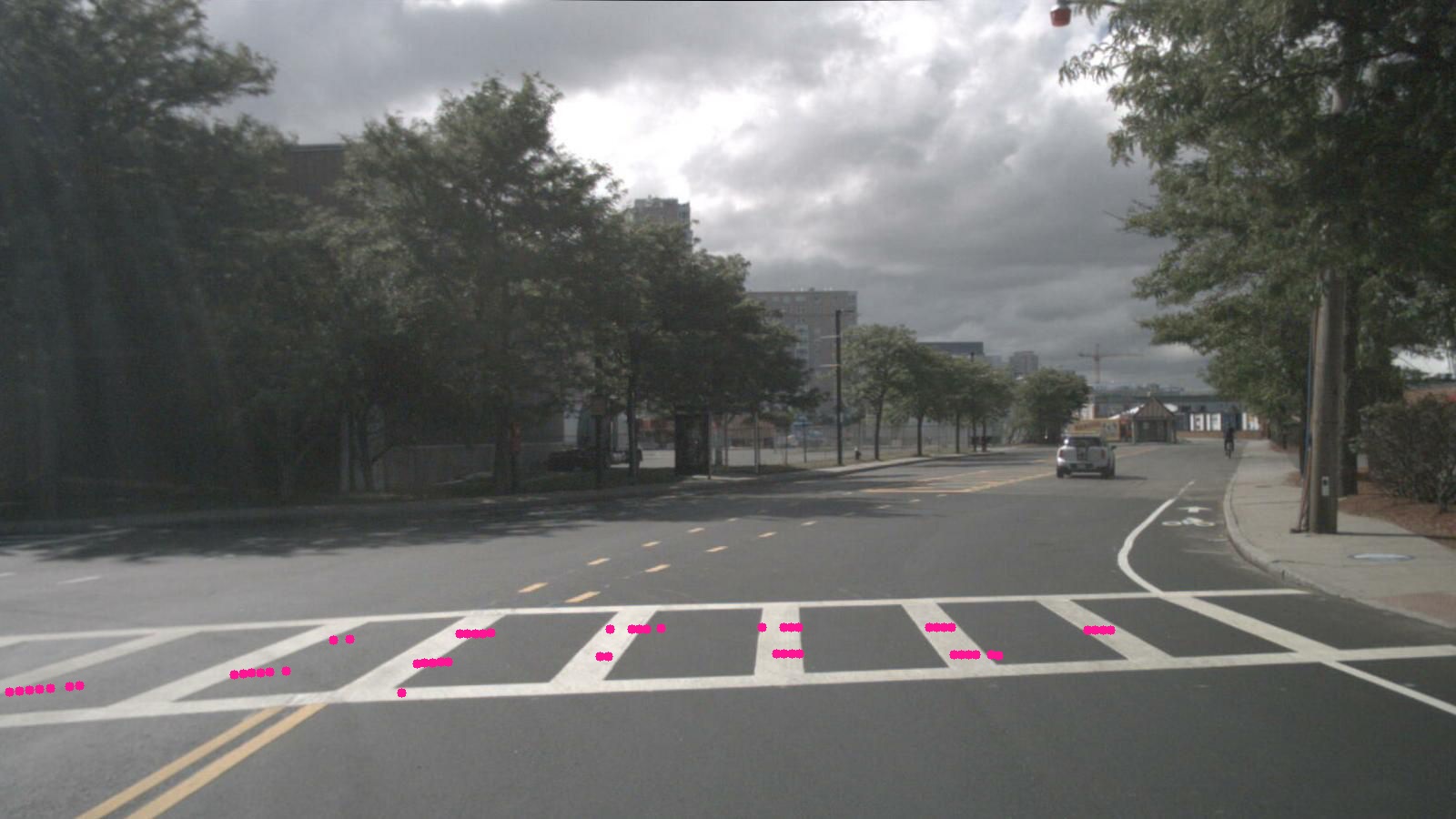} 
    \\
    \tinyafterresize{Wheelchair} & \tinyafterresize{Stroller} & \tinyafterresize{Crosswalk} \\
      \end{tabular}}
      \vspace*{-2mm}
  \caption{
\textbf{Examples of segment retrieval for arbitrary classes.}
As IGLOSS is OV, it can segment any class defined by a free text prompt.
We present (points reprojected onto images) OV results for usually-ignored classes \textit{wheelchair}, \textit{stroller}, and \textit{crosswalk}, which are not included in the 16 official nuScenes classes \cite{nuscenes}. \emph{Crosswalks} are not even identified in nuScenes and are just labeled as any \emph{driveable surface}. Although crosswalk points are geometrically on the road plane, their features identifies them specifically because the 2D-3D distillation takes lidar intensity into account \cite{scalr}.}
\label{fig:other_objects}
\vspace{-3mm}
\end{figure}

It explains why, rather than directly address OVSS in 3D, most 3D methods leverage existing 2D VLMs \cite{radford2021clip} and transfer 2D knowledge to 3D via 2D-3D (pixel-point) correspondences, using datasets where both 2D and 3D modalities are acquired jointly. To obtain a pure 3D model, not requiring image input, the image encoder of the 2D VLM is distilled into a 3D backbone to align point features and pixel features \cite{peng2023openscene}, which are already aligned with text in the VLM.

\begin{figure}[t]
\centering
\includegraphics[width=0.65\linewidth]{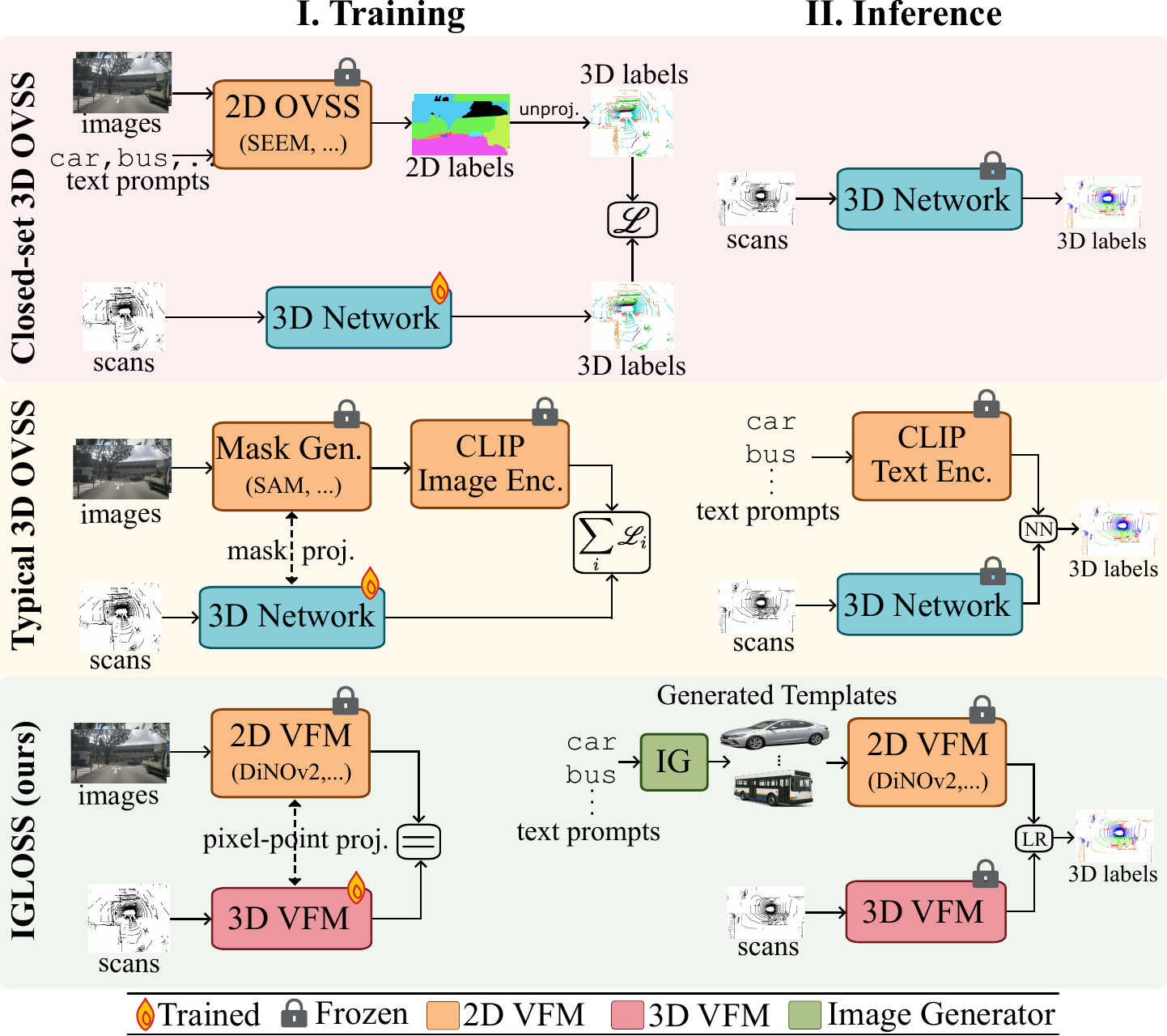}
\caption{\textbf{\ours vs typical 3D OVSS methods.} \ul{\emph{Closed-set 3D OVSS methods}} (top) train a 3D network on pseudo-labels provided by a 2D OVSS method. \ul{\emph{Typical 3D OVSS methods}} (middle) distill a 2D VLM into a 3D network, leveraging some mask generator at training time; inference consists in matching text features to 3D features, on a nearest prompt basis (NN). \ul{\emph{Our method, \ours}} (bottom), exploits a 2D VFM distilled into a 3D network, making it a 3D VFM. Inference consists in generating prototype images from text prompts and matching 2D features of these images to 3D features, using test-time % adaption with 
multinomial logistic regression (LR).}
\label{fig:teaser}
\vspace{-5mm}
\end{figure}

%\paragraph{Our method.} 
While it is hard to escape the image intermediation and the 2D-3D distillation, we propose to use a different approach to bridge text and images \cite{ovdiff, zhao2025dpseg}, as overviewed in \cref{fig:teaser}. % We leverage three foundation models. 
Given a text query, we first use a text-conditioned \emph{image generator (IG)}, such as ChatGPT~\cite{chatgpt5}, to produce images that we use as class prototypes. Second, using a 2D \emph{Vision Foundation Model (VFM)} that excels at dense downstream tasks, such as DINOv2 \cite{oquab2024dinov2}, we produce pixel-level features for the images, which we average and use to fit a multinomial logistic regression (LR) at test time. Third, given a 3D VFM distilled from the same 2D VFM, such as ScaLR \cite{scalr}, we use the LR classifier to assign labels to points according to their 3D features, which are by design aligned with the 2D features.

\renaud{A number of methods have been proposed to pretrain a 3D network from a pretrained 2D network, based on image-scan alignment. However, many of them use a contrastive loss \cite{slidr, stslidr, seal, xu2024superflow, kong2026largead}, possibly with extra loss terms \cite{cleverdistiller}, which produces strong 3D features for downstream tasks but does not specifically enforce 2D-3D feature alignment.} % , contrary to methods like ScaLR \cite{scalr}.} 
To construct such an alignment, \renaud{which is essential to our method,} we % actually
build \scalrplus, a 2D-3D distillation scheme improving over ScaLR~\cite{scalr}.

The advantages of our method, named \ours (acronym of the paper title), are as follows:

\stepcounter{benef}\edef\easyitalign{\strength{\arabic{benef}}}
%\textbf{\easyitalign.\,\,\,No difficult image-text alignment.} 
\textbf{\easyitalign.\,\,\,Versatile image-text alignment.} 
Our method does not rely on the classical image-text alignment \cite{radford2021clip} used by most other 3D OVSS methods. It thus does not suffer from the inherent and recognized modality gap between images and text, as shown by theoretical and empirical studies \cite{liang2022modalitygap, qian2023inmap, yamaguchi2025cliprefine, kang2025isclipideal}. This issue is only partly addressed by alternative pretraining architectures \cite{you2022msclip, chen2023fdt}, additional losses \cite{lee2022uniclip, goal2022cyclip} and specific training recipes \cite{lee2022uniclip, li2022declip, lee2022uniclip, doveh2023svlc}. Direct VFM-text aligments have been proposed \cite{jose2025dinov2meetstext, simeoni2025dinov3}, but preserving the nature of this this gap. Besides, they have not the same feature space as DINOv2, preventing our use of ScaLR.
In \ours, text and image are bridged with the versatility of image generation from text \cite{zhao2025dpseg}, and 2D-3D alignment operates on closely related visual modalities.

%, which captures intra-class distribution and inter-class textual ambiguity, and 
% Besides,the 2D-3D alignment operates in close modalities pertaining to vision.
% \nermin{Besides, the 2D-3D alignment operates between closely related vision-based modalities.}

\stepcounter{benef}\edef\nohugedata{\strength{\arabic{benef}}}
\textbf{\nohugedata.\,\,No large dataset to curate, no extra supervision.} 
Bridging text and image in a VLM requires a large amount of curated data, pairing images and captions \cite{radford2021clip}. Alternatively, aligning them with less data requires supervision, e.g., with object boxes \cite{gdino, openseed} or segmentation masks \cite{openseg, zou2023seem, openseed, sam}. In contrast, our method relies on image generation from text, which arguably requires less image-caption pairs \cite{degeorge2025howfarcanwego} and is less sensitive to noisy (meaningless) captions \cite{dufour2024dont}. 

\stepcounter{benef}\edef\betterfeat{\strength{\arabic{benef}}}
\textbf{\betterfeat.\,\,\,Better image features.} 
Early 2D VLMs were trained at image level which suits downstream tasks like image classification or retrieval \cite{radford2021clip}. For dense tasks like segmentation, additional apparatuses have to be used \cite{zhou2022maskclip, li2022lseg, ding2023maskclip, mukhoti2023pacl, dong2023maskclip, bousselham2024gem, shao2024cliptrase}. Nevertheless, imprecise segment boundaries led a number of 3D OVSS methods, including recent ones \cite{chen2023cns, osep2024sal, zhang2025sal4d, zou2025adaco, gebraad2025leap, li2025sas}, to integrate some form of denoising or use an additional FM for segmentation, such as SAM \cite{sam}. 
Meanwhile, image VFMs such as DINOv2 provide state-of-the-art (SOTA) dense encoders for segmentation tasks \cite{oquab2024dinov2}, pushing some to improve a VLM via a VFM teacher \cite{wysoczanska2024, stojnic2025lposs} or equip a VFM with text alignment \cite{jose2025dinov2meetstext}. 
Although recent 2D VLMs are stronger \cite{fang2024eva02, tschannen2025siglip2}, 2D VFMs remain SOTA for downstream segmentation tasks \cite{simeoni2025dinov3}.

\stepcounter{benef}\edef\bboxfm{\strength{\arabic{benef}}}
\textbf{\bboxfm.\,\,\,Foundation models (FMs) as black boxes.}
Existing 3D OVSS methods require ``opening the box'' of FMs, e.g., to get dense CLIP features \cite{zhou2022maskclip, ding2023maskclip} or to obtain masks via a cross-attention map in Stable Diffusion (SD) \cite{rombach2022sd}.
In contrast, our method uses FMs in a \emph{black box} manner: \emph{any} text-to-image generator, \emph{any} 2D VFM, and \emph{any} 3D backbone distilled from the same 2D VFM can do.

\stepcounter{benef}\edef\simplicity{\strength{\arabic{benef}}}
\textbf{\simplicity.\,\,\,Simplicity.}
Most 3D OVSS methods \cite{chen2023cns, kang2024hicl, zou2025adaco, gebraad2025leap, samet2026losc, wang2024ggsd, sun2025afov, li2025sas, jiang2024ov3d, wei20253davs, osep2024sal, zhang2025sal4d} are fairly complex, combining several other methods to extract masks or captions, adding multiple distillation losses for alignment and regularization, including modules to denoise labels or features, or requiring several training stages or recipes (details in appendix).  
In contrast, \ours is the simple composition of three FMs, 
which already exist, although we also improve the 3D VFM.

\stepcounter{benef}\edef\notrick{\strength{\arabic{benef}}}
\textbf{\notrick.\,\,\,No need to input accumulated scans or images.}
Some of the top-performing 3D OVSS methods operate on a sequence of scans \cite{clip2scene, zhang2025sal4d, wei20253davs, zou2025adaco}, or get their best results by additionally using images or scan sequences at test time \cite{clip2scene, peng2023openscene, jiang2024ov3d, wei20253davs, sun2025afov}. To reach the SOTA, our method does not need such strategies, that have been argued to have high storage and computational costs \cite{wang2024ggsd}. 

\stepcounter{benef}\edef\openvoc{\strength{\arabic{benef}}}
\textbf{\openvoc.\,\,\,Vocabulary and semantic concept openness},
\emph{for all methods,} is bounded by their training sets. For 3D OVSS, the bottleneck is the 2D-3D correspondence dataset, which is in practice much smaller than 2D VLM training sets. Only things seen at train time (yet unlabeled) have a chance to be segmented at inference. As it uses a 2D-3D VFM trained on several datasets, \ours incurs less bias from single datasets used to train 2D-3D distillation in other methods.

Some methods use text prompts to generate 2D pseudo-labels and train in 3D, 
making their model closed-set \cite{chen2023cns, kang2024hicl, zou2025adaco, gebraad2025leap, samet2026losc}. Other methods build a model that can \emph{functionally} use \emph{any} text prompt, but that is actually trained using fixed prompts \cite{clip2scene, wang2024ggsd, sun2025afov, li2025sas}, introducing a \emph{bias towards predefined classes}. Still others use an image captioner at training time instead of fixed prompts, leading to trained 3D backbones \emph{biased to classes the captioner can recognize} \cite{jiang2024ov3d, wei20253davs}. (See \appref{sec:moredetail}.)  

We argue that captioning an image \emph{with details}, especially rich driving scenes, is harder than
getting strong image features 
or generating an image for a class name. Besides, driving scenes are often less represented in VLM training data due to the scarcity of text alignment, while DINOv2 easily trains on images only. 
In contrast, \ours has no such bias.
\nermin{Moreover}, text-image correspondence in IG is particularly flexible and productive.
\cref{fig:other_objects} shows results for ``arbitrary'' classes. 

\stepcounter{benef}\edef\easyprompt{\strength{\arabic{benef}}}
\textbf{\easyprompt.\,\,\,No need for heavy prompt engineering.}\,%
VLMs are sensitive to text prompts, e.g., pushing CLIP authors to ensemble 80 different \emph{context prompts} (per actual prompt) to improve their results \cite{radford2021clip}. It led to a whole literature on prompt engineering \cite{openseg, gu2022ovobjdet, shu2022tpt, roth2023waffling, parisot2023learning, lin2023clipes, vobecky2023pop3d, huang2024renovate, esfandiarpoor2024ifclipcouldtalk, lafon2024gallop, benigmim2025floss}\rlap. We argue that text-conditioned image generation is less sensitive to prompt formulation. Broad class names (e.g., \emph{manmade} \cite{nuscenes}) and negative definitions (e.g., \emph{other flat}) have however to be made explicit, as other OV methods do too.
While zero shot does not mean zero effort, IGLOSS has the simplest prompt templates (\cref{tab:rwprompts}).

\begin{table*}[t!]
\centering\tabcolsep 3pt
\caption{\textbf{Characteristics of 3D OVSS methods.}
\ul{\emph{Image-text bridge}} and \ul{\emph{Extra foundation models used:}} \quot\ift~= image and text encoders, producing aligned features; \quot\igm~= image mask generator; \quot\igt~= image captioner; \quot\itgb\ (resp.\ \quot\itgm)~= image object detector producing bounding boxes (resp.\ masks) given classes defined with text prompts; \quot\tgi~= image generator conditioned on text; \quot\pgf~= point encoder.
\ul{\emph{Label assignment:}} nearest neighbor (NN) to prototype features, class probability (CP) of underlying model, or multinomial logistic regression (LR) on prototype features. \ul{\emph{OV at test time:}} \xmark~= the model is closed-set; \mmmark~= the model is biased at training time with a pre-defined vocabulary or with text prompts, but remains functionally OV; \mmark~= the model is biased by a captioner but remains OV; \cmark~= the model is fully OV, the classes of interest are decided at test time. \ul{\emph{No image at test time:}} \cmark~= input is only 3D points; \mmark~= 2D-3D ensembling for best results. \ul{\emph{No scan sequence at test time:}}  \cmark~= yes; \xmark~= no. \ul{\emph{Code/checkpoint available:}} \codeavail~= yes for both nuScenes (NS) and SemanticKITTI (SK); \codeavailnoSK~= for NS, not for SK; \codeavailissues~= partial or with long-standing open issues; \nocodeavail~= no; $^\dagger$: upon publication.
}
\resizebox{0.96\linewidth}{!}{
\begin{tabular}{l@{~}r|cl@{~}ccccc} % Open-voc at test time, 
\toprule
        && Image   & Extra         & Label   & OV     & No img      & No scan    & Code/\\
\multicolumn{2}{l|}{3D OVSS}  
        & -text    & foundation    & assign- & at test  & at test & seq.\ at  & chkpt \\
\multicolumn{2}{l|}{method}  
        & bridge   & model(s) used & ment    & time & time & test time & avail. \\
% was No VLM for image-to-text generation
\midrule
CNS & \cite{chen2023cns}
        & \ift     & \igm          & NN & \xmark    & \cmark & \cmark & \nocodeavail \\
HICL & \cite{kang2024hicl}
        & \ift     &               & NN & \xmark    & \cmark & \cmark & \nocodeavail \\ % + \igm = VCCS 
AdaCo & \cite{zou2025adaco} %%% SEQ
        & \igt     & \igm, \tgf  & NN & \xmark    & \cmark & \cmark & \nocodeavail \\
AFOV & \cite{sun2025afov}
        & \ift     & \igm          & CP & \xmark    & \mmark & \cmark & \codeavailnoSK \\
LeAP & \cite{gebraad2025leap}
        & \itgb    & \igm          & CP & \xmark    & \cmark & \cmark & \nocodeavail \\
LOSC & \cite{samet2026losc}
        & \itgm    &               & CP & \xmark    & \cmark & \cmark & \nocodeavail \\
\midrule
CLIP2Scene & \cite{clip2scene}
        & \ift     &               & NN & \mmmark   & \cmark & \xmark & \codeavailissuesnoSK \\
GGSD & \cite{wang2024ggsd}
        & \ift     &               & NN & \mmmark   & \cmark & \cmark & \codeavailissuesnoSK \\ 
SAS & \cite{li2025sas}
        & $($\ift$)$\rlap{$^k$} & \igm, \igt, \tgi & NN & \mmmark & \cmark & \cmark & \codeavailissuesnoSK \\ 
\midrule
OV3D & \cite{jiang2024ov3d} 
        & \ift     & \igm, \igt  & NN & \mmark    & \cmark & \cmark & \nocodeavail \\
\multicolumn{2}{l|}{OV3D\,w/\,OpenScene} 
        & \ift     & \igm, \igt  & NN & \mmark    & \mmark & \cmark & \nocodeavail \\
3D-AVS & \cite{wei20253davs} %%% SEQ
        & \ift     & \igt, \pgf  & NN & \mmark    & \mmark & \xmark & \codeavailissuesnoSK\\
\midrule
OpenScene & \cite{peng2023openscene} %%% SEQ
        & \ift     &               & NN & \cmark    & \mmark & \cmark & \codeavailnoSK \\
SAL & \cite{osep2024sal}
        & \ift     & \igm          & NN & \cmark    & \cmark & \cmark & \nocodeavail \\
\rowcolor{orange!18}
\ours & (ours)
        & \tgi     & \ifp          & LR & \cmark    & \cmark & \cmark & \codeavail\rlap{$^\dagger$} \\
\bottomrule
\end{tabular}
}
\label{tab:relwork}
%\vspace{-1mm}
\end{table*}

\stepcounter{benef}\edef\annotinstr{\strength{\arabic{benef}}}
\textbf{\annotinstr.\,\,\,Adequacy to annotation instructions.}
Instead of images generated from text prompts, our method also applies to exemplar images as provided, e.g., in annotation instructions \cite{nuscenesannotinstruct}.
This is meaningful if we assume some truth in the saying that ``one image is worth one thousand words''.

\stepcounter{benef}\edef\sota{\strength{\arabic{benef}}}
\textbf{\sota.\,\,\,SOTA.}
Thanks to a better text-pixel alignment, an enhanced 2D-3D distillation, and the use of logistic regression for label assignment, we reach state-of-the-art results in zero-shot 3D OVSS on both nuScenes~\cite{nuscenes} and SemanticKITTI~\cite{semantickitti}.
{Optionally}, for an improved performance, our method can be complemented by a self-training stage \cite{zou2018selftraining}; the resulting model however loses then vocabulary openness and becomes closed-set. Our method also very easily supports the mixing of several 2D-3D VFMs for an increased performance.

\section{Related work}
\label{sec:rw}

The introduction largely reviews related work. We focus here on OV methods using image generation (IG). For a more comprehensive discussion of related work, please refer to \appref{sec:moredetail}.

Among 3D OVSS methods, only SAS \cite{li2025sas} uses IG, but for different purposes than \ours. It uses Stable Diffusion \cite{rombach2022sd} (SD) to score VLMs in order to ensemble them.

A few uses of IG to produce OV prototypes can be found in 2D. To achieve zero-shot OV image classification, SuS-X \cite{udandarao2023susx} uses few-shot classification techniques \cite{zhang2022tipadapter}, providing a support set made of images generated from SD or retrieved from a vision-language data bank \cite{schuhmann2022laion5b}. Other classification methods \cite{li2023diffusionclassifier, clark2023texttoimage} open the SD network, choosing the text conditioning that best predicts the noise added to the input image. It is however not applicable to segmentation.

Regarding 2D OV segmentation, DPSeg \cite{zhao2025dpseg} tightly integrates in the segmentation architecture features of visual prompts obtained by generating images from text prompts.
Closer to \ours, OVDiff \cite{ovdiff} addresses 2D OVSS by using Stable Diffusion (SD) for both generation and segmentation. It generates foreground and background prototypes by masking images with CutLER \cite{cutler} guided by SD cross-attention. Features are then extracted using DINO, CLIP, and SD \cite{caron2021dino, radford2021clip, rombach2022sd} and clustered via k-means. Finally, OVDiff uses ChatGPT for class categorization, CLIP for image filtering, and a rejection mechanism to ensure foreground and background prototypes remain distinct.

In contrast, \ours is much simpler: we ask the image generator to generate images on a white background, avoiding the need for explicit prototype segmentation. No additional models, sliding window, part-level clustering, foreground/background separation, or thing/stuff split are required; we only average patch features within each image. 

\section{Method}
\label{sec:method}

Our method generates prototype images from text prompts and labels the point cloud by matching 3D features with 2D features of these prototypes (\cref{fig:teaser}).
We describe here the 2D-3D distillation and the two inference stages: image prototype generation and point-cloud segmentation. We also show an annotation-free extension for closed-set segmentation and an easy way to ensemble VFMs.

\subsection{2D-3D Visual Foundation Model alignment}
\label{sec:2d3dvfmdistill}

Our key ingredient is a pair $\langle M,M'\rangle$ of a 2D VFM $M$ and an aligned 3D VFM~$M'$\!\!. Some are publicly available, like \lrangle{OpenSeg, OpenScene} \cite{openseed, peng2023openscene}, as part of existing 3D OVSS methods, % . They are 
aligned with CLIP text features, which adds a constraint on 2D features that we do not require.

\simpletextparagraph{ScaLR.}
One of the most prominent alignment of a strong 2D VFM and a 3D network is \lrangle{DINOv2, ScaLR}. While ``ScaLR'' is originally the name of 2D-3D distillation method \cite{scalr}, it is also understood as is a multi-domain ScaLR distillation of DINOv2 \cite{oquab2024dinov2} into WaffleIron \cite{waffleiron}, jointly trained on several lidar datasets: nuScenes \cite{nuscenes}, SemanticKITTI \cite{semantickitti}, Panda\-64 and PandaGT \cite{xiao2021pandaset}. As a pretrained 3D backbone \cite{scalr}, ScaLR generalizes to other datasets, e.g., SemanticPOSS \cite{pan2020semanticposs}. Other authors \cite{michele2025muddos} have retrained it to include WOD \cite{ettinger2021waymoopen}. 

\simpletextparagraph{{\scalrplus}.}
Although the supervision signal of ScaLR is based on the similarity between pixel and point features, ScaLR, \emph{as a general 3D VFM,} is not made to output 3D features that resemble 
the most to the 2D features; it is designed and tuned to produce  strong general 3D features, which are actually extracted before the 3D-2D projection head \cite{scalr}, and thus dissimilar from the 2D features. 

To get a 3D network that mimics as much as possible the 2D VFM is crucial to our approach. To that end, we modify the ScaLR architecture: we reintroduce the (linear) projection head and extend it to an MLP; our features are extracted after this head. Additionally, to further improve performance, we improve the representation learning by replacing ReLUs by GELUs \cite{hendrycks2016gelu} and adding a drop path regularization \cite{huang2016stochasticdepth}. We also increase the resolution of images and train for longer. We call the resulting model \scalrplus. The ablation study justifying these choices appears in \cref{tab:scalrplusablat}.

\subsection{Prototype generation}
\label{sec:protogen}

Our second ingredient is a text-conditioned image generator (IG), to generate prototype images from text prompts defining classes of interest (cf.\ \cref{fig:templates}).

\simpletextparagraph{Explicit prompting.}
Given a set of classes ${C} = \{C_1, C_2, \dots, C_N\}$ to segment, we first build corresponding text prompts. It is common to find broad classes that contain widely different subclasses. For instance, in nuScenes, the \emph{manmade} class is defined in the annotator instructions \cite{nuscenesannotinstruct} to contain things as different as building, wall, stairs, pole, fire hydrant, traffic sign and traffic light. It does not matter when training with full supervision, but it does matter when trying to represent a class with a \emph{single} prototype. Therefore, contrary, e.g., to the \emph{bicycle} class, which can easily be defined by just one prototype, the \emph{manmade} class has to be split into different subclasses, defined with separate prototypes, as all other 3D OVSS methods also do. Such subclasses $\{c_{i,1}, c_{i,2}, \dots, c_{i,N_i}\}$ are mapped to their corresponding class $C_i$ at a later stage, for evaluation. This applies as well to ``negative'' classes, e.g., \emph{other flat} in nuScenes, which we actually define as \emph{traffic island}, following again the nuScenes' annotator instructions \cite{nuscenesannotinstruct}.

% --- Image cell size (edit these) ---
% \newlength{\cellw}
% \newlength{\cellh}
\setlength{\cellw}{1.0cm} 
\setlength{\cellh}{1.0cm} 

\newcommand{\cellimgg}[1]{%
  \adjincludegraphics[width=\cellw,height=\cellh,keepaspectratio]{#1}%
}

\begin{figure}[t!]
\setlength{\tabcolsep}{2pt}
\renewcommand{\arraystretch}{1.0}
\resizebox{1.0\linewidth}{!}{
  \centering

  \begin{tabular}{*{16}{>{\centering\arraybackslash}m{\cellw}}}
    % Row 1

    \cellimgg{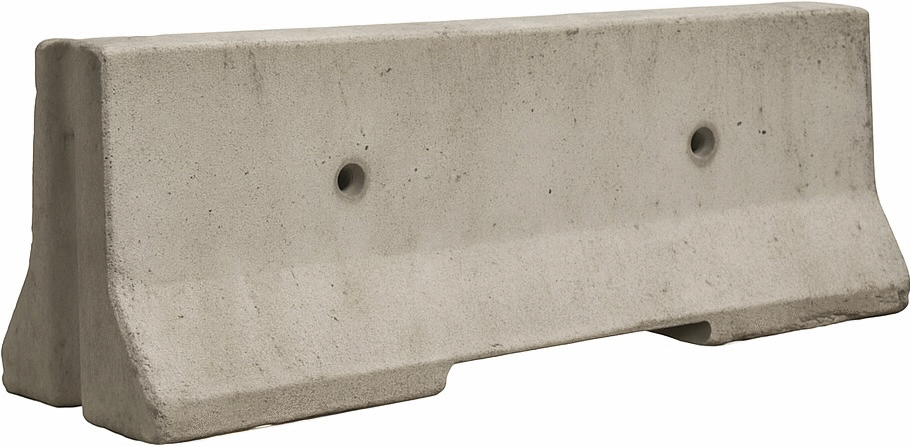} 
    & \cellimgg{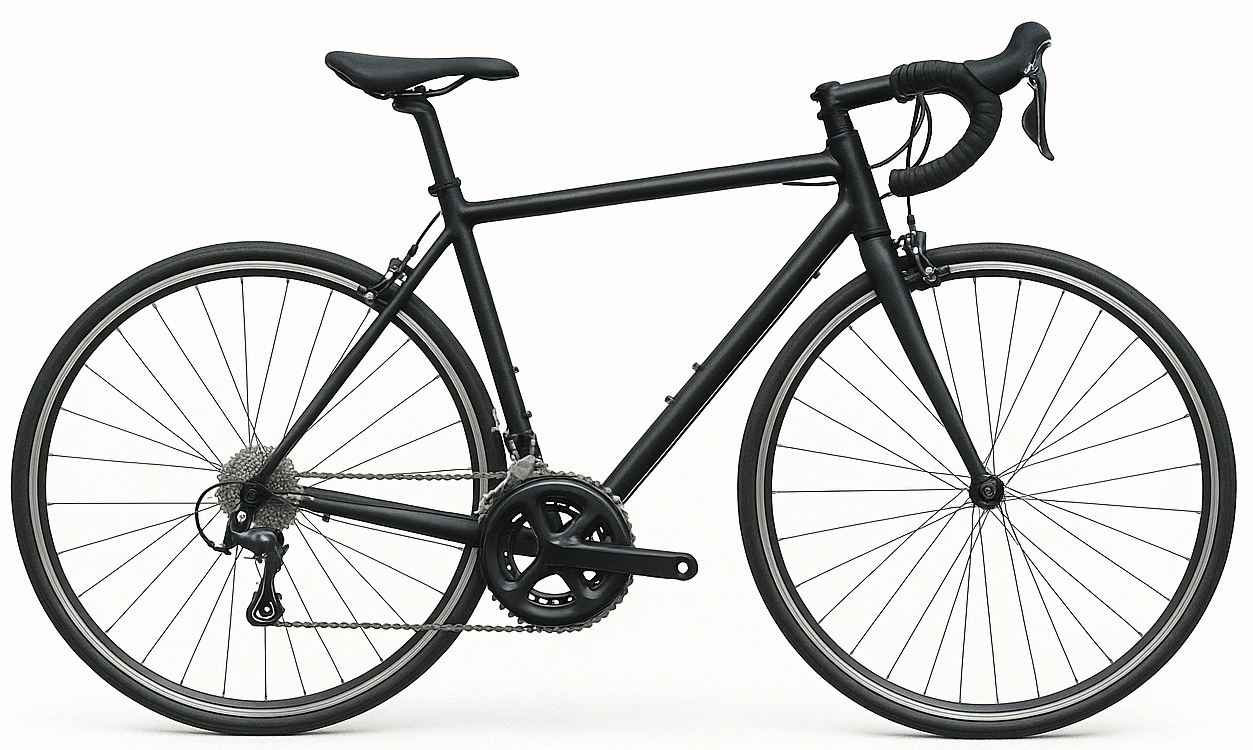} 
    & \cellimgg{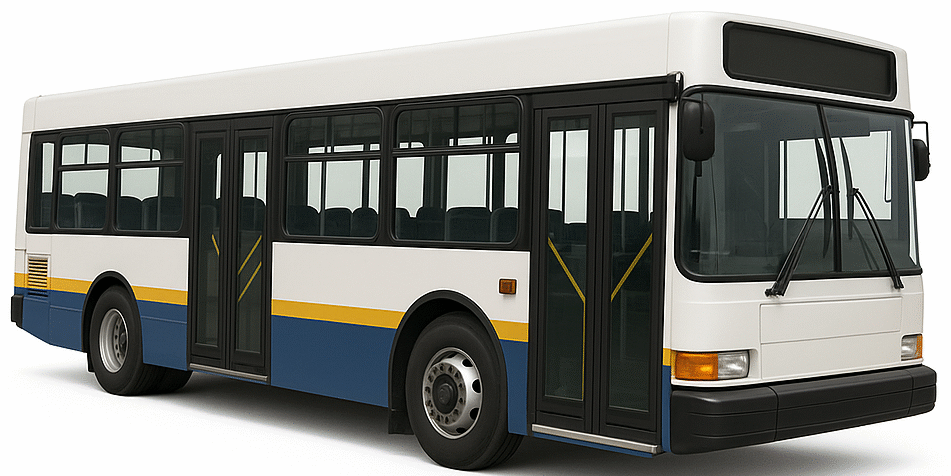} 
    & \cellimgg{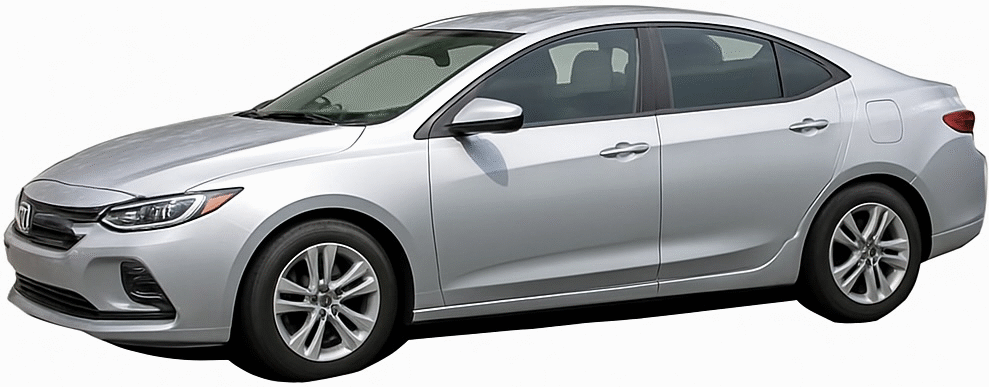} 
    & \cellimgg{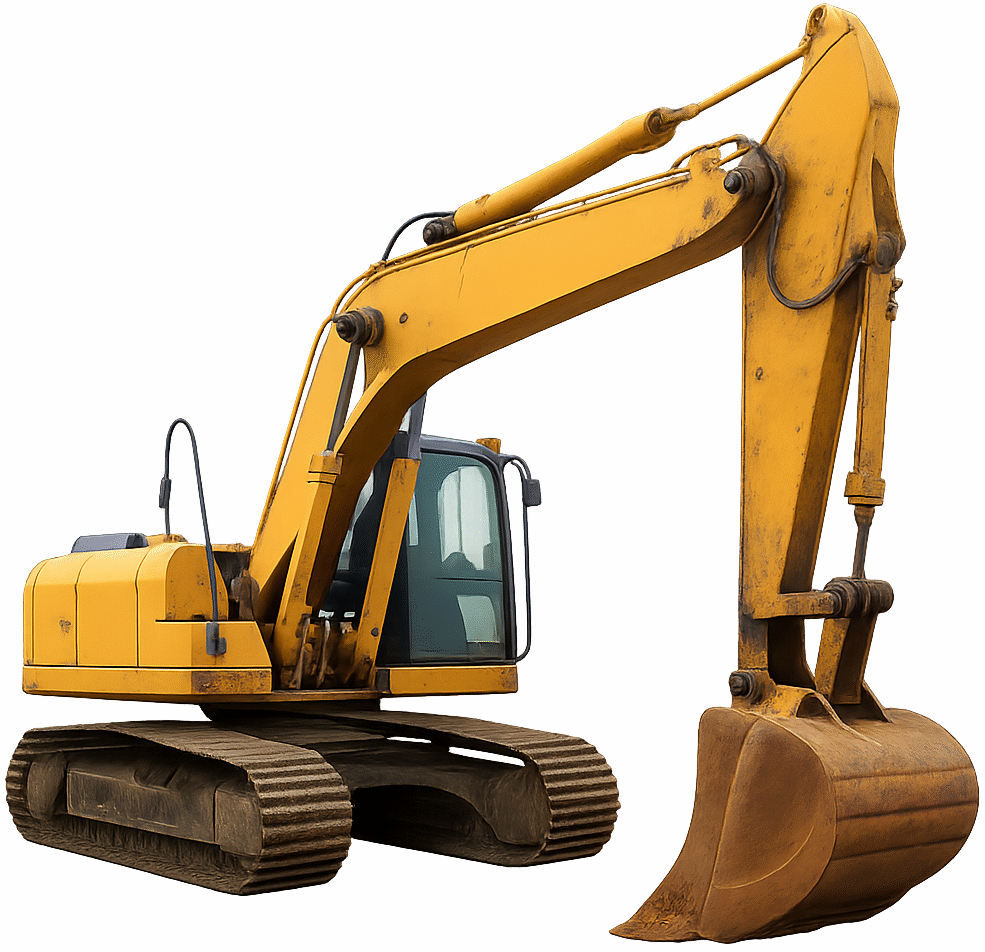} 
    & \cellimgg{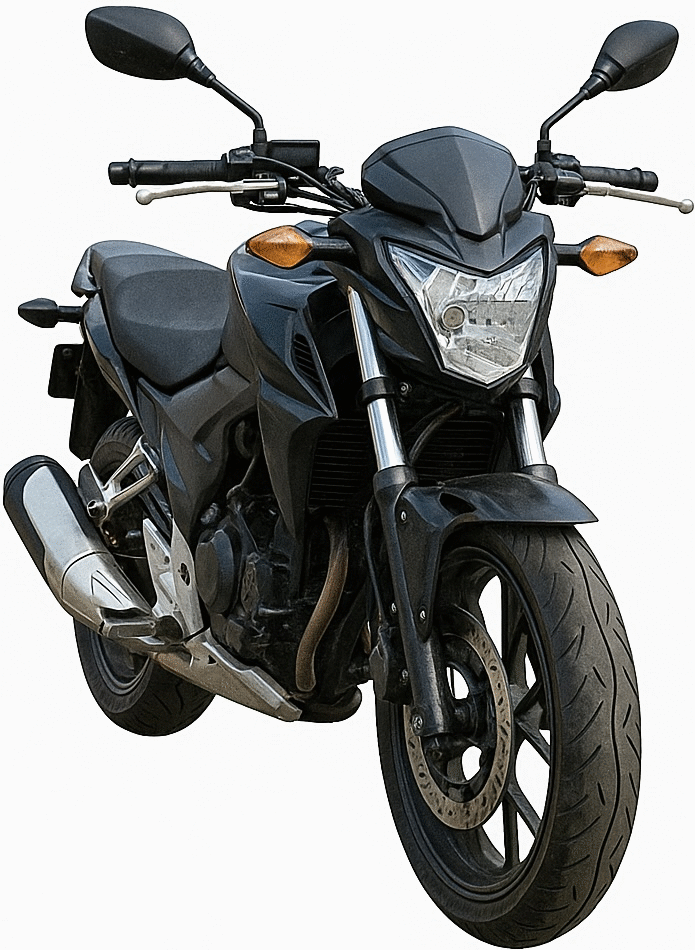} 
    & \cellimgg{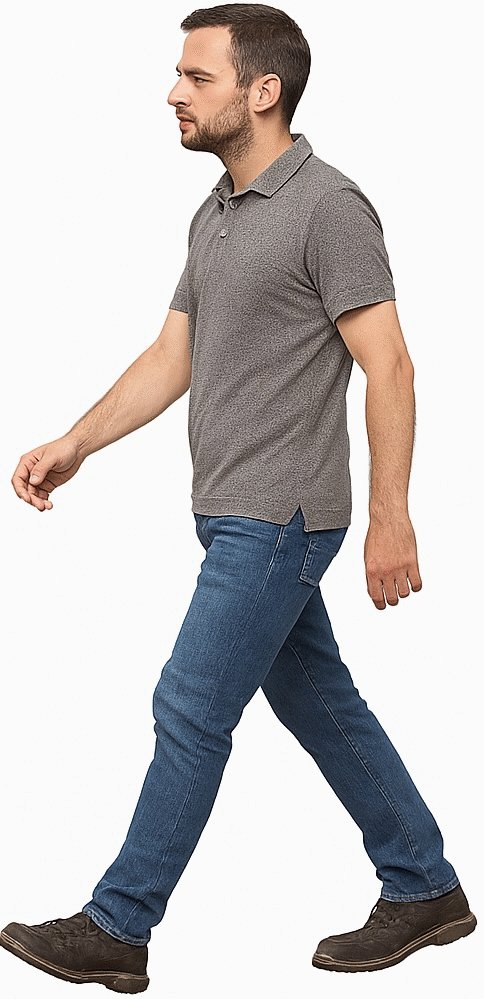} 
    & \cellimgg{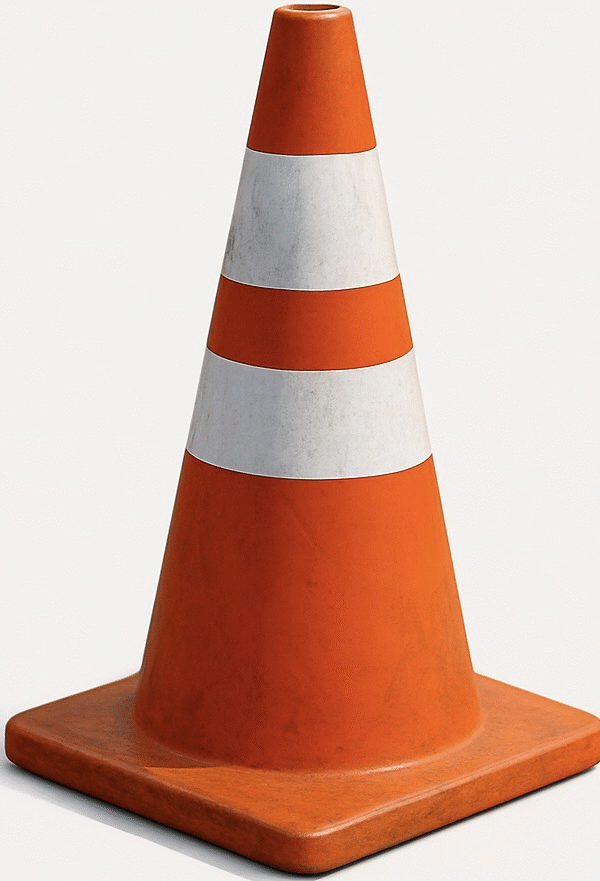} 
    & \cellimgg{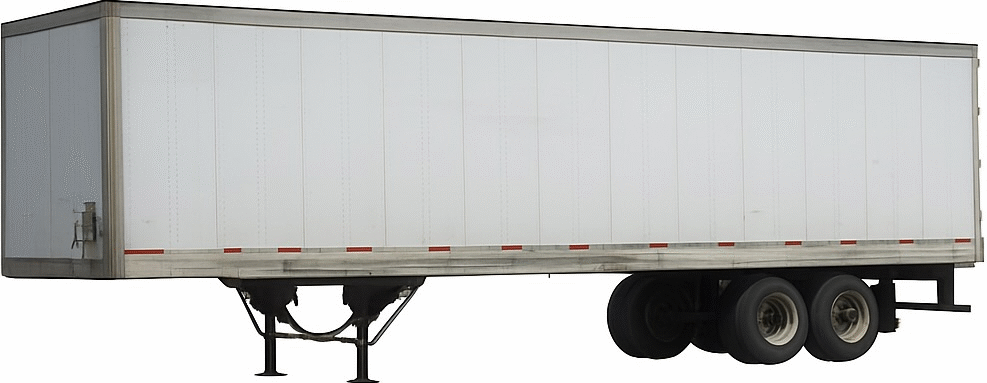} 
    & \cellimgg{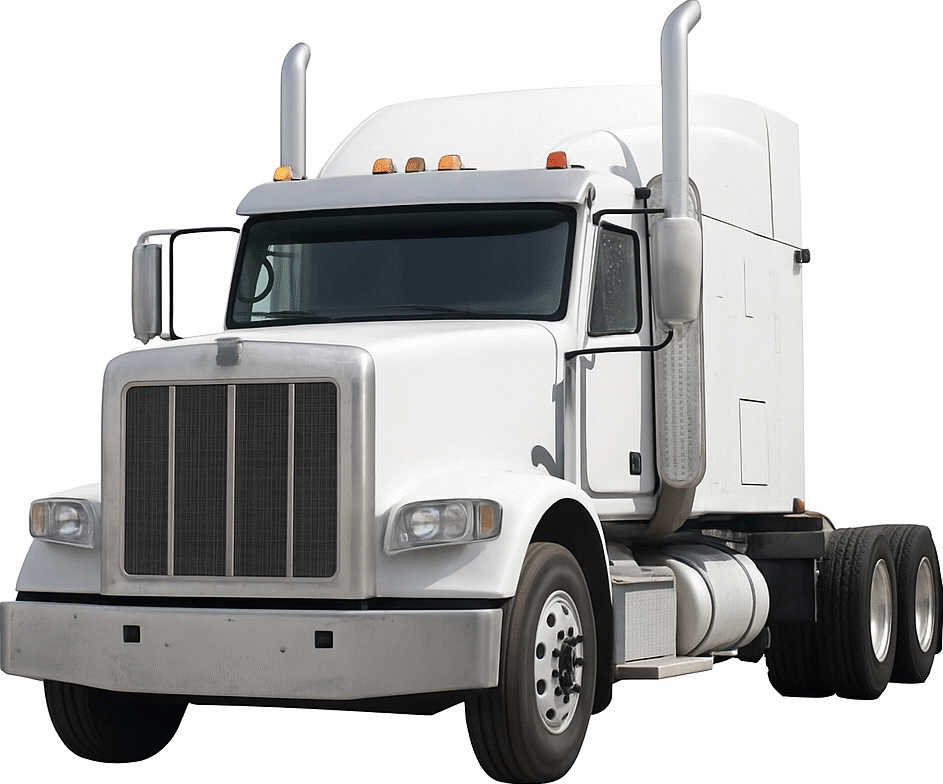} 
    & \cellimgg{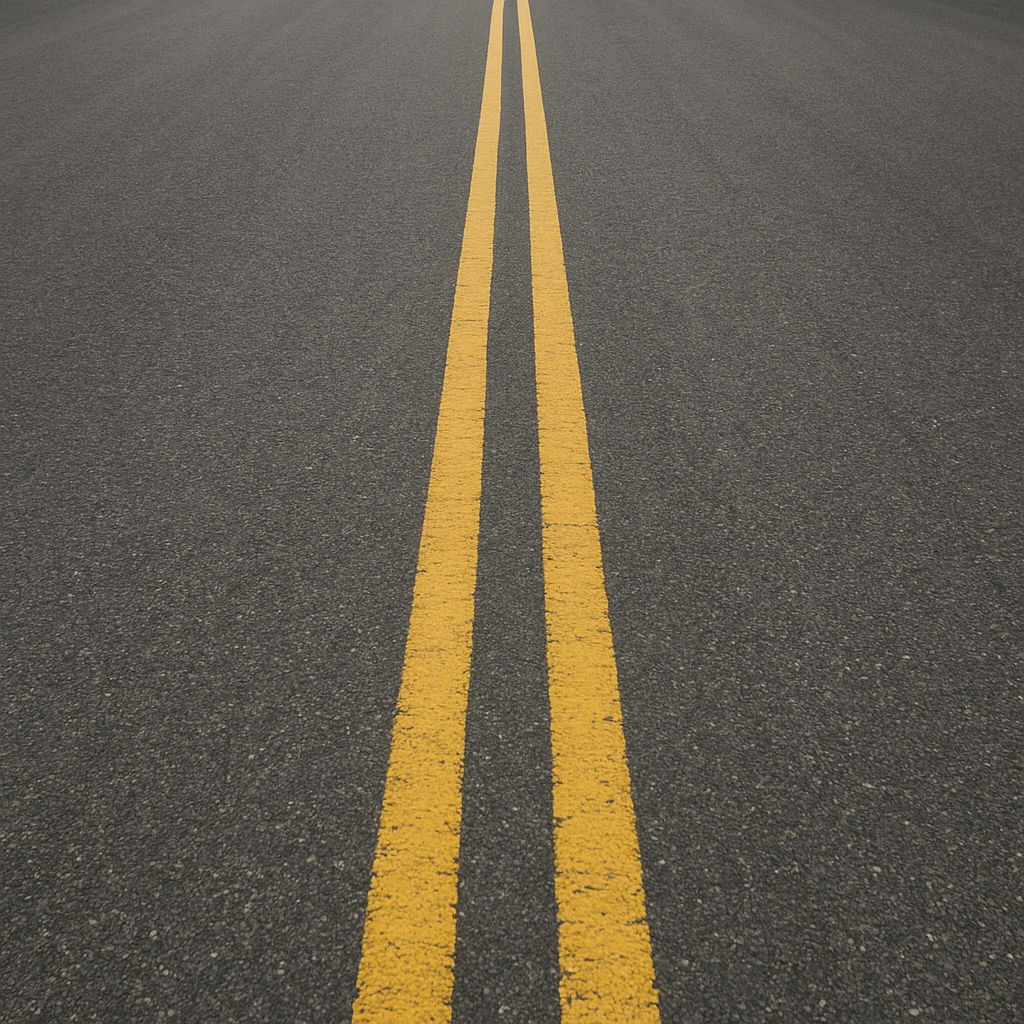} 
    & \cellimgg{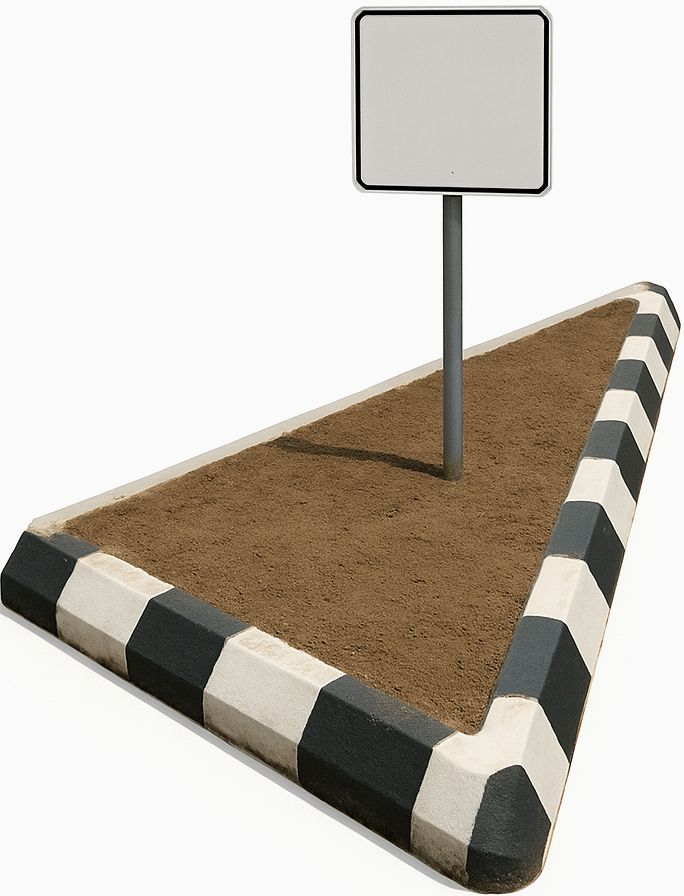} 
    & \cellimgg{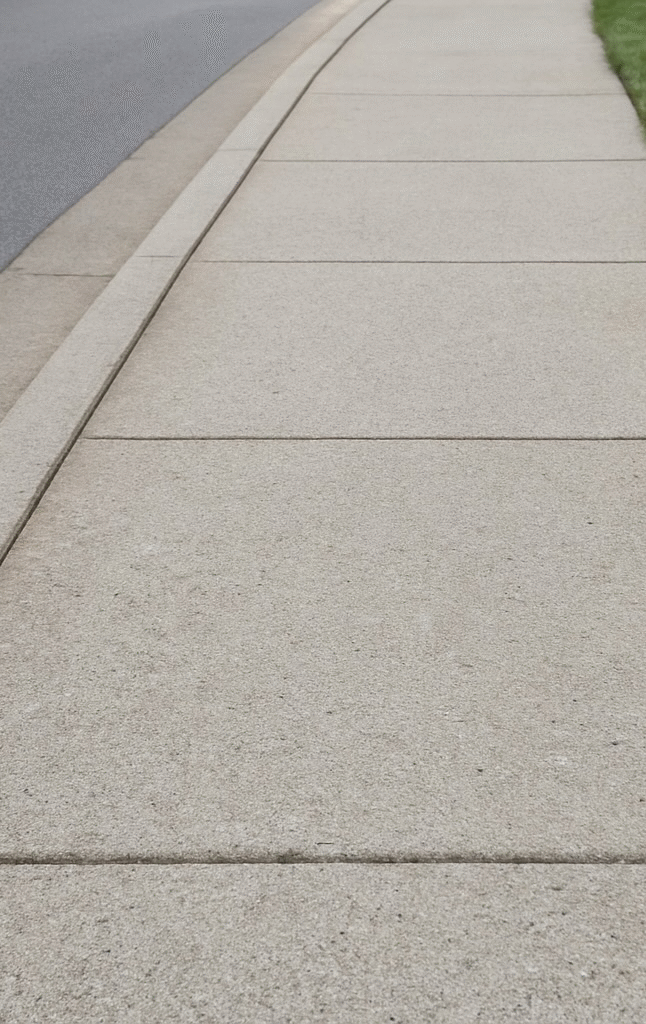} 
    & \cellimgg{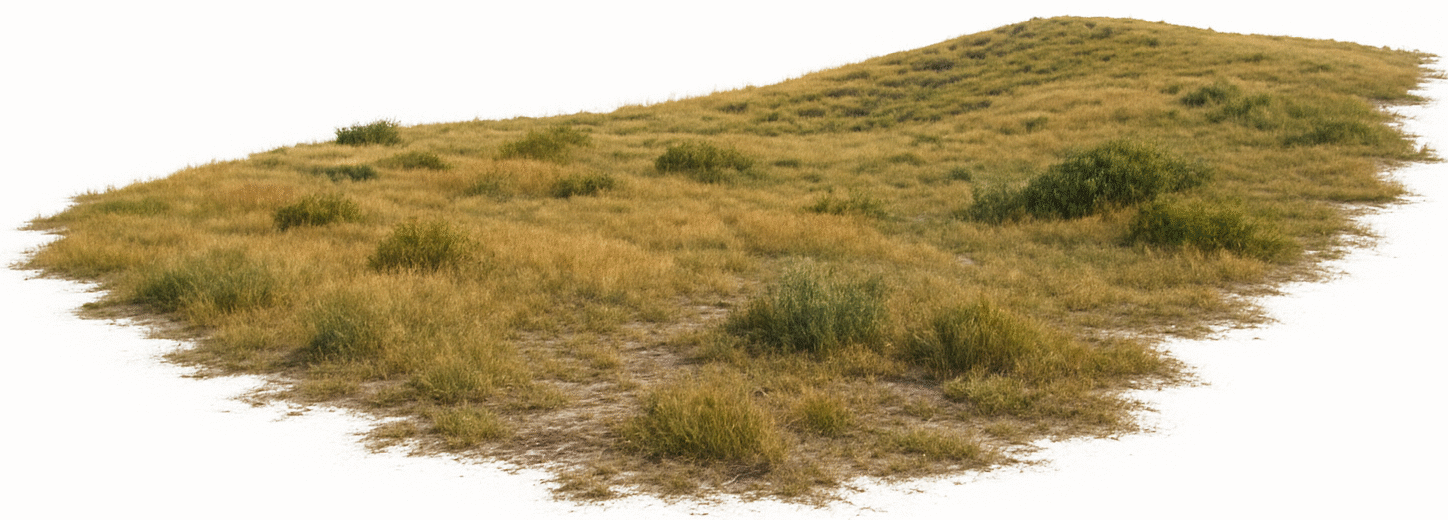} 
    & \cellimgg{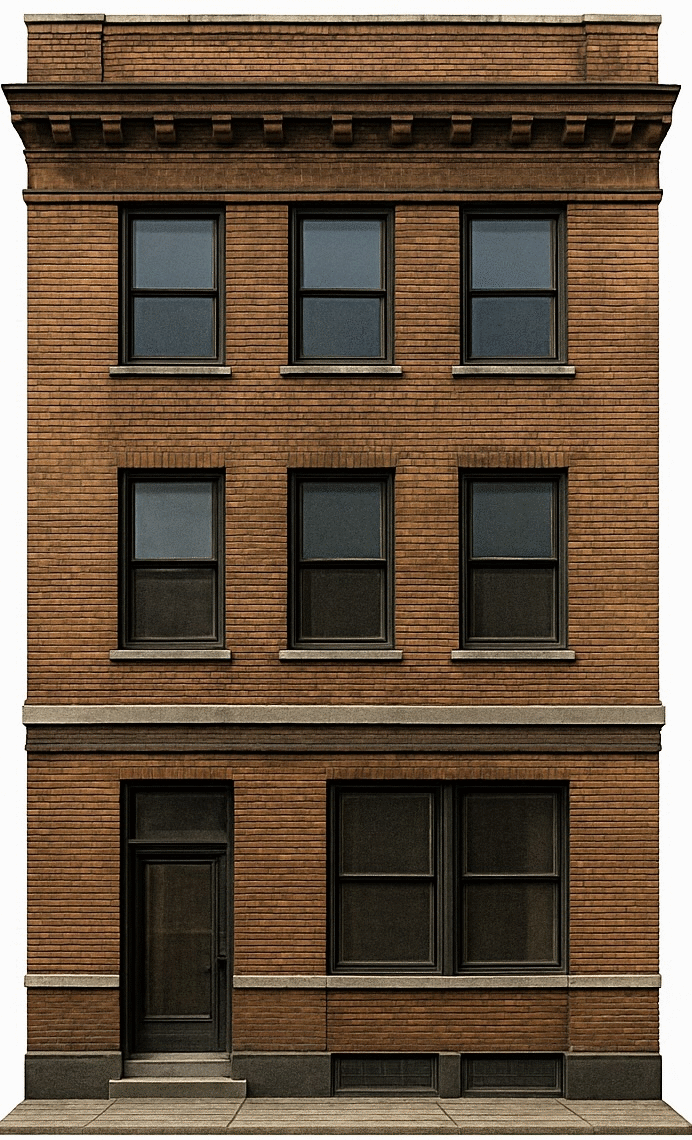} 
    & \cellimgg{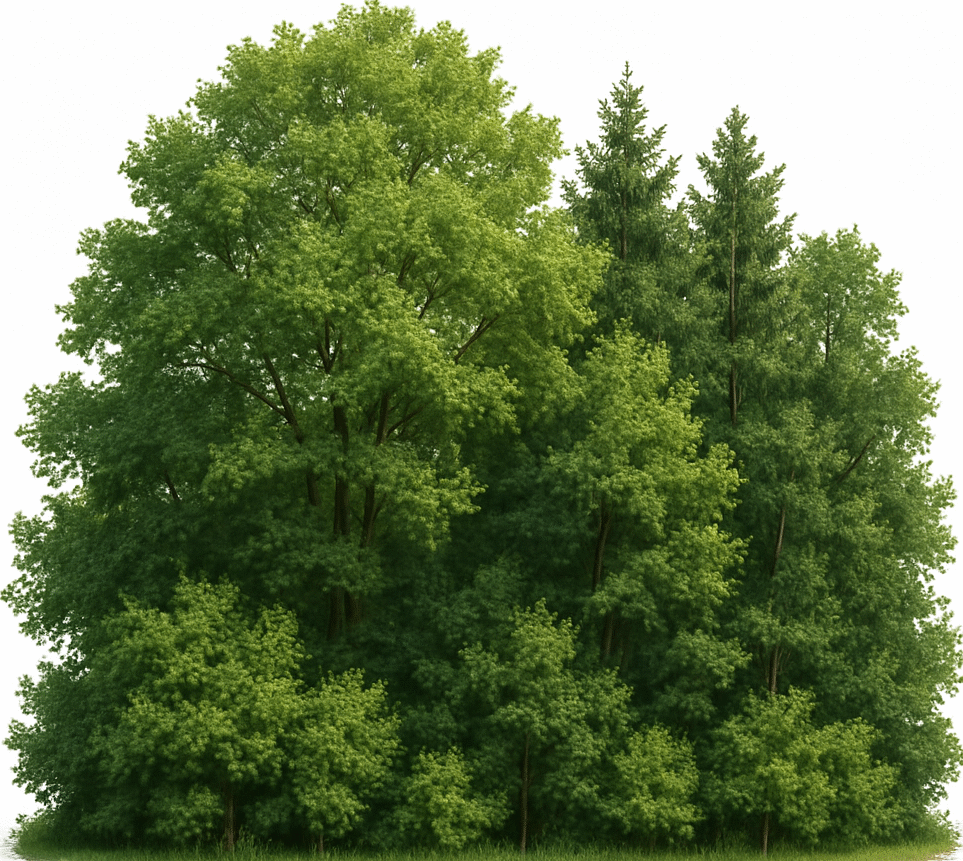} \\
   
 % Row 2
    \cellimgg{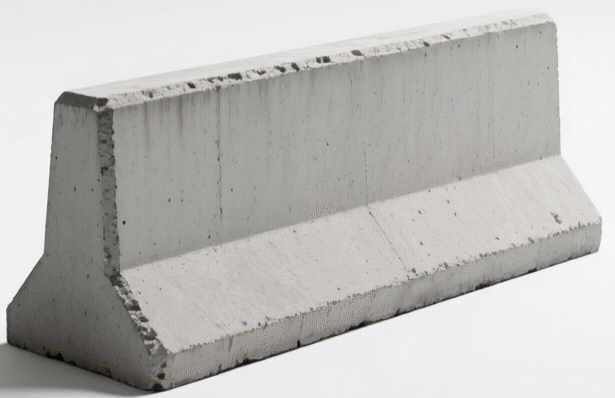} 
    & \cellimgg{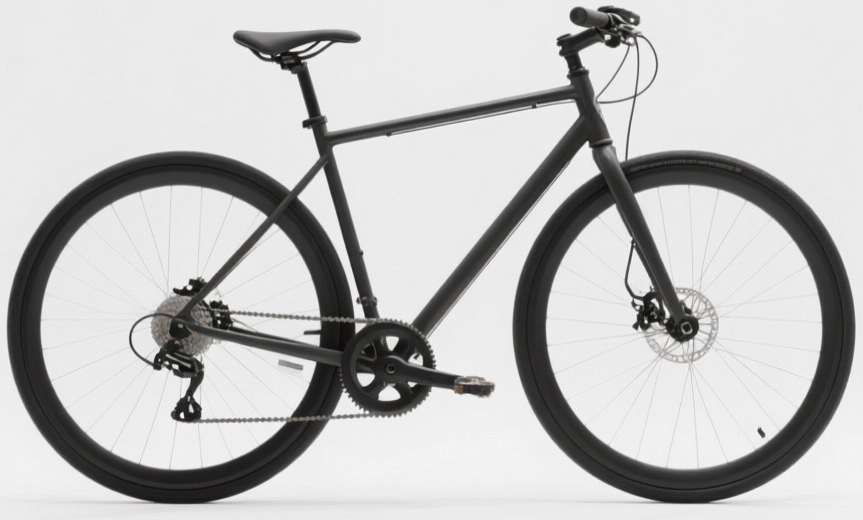} 
    & \cellimgg{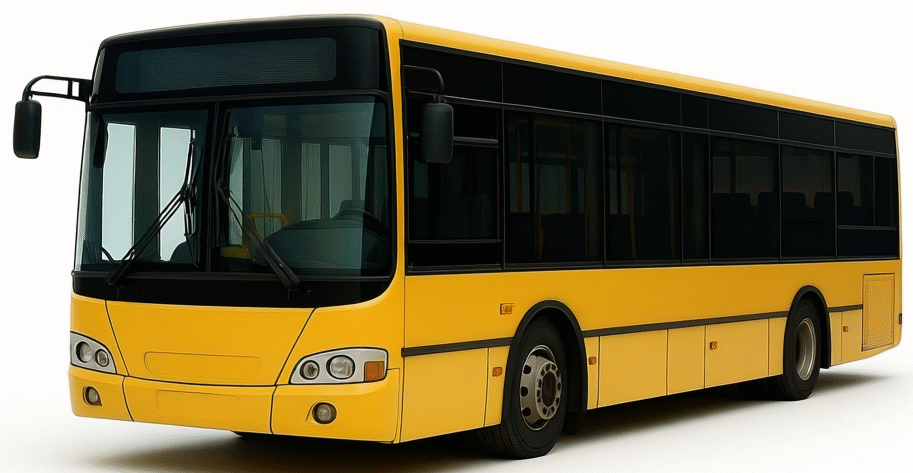} 
    & \cellimgg{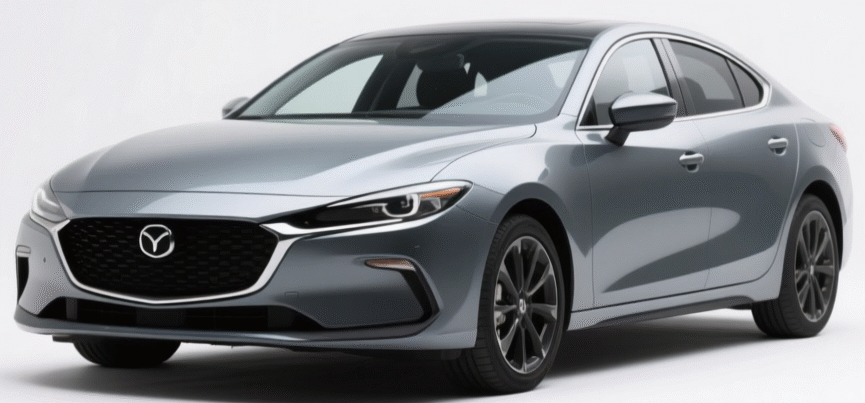} 
    & \cellimgg{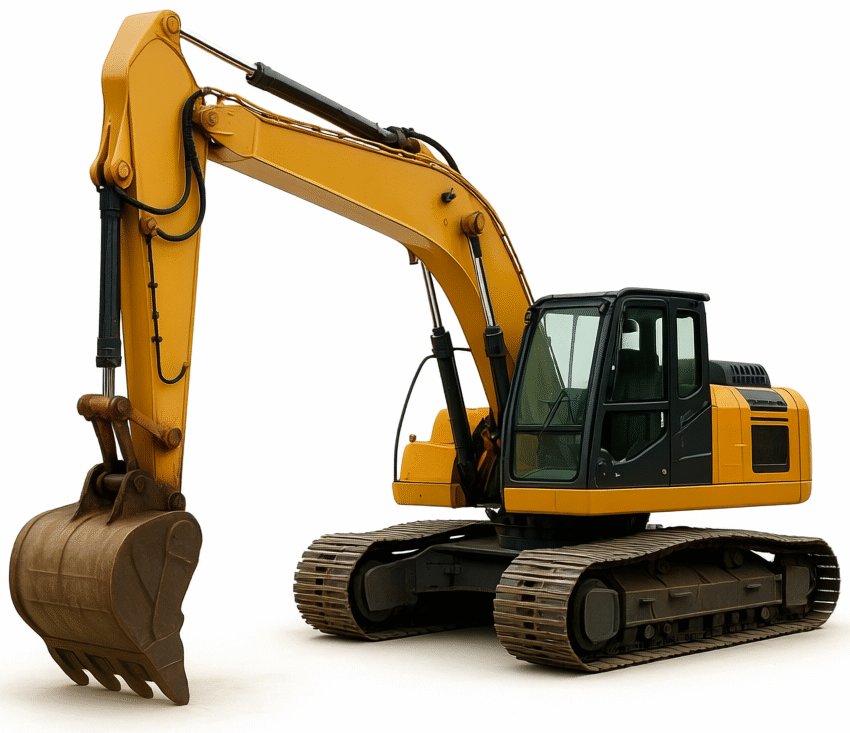} 
    & \cellimgg{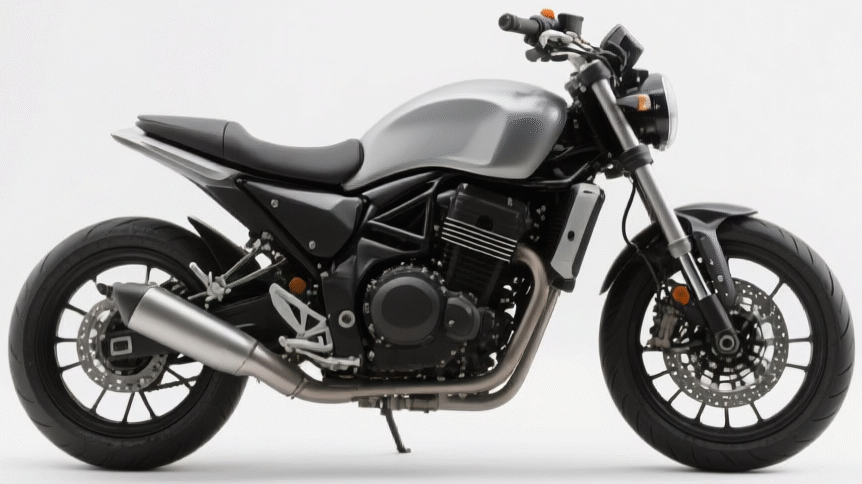} 
    & \cellimgg{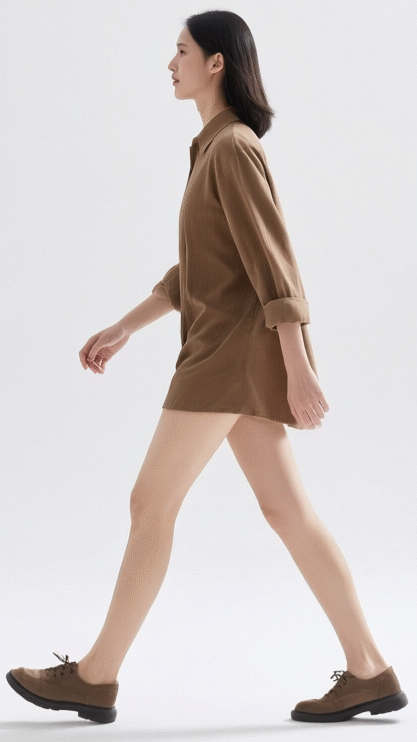} 
    & \cellimgg{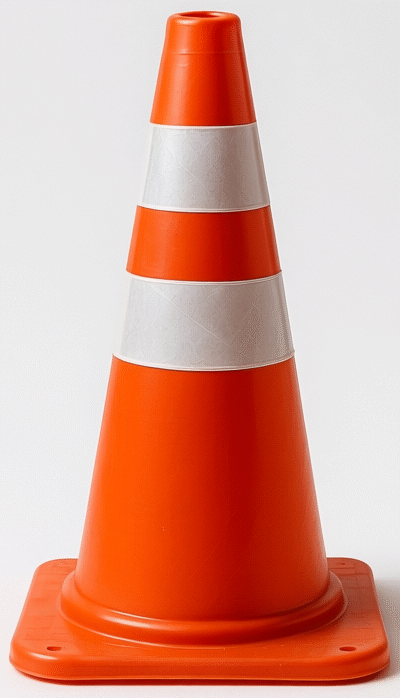} 
    & \cellimgg{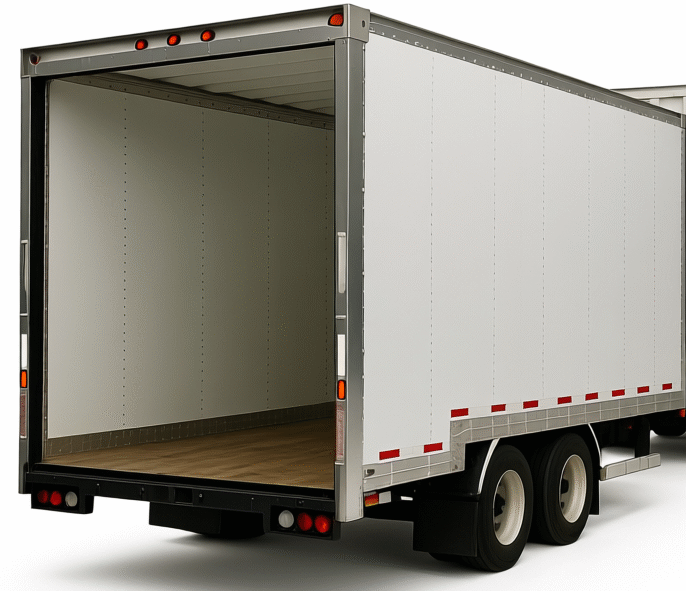} 
    & \cellimgg{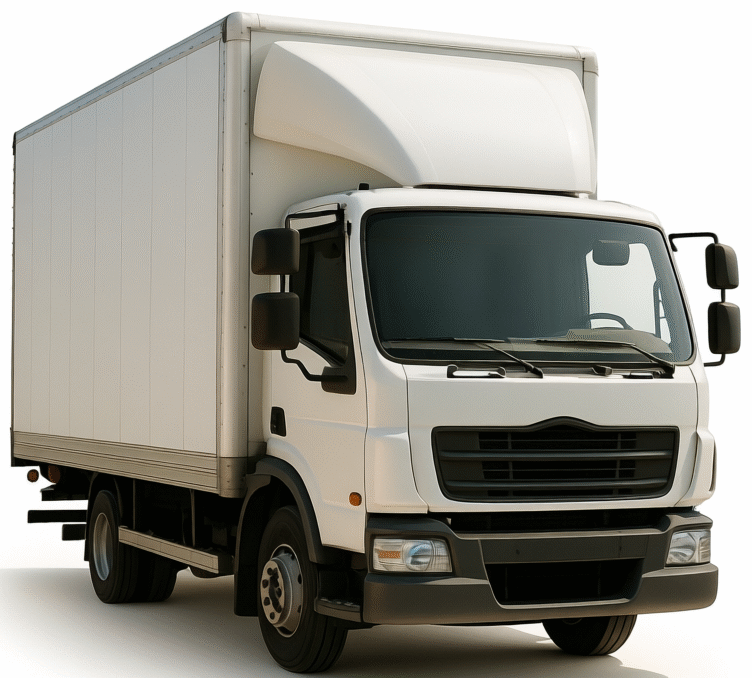} 
    & \cellimgg{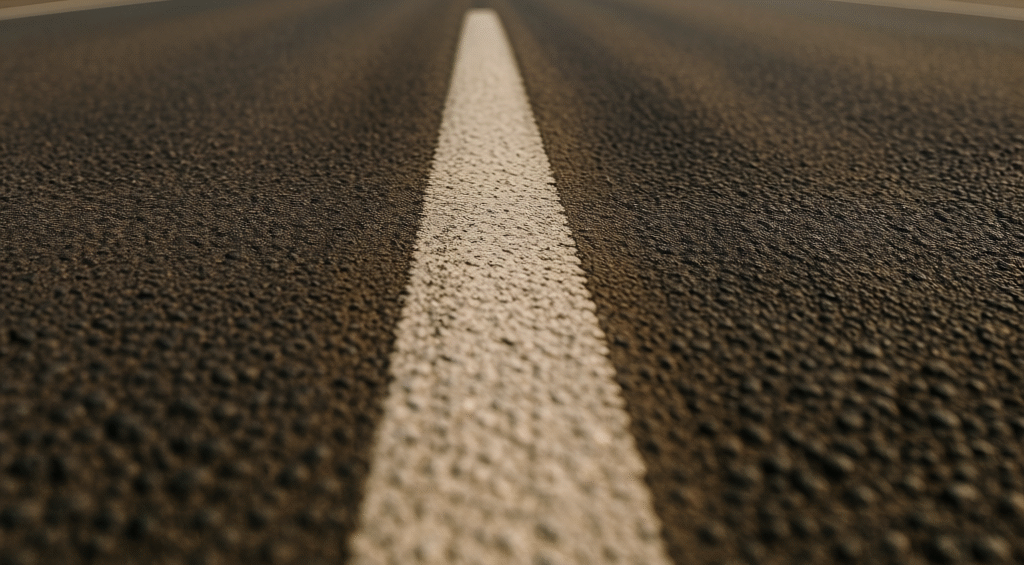} 
    & \cellimgg{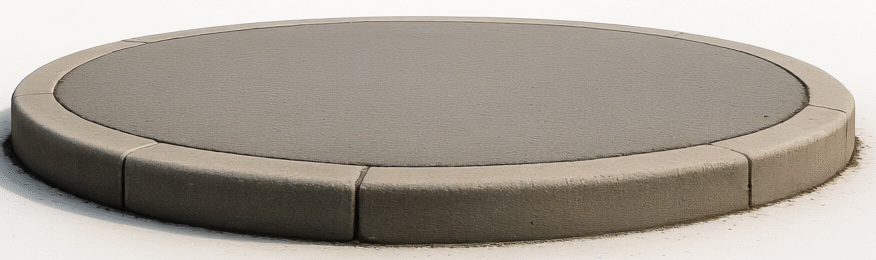} 
    & \cellimgg{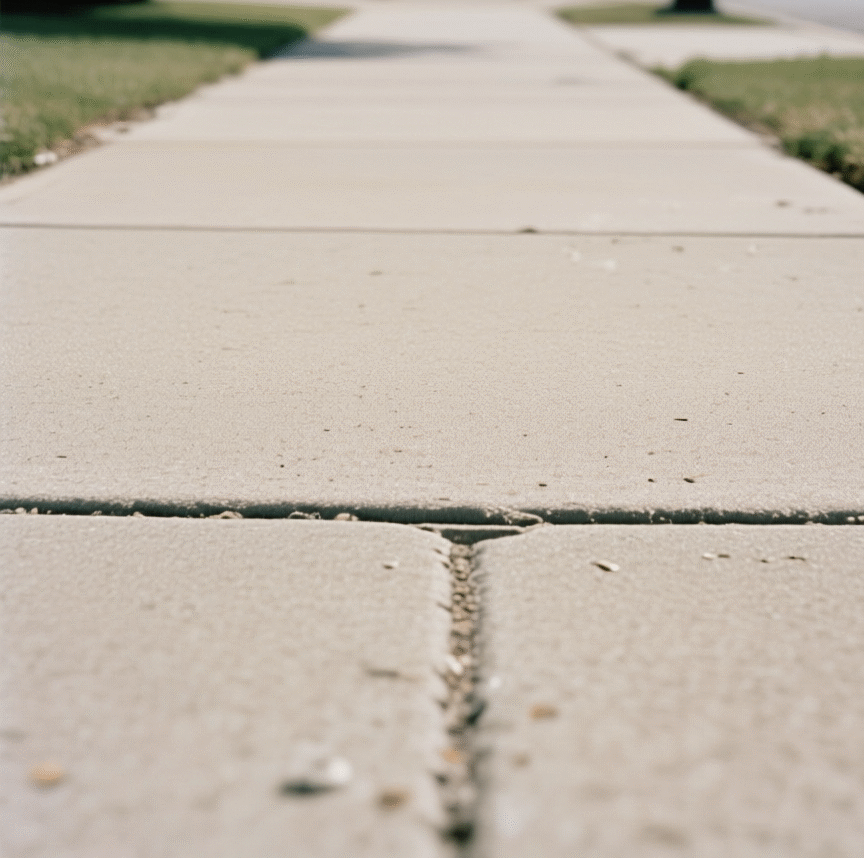} 
    & \cellimgg{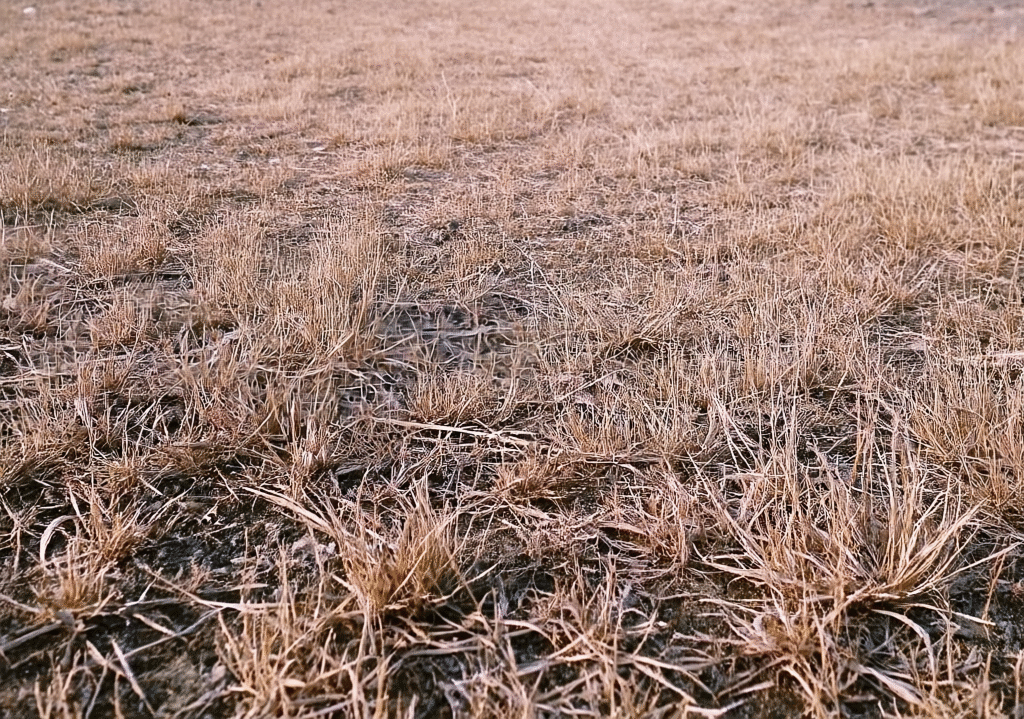} 
    & \cellimgg{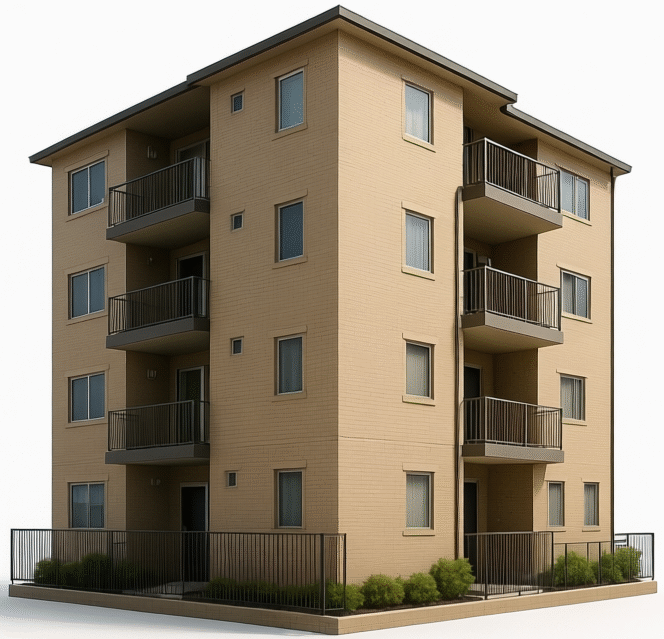} 
    & \cellimgg{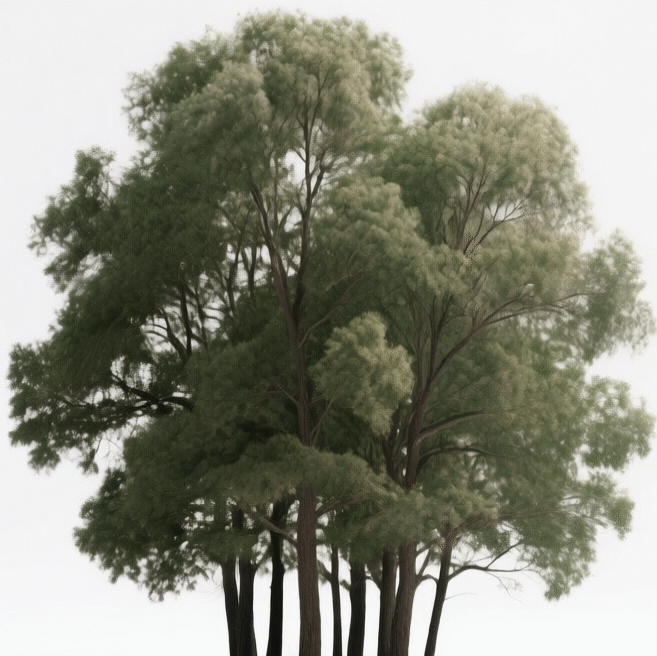}
       
     \\ 
    \cellimgg{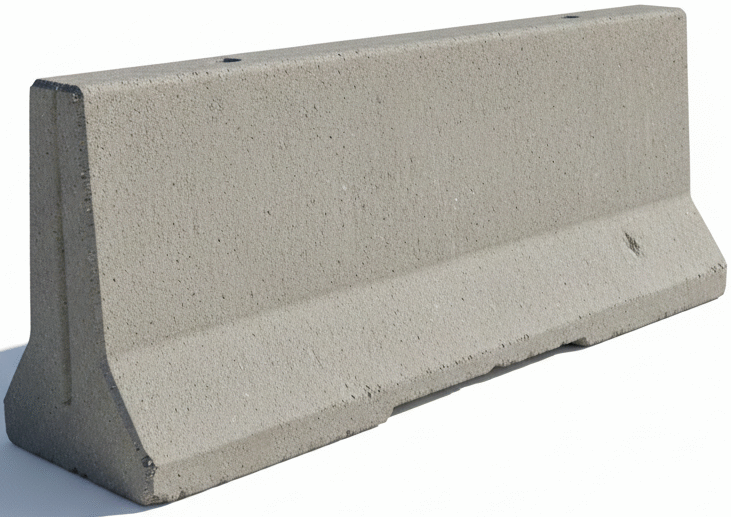} 
    & \cellimgg{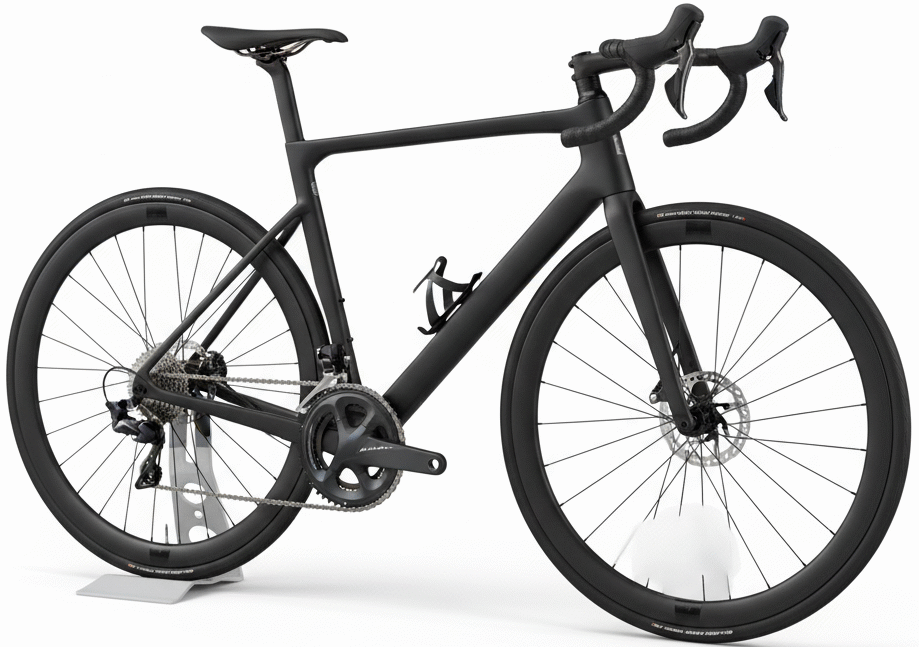} 
    & \cellimgg{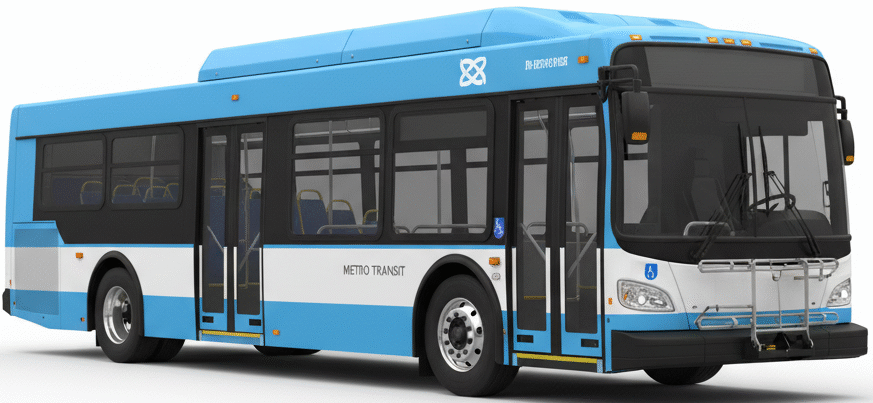} 
    & \cellimgg{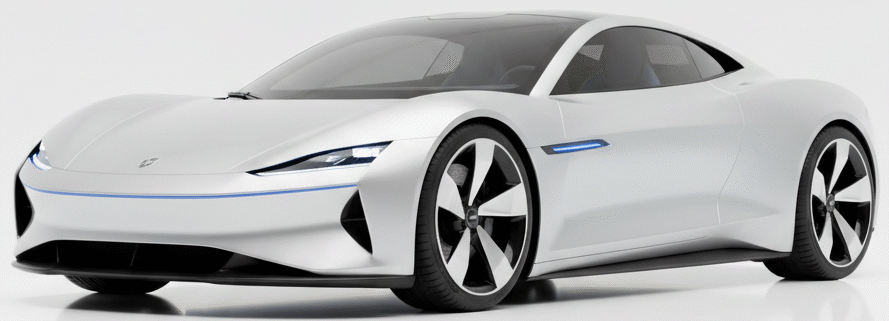} 
    & \cellimgg{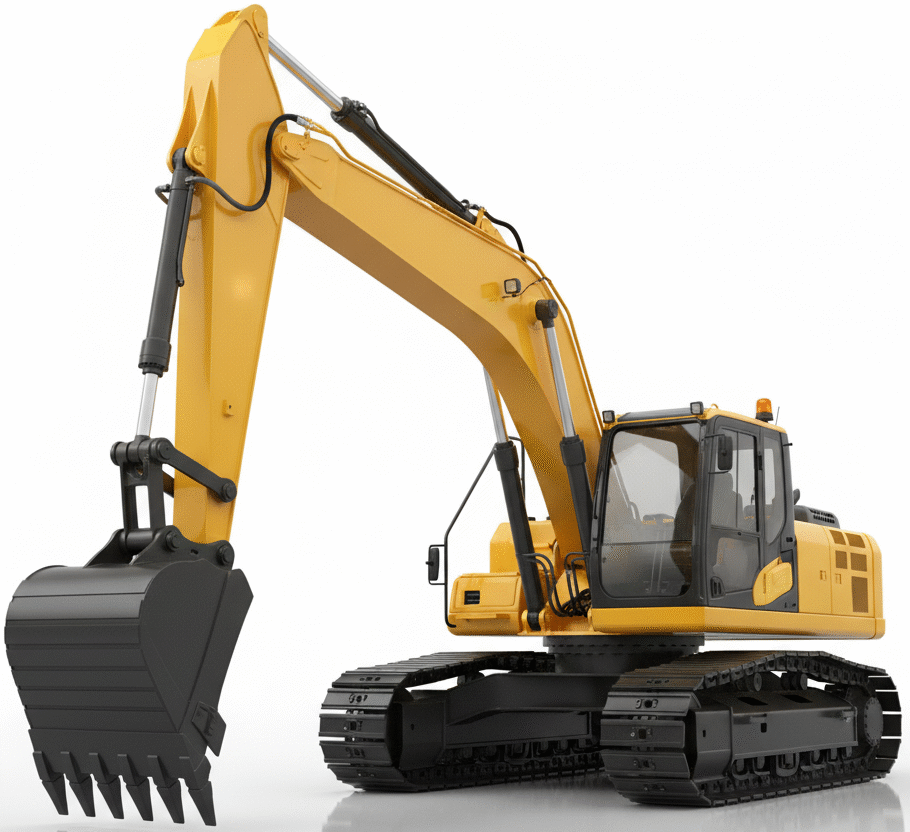} 
    & \cellimgg{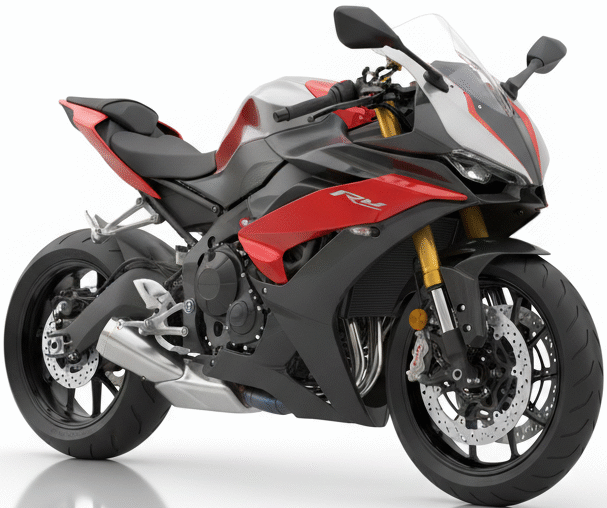} 
    & \cellimgg{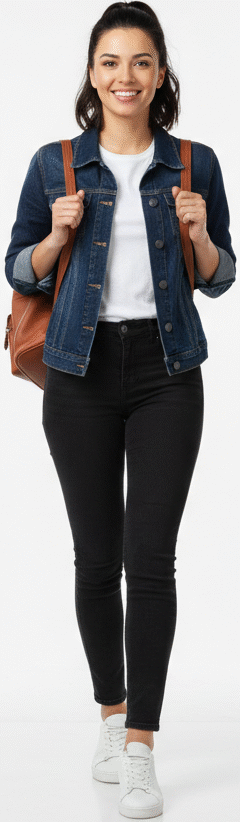} 
    & \cellimgg{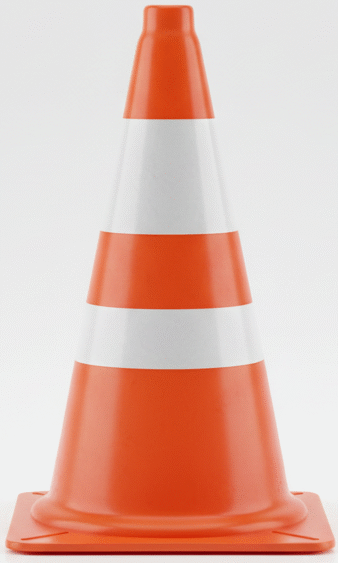} 
    & \cellimgg{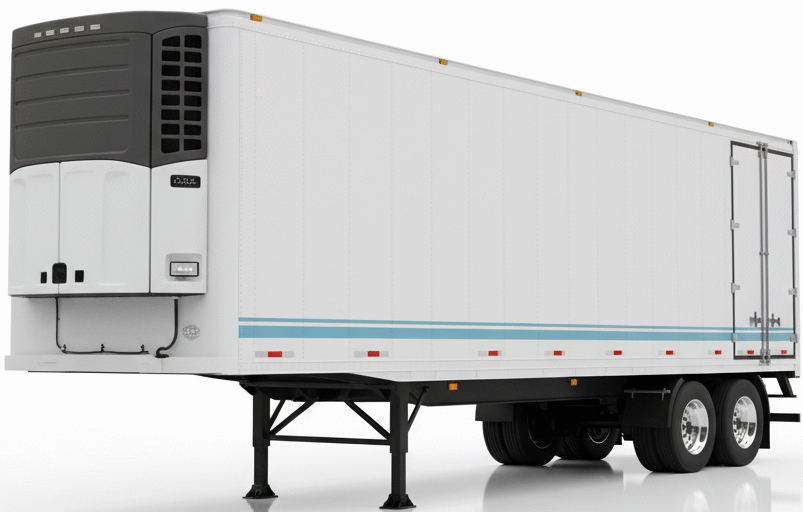} 
    & \cellimgg{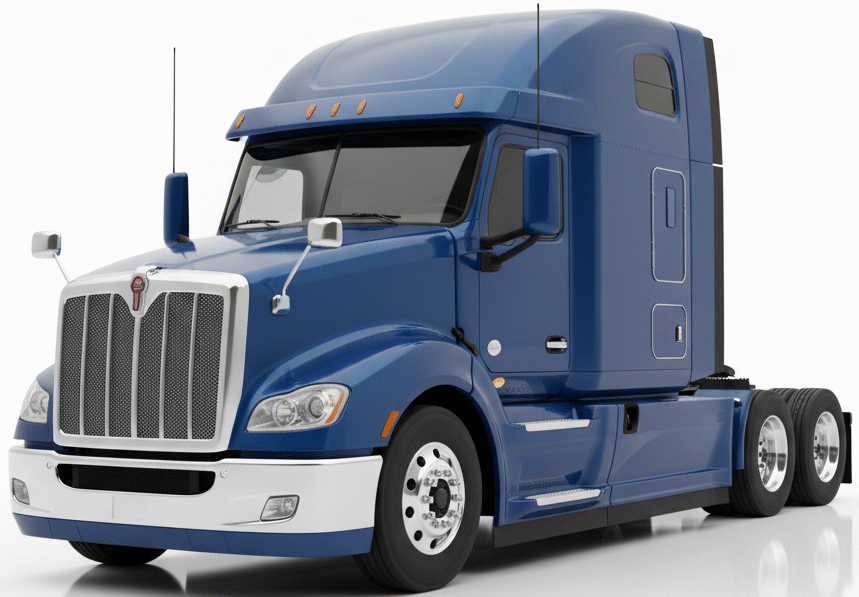} 
    & \cellimgg{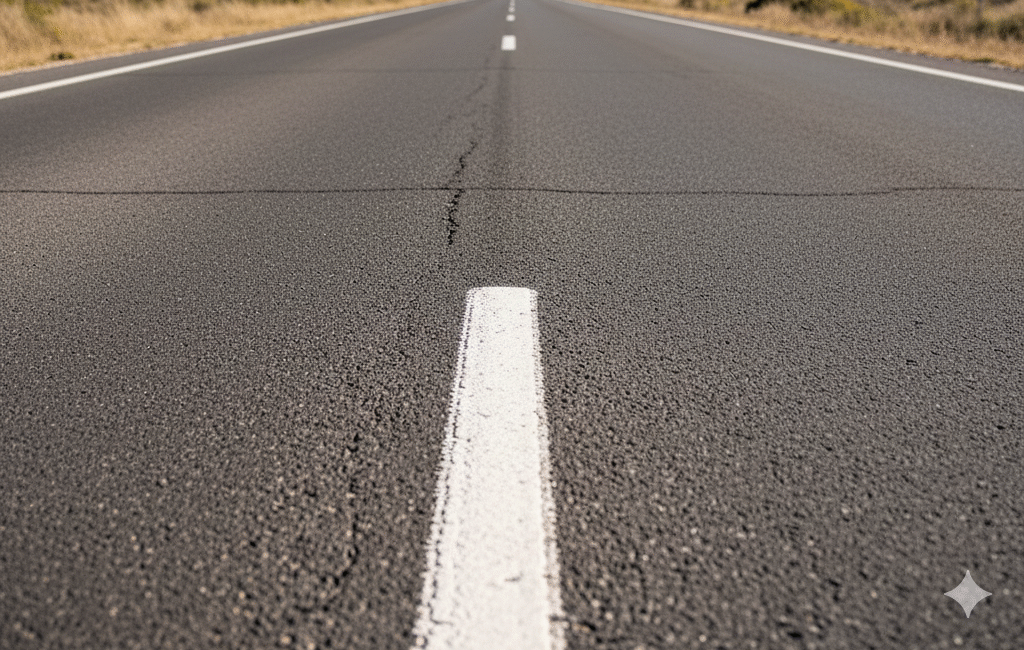} 
    & \cellimgg{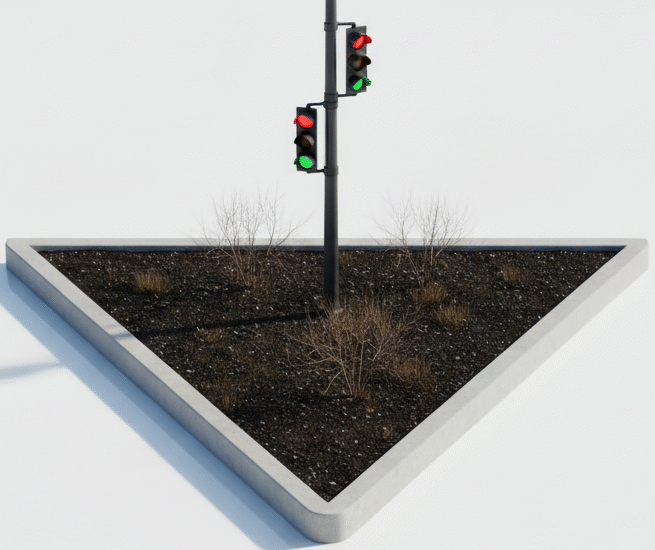} 
    & \cellimgg{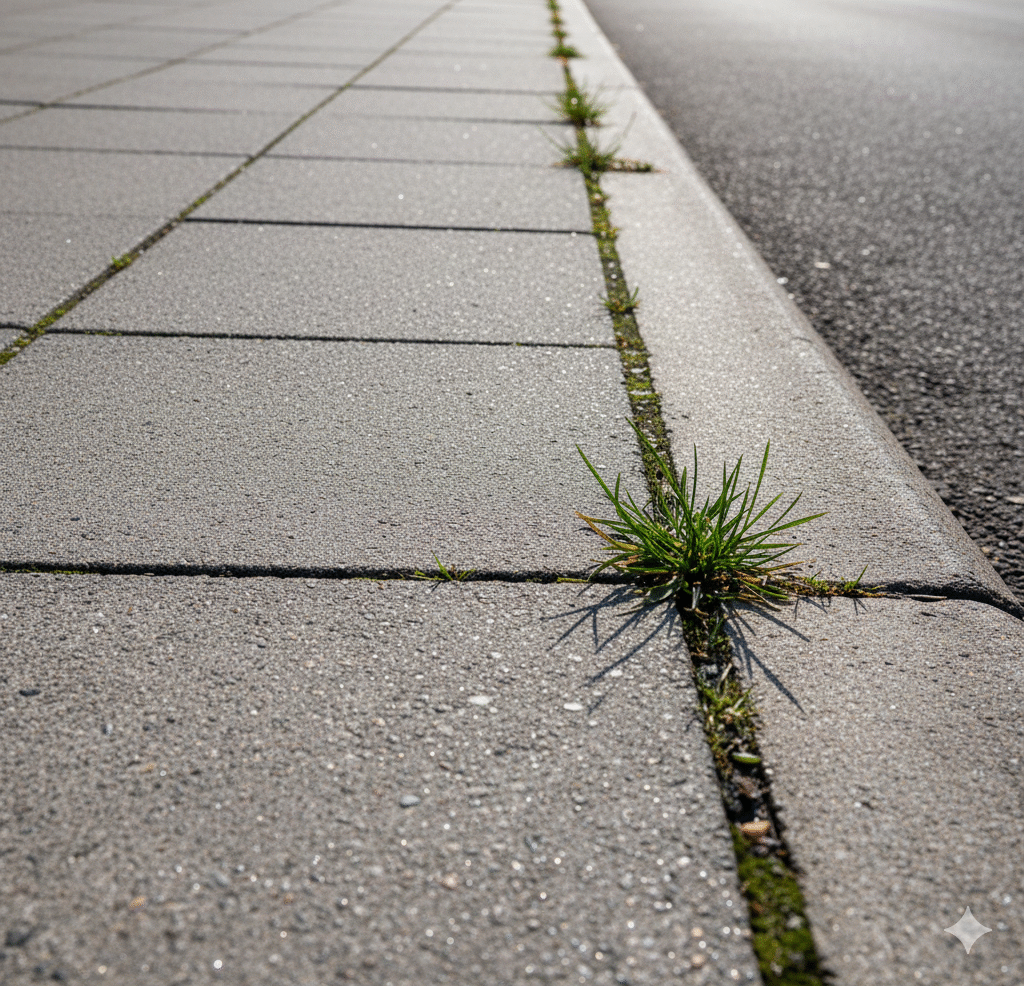} 
    & \cellimgg{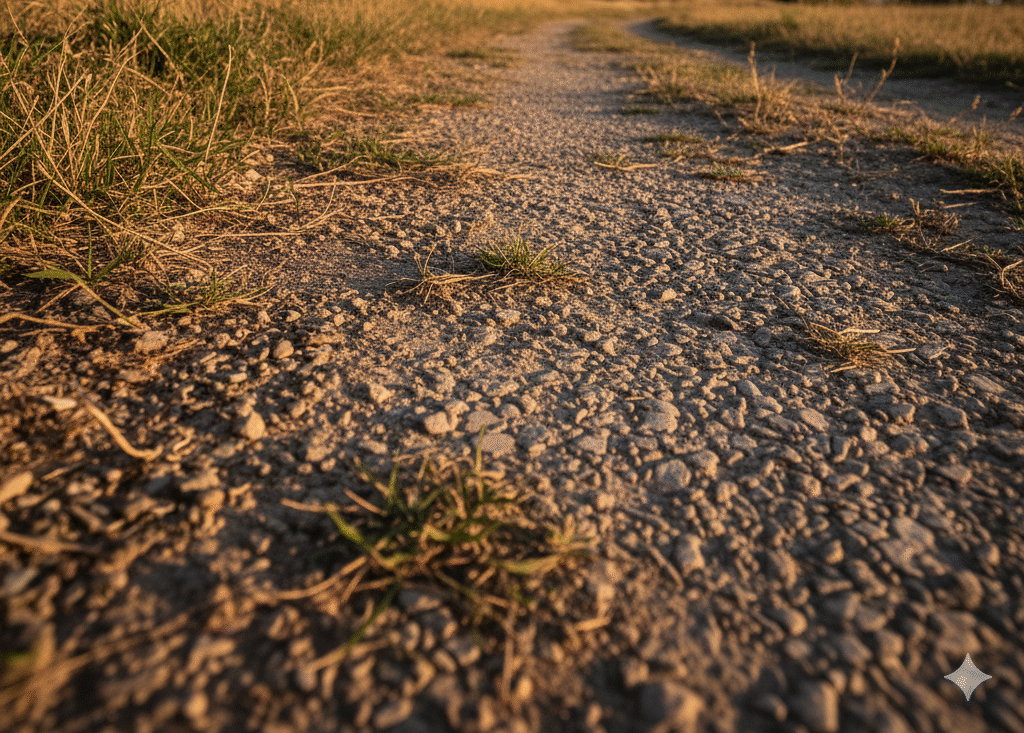} 
    & \cellimgg{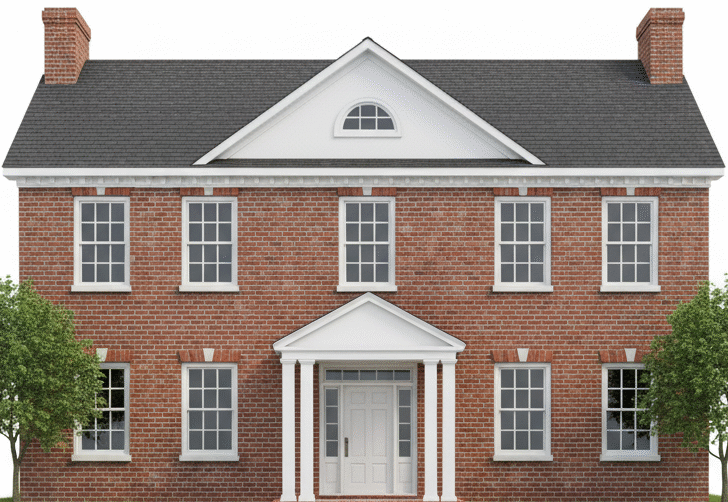} 
    & \cellimgg{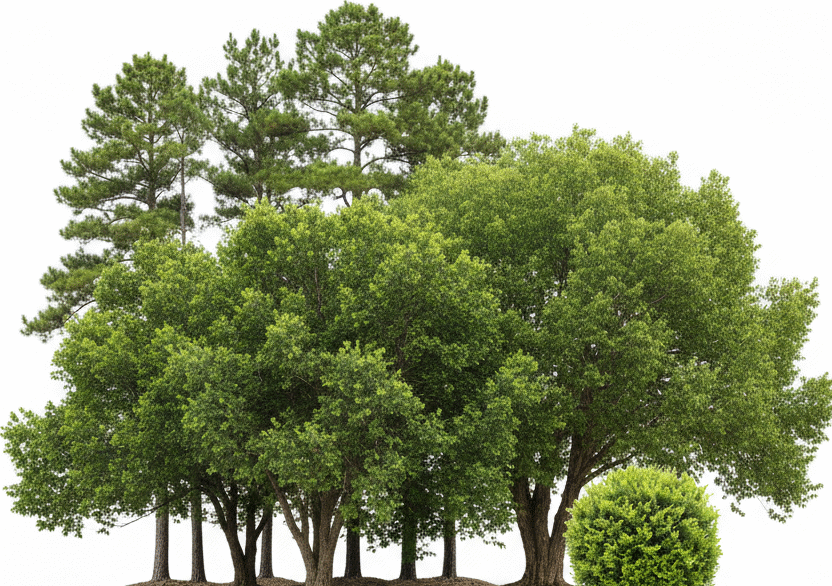} 
  \end{tabular}}
    \caption{\textbf{Examples of generated images} (see \cref{sec:experiments}) with ChatGPT (top), Qwen Image (middle) and Gemini (bottom). Web images are not shown due to copyright restrictions.}
    \label{fig:templates}
    \vspace{-3mm}
\end{figure}

\simpletextparagraph{Focus on objects and stuff.}
When asked to generate an image picturing an object, image generators tend to include a background, which may confuse the extraction of object-specific features. As a workaround, OVDiff \cite{ovdiff} uses CutLER \cite{cutler} to discard the background.
We prefer to ask the IG to generate, for \emph{things}, the object in a white background, and, for \emph{stuff}, some stuff covering the whole image. 
For each subclass $c_{i,j}$ of the \emph{thing} (resp.\ \emph{stuff}) type, we create a text prompt $t_{i,j}$ of the form ``\relax{Generate an image of $c_{i,j}$ \relax{with a white background}}'' (resp.\ ``\relax{covering the whole image}'').

After we generate a prototype image $I_{i,j}$ for a subclass $c_{i,j}$, 
\nermin{to obtain the corresponding prototype feature $f^{\text{2D}}_{i,j}$, we first normalize the individual patch features extracted from the encoded prototype image $I_{i,j}$; we then compute their average and normalize the resulting feature vector.}

In driving \renaud{settings}, % datasets, 
cameras and lidars have \renaud{close} % similar 
viewpoints, thus ``see'' similar occlusions. When distilling, % During distillation, 
the 3D network is trained to produce point features similar to 2D patch features, regardless of occlusions. As patch
features tend to be similar within an object, \renaud{we take} % it justifies taking 
the average as prototype.

\subsection{Point cloud OV semantic segmentation}

Finally, given a scan, we obtain normalized 3D features $f^{\text{3D}}_p$ of any point $p$ using the 3D VFM, and compare them to the prototype features $f^{\text{2D}}_{i,j}$.

When the task is to segment one object class on an unknown background, a basic labeling strategy is to set a threshold on $f^{\text{3D}}_p \cdot f^{\text{2D}}_{i,j}$. But it is hard to set, class sensitive, and it can generate overlapping segments.
When there are at least two classes, the usual labeling strategy (\cref{tab:relwork}) is to associate with what has the highest cosine similarity. This nearest neighbor (NN) strategy assigns to point $p$ the subclass $c_p = \NN(f^{\text{3D}}_p) = \argmax_{f^{\text{2D}}_{i,j}} (f^{\text{3D}}_p \cdot f^{\text{2D}}_{i,j})$.

We \nermin{also} consider a labeling strategy that consists in fitting a multinomial logistic regression (LR) on features $f^{\text{2D}}_{i,j}$. It is done after the images and 2D features are generated, and produces a shallow classifier $\LR$. 
Moreover, if new prototypes are added, the classifier can be rapidly refitted. 
\nermin{At test time,} a scan point $p$ is then classified as $c_p = \LR(f^{\text{3D}}_p)$, based on the maximum logit index.

LR was already found to outperform NN in few-shot classification \cite{tian2020rethining}. LR adds a small initial overhead ($\sim$\,70\,ms) but becomes faster than NN when the same query is made on several point clouds (e.g., for auto-labeling a dataset) or involves several prototypes per (sub)classes. We thus argue that LR does not break the OV paradigm, which does not have to be purely training-free. In fact, adding an online query preprocessing has already been used in OV methods, as test-time adaptation (TTA) \cite{karmanov2024efficienttta,noori2025testtimeadaptation}, test-time training (TTT) \cite{zhou2026zeroshotovhumanmotion} or test-time prompt tuning (TPT) \cite{shu2022tpt, feng2023difftpt}.

\subsection{\ours variant 1: closed-set self-training}
\label{sec:selftraining}

\nermin{Our method, \ours, is open-vocabulary at inference time. 
It can segment arbitrary classes given by free text prompts without retraining a 3D model, e.g., for an interactive retrieval.}
But when the classes are fixed, e.g., for auto-labeling, it is possible, after label assignment, to complement \ours with a self-training phase \cite{zou2018selftraining} to improve the label predictions. For this \emph{closed-set self-training}, denoted \oursclosed, we first improve the spatial and temporal consistency by aggregating each sequence of labeled scans into the same coordinate system and enforcing label consistency by majority voting of the points falling into a voxel \cite{samet2026losc}. We then propagate voxel classes back to points of individual scans, and consider them as pseudo-labels to finetune the distilled 3D backbone of the 3D VFM (cf.\ \cref{sec:2d3dvfmdistill}).

\subsection{\ours variant 2: 2D-3D VLM mixing}
\label{sec:ensembling}

If several 2D-3D distillations are available and complementary with respect to their features, it is profitable to combine them. In particular, it is possible to make a single 3D distillation of concatenated 2D features, but the resulting high-dimensional mixed features can be difficult for a 3D network to regress. Alternatively, we just concatenate the feature vectors of the 2D VLMS to get images features, and of the corresponding 3D  distillations to get point features.

\section{Experiments}
\label{sec:experiments}

We now provide implementation details (\cref{sec:implementation}), quantitative results for both open-set open-vocabulary and closed-set annotation-free settings (\cref{sec:openvoc}), an analysis of the image-text modality gap (\cref{sec:modgap}), and ablation studies (\cref{sec:ablation}).

\subsection{Implementation details}
\label{sec:implementation}

\hspace{\parindent}\simpletextparagraph{Lidar datasets.}
We conduct our experiments  on classical automotive datasets, \ns (NS) \cite{nuscenes} and \sk (SK) \cite{semantickitti}, with identical parameters. 

\simpletextparagraph{Aligned 2D-3D VFMs.}
We use a \scalrplus (\cref{sec:2d3dvfmdistill}) with a multi-dataset \cite{nuscenes, semantickitti, xiao2021pandaset} distillation of DINOv2 \cite{oquab2024dinov2}. A single model can thus be shared across NS and SK, in contrast to prior work that trains separate models for each dataset.

\simpletextparagraph{Image generation.}
We generate or get prototype images from four sources: ChatGPT-5~\cite{chatgpt5} (Plus), Qwen-Image 20B~\cite{qwen}, Gemini 2.5~\cite{gemini} (Pro)
and Google Image Search. For Web-based collection, we query images with transparent backgrounds to be consistent with generated images (\cref{sec:protogen}). 
\cref{fig:templates} shows image examples % of images produced by ChatGPT, Gemini and FLUX 
for nuScenes classes. For SemanticKITTI, we use the same set of images as for nuScenes, generating additional images only for classes that are not in nuScenes; specifically, we add images for \textit{motorcyclist}, \textit{bicyclist}, and \textit{parking}.

\simpletextparagraph{Feature extraction and classification.}
We resize prototype images to 448\,$\times$\,448 and extract DINOv2 features at the layer used for 2D-3D distillation. Patch features are averaged into a single 1024-d feature vector per prototype.

Logistic regression uses liblinear solver \cite{liblinear} with default regularization $C\,{=}\,1$.

\simpletextparagraph{\ours variants.} 
Unless otherwise stated, \ours is our method with a \scalrplus distillation of DINOv2 in WaffleIron-48-768 \cite{waffleiron}, two ChatGPT images per subclass prototype (prompt), and classification with logistic regression (LR). % 

We denote by \oursclosed the closed-set model after self-training with labels given by \ours (cf. \cref{sec:selftraining}). Voxel-based point consistency operates with voxels of 10\,cm. We then finetune \scalrplus with these pseudo-labels as done in ScaLR \cite{scalr}, for 10 epochs, with a learning rate of 10\textsuperscript{-4} and a layer-wise learning rate decay of 0.99. While \ours is OV at inference, \oursclosed is not.

We call \oursens our method applied to a mix (\cref{sec:ensembling}) of \scalrplus distilled from DINOv2 and OpenScene distilled from OpenSeg \cite{peng2023openscene}. It only applies to NS as no OpenScene model is available for SK.
\oursclosedens first does the mixing, then self-training.

\simpletextparagraph{Runtime.}
Generating each prototype image takes approximately 3--5 s, while extracting all prototype features and fitting the LR classifier require only 80 ms and 70 ms, respectively, on a single A100 GPU. These steps are performed once for a given vocabulary, after which the resulting prototypes and classifier can be reused across all scans. Consequently, the only per-scan cost is point-cloud inference, which takes approximately 200 ms.

\begin{table}[t]
\tabcolsep 3pt
\caption{\textbf{Test-time and train-time OVSS results} (mIoU\%) on the  \ns (NS) and \sk (SK) validation sets. We denote {\ours} = {\ours}({\scalrplus}(DINOv2)), {\oursens} = {\ours}({\scalrplus}(DINOv2)\,\conc\,OpenScene), OpenScene is distilled from OpenSeg. 
\ul{\emph{OV:}} \xmark~= not OV at test time; \mmmark~= bias at training time with a predefined vocabulary or text prompts, but functionally OV; \mmark~= bias at training time from a captioner, but still OV; \cmark~= fully OV. \ul{\emph{3D:}} \cmark~= input is only 3D points; \xmark~= input includes images in addition to 3D points. \ul{\emph{ens:}} \xmark~= no ensembling; \cmark~= VFM ensembling (at train or test time). \renaud{``-'' = no such experiment reported.}
}%
\label{table:sota_semantic_ovss}%
\vspace{1mm}%
\noindent
\begin{minipage}[t]{0.532\textwidth}
\resizebox{0.95\textwidth}{!}{ 
\begin{tabular}[t]{l@{}r|ccc|c|c}
\multicolumn{7}{c}{\!\!\!(a) Open-vocabulary semantic segmentation\!\!\!}
\\[2mm]
\toprule
\multicolumn{2}{l|}{Method}
& OV
& 3D
& \!ens
    & NS
    & SK
    % & Code
    % & Backbone
\\
\midrule
\rowcolor{black!15} 
MaskCLIP$\rightarrow$3D & \cite{zou2025adaco} % \!\cite{zhou2022maskclip}
& \cmark & \xmark & \xmark
    & 16.6 
    & 8.1
    % & \cmark
    % & \na
\\
\rowcolor{black!15} 
3D-AVS {\scriptsize (I+L)} & \cite{wei20253davs}  
& \mmark & \xmark & \xmark
    & 36.2 
    & -
\\
\rowcolor{black!15} 
OpenScene {\scriptsize (I+L)} & \cite{peng2023openscene}
& \cmark & \xmark & \xmark
    & 42.1 
    & -
\\
\midrule
CLIP2Scene & \cite{clip2scene}
& \mmmark   & \cmark & \xmark
    & 20.8
    & -
\\
3D-AVS {\scriptsize (L)} & \cite{wei20253davs} 
& \mmark & \cmark & \xmark
    & 33.4 
    & -
\\
SAL & \cite{osep2024sal}
& \cmark & \cmark & \xmark
    & 33.9 
    & 28.7
\\
OpenScene {\scriptsize (L)} & \cite{jiang2024ov3d}
& \cmark & \cmark & \xmark
    & 41.3 
    & -
\\
OV3D & \cite{jiang2024ov3d} 
& \mmark & \cmark & \xmark
    & 44.6
    & - 
\\
GGSD & \cite{wang2024ggsd}
& \mmmark   & \cmark & \xmark
    & 46.1
    & -
\\
{\ours} & \llap{(ours)}
& \cmark & \cmark & \xmark
    & \textbf{47.5}
    & \textbf{34.3}
\\
\midrule
\rowcolor{blue!10}
OV3D\textsubscript{w/\,OpenScene~} & \cite{jiang2024ov3d}  
& \mmark & \cmark & \cmark
    & 45.5 
    & -%
    \rlap{\quad: ensembles SEEM \cite{zou2023seem} and OpenSeg \cite{openseg}.}
\\    
\rowcolor{blue!10}
SAS & \cite{li2025sas}
& \mmmark   & \cmark & \cmark
    & 47.5
    & -%
    \rlap{\quad: ensembles SEEM \cite{zou2023seem} and LSeg \cite{li2022lseg}.}
\\
\rowcolor{blue!10}
{\oursens} & \llap{(ours)}
& \cmark & \cmark & \cmark
    & \textbf{51.1}
    & -%
    \rlap{\quad: ensembles {\scalrplus}(DINOv2) and OpenScene.}
\\
\bottomrule 
\end{tabular}
}
\end{minipage}%
\begin{minipage}[t]{0.468\textwidth}
\resizebox{0.96\textwidth}{!}{
\begin{tabular}[t]{l@{~}r|cc|c|c}%|c|l}
\multicolumn{6}{c}{\!\!\!(b) Annotation-free closed-set segmentation\!\!\!}
\\[2mm]
\toprule
\multicolumn{2}{l|}{Method}
    & OV
    & 3D
    & {NS}
    & {SK}
\\
\midrule
\rowcolor{black!15} 
AFOV {\scriptsize (I+L)} & \cite{sun2025afov}
    & \xmark 
    & \xmark 
    & 47.9 
    & -
\\
\midrule
HICL & \cite{kang2024hicl}  
    & \xmark 
    & \cmark & 23.0 & - 
\\
CNS & \cite{chen2023cns}
    & \xmark 
    & \cmark & 26.8
    & -
\\  

AdaCo & \cite{zou2025adaco}
    & \xmark 
    & \cmark & 31.2
    & 25.7
\\
AFOV {\scriptsize (L)} & \cite{sun2025afov}
    & \xmark 
    & \cmark 
    & 47.7 
    & -
\\

LOSC  & \cite{samet2026losc}
    & \xmark 
    & \cmark 
    & 49.3
    & 35.2
\\

{\oursclosed} 
    & \llap{(ours)}
    & \xmark 
    & \cmark 
    & \textbf{49.6}
    & \textbf{39.4}

\\
\midrule
\rowcolor{blue!10}
{\oursclosedens}~~  
    & \llap{(ours)}
    & \xmark 
    & \cmark 
    & \textbf{54.1}
    & -
\\
\bottomrule
\end{tabular}
}
\end{minipage}

\end{table}

\subsection{Open-vocabulary and annotation-free semantic segmentation}
\label{sec:openvoc}

\simpletextparagraph{Open-set OVSS.}
\cref{table:sota_semantic_ovss}(a) compares test-time OV methods.
\ours sets a new SOTA among methods without VFM ensembling, gaining +1.4 mIoU pts on nuScenes. It gains +5.6 pts on SemanticKITTI, which is however much less evaluated by other methods, maybe because it is harder for them to adapt to a single camera viewing only a fraction of the 360\textdegree\ lidar scan, yet to fully segment.
{\oursens} is also SOTA among methods with train-time or test-time ensembling of VFMs, improving mIoU by +3.6 pts on nuScenes.

Although IGLOSS benefits from a strong backbones and pretraining, these gains nevertheless highlight the effectiveness of relying solely on image features, avoiding the text–image alignment used by other methods.
Meanwhile, \ours is considerably simpler than most of the other methods.

\simpletextparagraph{Closed-set annotation-free semantic segmentation.}
\cref{table:sota_semantic_ovss}(b) compares methods whose vocabulary is open only at training time.
\oursclosed surpasses the leading baseline LOSC on both \ns and \sk. Furthermore, \oursclosedens improves on \ns by an additional 4.5 points.
Compared to the open-set (OV) setting of \ours, the closed-set self-training of \oursclosed (\cref{sec:selftraining}) further improves semantic segmentation by +2.1 mIoU on NS and +5.1 mIoU on SK.

We note that \sk is rarely evaluated in this literature, possibly because its single front camera supervises only a fraction of the 360$^\circ$ LiDAR scan. Moreover, many prior methods do not publicly release their code, limiting additional evaluation on \sk. We therefore report all directly available \sk results we could identify.

\subsection{\renaud{Analyzing the image–text modality gap} at 2D level}
\label{sec:modgap}

\renaud{To support our choice of doing the alignment only in visual space rather than in a joint image-text space,} we % first
evaluate \renaud{at 2D level, without 2D-3D distillation,} how well % both
text and image prototype representations align with real object appearances in the CLIP embedding space. We use Cityscapes \cite{cityscapes} as an independent pure 2D benchmark, % which was 
not used for training or evaluation in our method. 

We first identify the % ten 
classes shared between Cityscapes and nuScenes, matched either by identical class names (bicycle, bus, car, motorcycle, truck, sidewalk, terrain, vegetation) or by clear semantic correspondence (pedestrian $\leftrightarrow$ person, driveable\_surface $\leftrightarrow$ road).
For each Cityscapes validation image, we then extract object and background class crops using the ground-truth semantic masks and encode them with the image encoder of OpenCLIP ViT-L/14 \cite{openclip}. Last, we compare these embeddings against two different types of class prototypes:

\textit{CLIP text embeddings} are generated from the same prompts used to create the IGLOSS prototype images, after removing modifiers specific to image generation (e.g., "with white background" or "covering the whole image").

\textit{CLIP image embeddings} are extracted from the generated prototype images used in IGLOSS.
\renaud{For classes with multiple prototypes,
% When multiple prototype images existed for a class, 
we adopt nearest-template matching by assigning each crop the maximum cosine similarity across \renaud{the templates of that class} % that class's templates 
before averaging over all instances.
}

\begin{table}[t]
\centering
\caption{\renaud{\textbf{CLIP alignment of text or generated-image prototypes with Cityscapes.} We measure the cosine similarity of CLIP embeddings of class prototypes with image masks in Cityscapes val.}}
\label{tab:clip_exp}
\vspace*{5pt}
\resizebox{\textwidth}{!}{%
\begin{tabular}{l c ccccccccccc}
\toprule
\textbf{Prototype} &  \textbf{bus} & \textbf{truck} & \textbf{car} & \textbf{motorcycle} & \textbf{bicycle} & \textbf{road} & \textbf{pedestrian} & \textbf{terrain} & \textbf{sidewalk} & \textbf{vegetation} &  \textbf{mIoU} (\%) \\
\midrule
Text % sim  
& 27.7 & 26.2 & 26.1 & 25.2 & 24.9 & 24.4 & 23.9 & 23.4 & 23.1 & 22.0 & 24.7 \\
Image % sim  
& \textbf{55.7} & \textbf{47.2} & \textbf{38.2} & \textbf{31.2} & \textbf{26.4} & \textbf{43.4} & \textbf{28.7} & \textbf{45.5} & \textbf{52.7} & \textbf{47.1} & \textbf{41.6} \\
\bottomrule
\end{tabular}%
}
\vspace{-5pt}
\end{table}

We show classwise semantic segmentation results (IoU\%) in \cref{tab:clip_exp}. These results consistently show that generated \renaud{visual} % -image 
prototypes exhibit higher similarity to real object instances than text embeddings, across all ten classes.

\subsection{Ablation, alternative and parameter studies}
\label{sec:ablation}

We present \renaud{here various} studies that justify our design choices. For efficiency, unless otherwise mentioned, we evaluate in simplified settings, using \textsl{only one image} per prompt and, sometimes, a reduced nuScenes dataset. This explains the differences with the ``final'' results reported in \cref{table:sota_semantic_ovss}.

\begin{table}[t]
\small
\caption{\textbf{2D-3D VFM evaluation.} We generate \textsl{one image per prompt}  with ChatGPT, except for OVDiff that generates 32 images using Stable Diffusion \cite{rombach2022sd}. (a)~We match 2D VFM features of prototypes and of nuScenes val set images, with nearest neighbor (NN), unproject classes into seen 3D points, and measure mIoU\%. $^{\smash\ast}$: OVDiff-like evaluation, i.e., with all foreground classes and one merged background class. (b)~We match 2D VFM prototype features with point features of a distilled 3D VFM, using NN or LR, and measure mIoU\%. 
As OpenSeg, hence also OpenScene, have the same feature space as CLIP, we also simply experimented with CLIP visual features.
}
\label{tab:2d_3d_features_abl}
%\begin{minipage}[t]{0.532\textwidth}
\centering
\vspace{1mm}
\noindent\begin{tabular}{l@{~}r|c@{~}c}
\multicolumn{4}{c}{\small (a) 2D VFM evaluation}
\\[1mm]
\toprule
    \multicolumn{2}{l|}{2D VFM}
    &  NN 
    &  \!NN$^{\smash\ast}$\!
\\
\midrule    
OVDiff      & \cite{ovdiff}                 & \na   & 23.5  \\ 
SigLIP\,2    & \cite{tschannen2025siglip2}   & 18.7  & 23.7 \\ 
OpenSeg     & \cite{openseg}                & 19.8  & 23.7 \\ 
DiNOv2      & \cite{oquab2024dinov2}        & \textbf{23.3}  & \textbf{25.1} \\ 
\bottomrule
\end{tabular}%
%\end{minipage}
%\resizebox{1.0\linewidth}{!}{
\qquad
\qquad
\begin{tabular}{l|l@{~}r|c@{~}c}
\multicolumn{4}{c}{\small (b) 2D-3D VFM distillation evaluation}
\\[1mm]
\toprule
2D VFM &
3D VFM distillation & &
NN & LR \\
\midrule
OpenSeg   & OpenScene(OpenSeg) & \cite{openseg} & 35.5 &  29.8 \\
CLIP \cite{radford2021clip} & OpenScene(OpenSeg) & \cite{openseg} & 39.5 & 30.0  \\

DiNOv2      & ScaLR(DiNOv2) & \cite{scalr} & 34.8 & 39.7   \\
% \rowcolor{green!15} 
DiNOv2       & {\scalrplus}(DINOv2) & \llap{(ours)} & \textbf{42.9} & \textbf{46.6}   \\
\bottomrule
\end{tabular}
%}
\vspace{-5pt}
%\vspace{-3mm}
\end{table}

\begin{table}[t]
\small
\caption{\textbf{\scalrplus ablation study.} We study \scalrplus distillation of DINOv2 in WaffleIron \cite{waffleiron} (WI-48-256) on the \textsl{mini-train and mini-val sets of nuScenes} defined in \cite{slidr, scalr}, varying: the drop-path proportion (Drop), the type of non linearity (Non lin.), the backbone head (Head), the image resolution (Resol.) and the number of epochs (Epochs). We use a 2-layer MLP. Hidden dimension is 2048 as in \cite{cleverdistiller}. We measure the quality of features extracted before the head with linear probing (LP) and the impact on \ours 
when classifying with NN or LR.}  
\label{tab:scalrplusablat}
\vspace{1mm}
\centering
\begin{tabular}{l@{~}r|ccccc|ccc}
\toprule
\multicolumn{2}{l|}{Method} & Drop& \!Non\,lin. & Head & Resol. & \!\!\!\!Epochs\! & LP & NN & LR \\
\midrule
ScaLR & \cite{scalr}     & 0   & ReLU & lin. & 224\,$\times$\,448 & 25 & 64.3 & 37.1 & 40.1\\
           && 0.2 & ReLU & lin. & 224\,$\times$\,448 & 25 & 65.4 & 39.9 & 42.3\\
\multirow{2}{*}{\quad$\vdots$}
           && 0.2 & GELU & lin. & 224\,$\times$\,448 & 25 & 66.2 & 40.7 & 43.7\\
           && 0.2 & GELU & MLP  & 448\,$\times$\,896 & 25 & 67.9 & 43.1 & 45.7\\
           && 0.2 & GELU & MLP  & 448\,$\times$\,896 & 45 & 68.7 & 43.3 & 46.3\\
\scalrplus & (ours) & 0.2 & GELU & MLP  & 448\,$\times$\,896 & 65 & \textbf{68.7} & \textbf{43.7} & \textbf{46.5} \\
\bottomrule
\end{tabular}
\vspace{-2pt}
% \vspace{-3mm}
\end{table}

\simpletextparagraph{2D VFM \renaud{unprojection}.} % evaluation.}
We evaluate 2D VFMs for their ability to label lidar scans by unprojecting image labels onto 3D points. 
We compare to OVDiff \cite{ovdiff}, that also uses image generation, SigLIP~2 \cite{tschannen2025siglip2}, OpenSeg \cite{openseg}, that is the strongest image feature extractor used with OpenScene \cite{peng2023openscene}, and DINOv2 \cite{oquab2024dinov2}.  
Results in \cref{tab:2d_3d_features_abl}(a) show that DINOv2 yields the best performance. It confirms that current CLIP-like 2D VLM features are not as strong as pure 2D VFM features, likely because of \renaud{weaker dense (pixel-level) representations and because of the constraint} to also align with text features, which also limits training data to captioned images. In contrast, a 2D VFM like DINOv2, that does not need any training annotation, can exploit any image. It also shows that OVDiff with StableDiffusion is not as strong as DINOv2 with ChatGPT.

\simpletextparagraph{2D-3D VFM distillation evaluation, NN vs LR.}
The above unprojection results do not reflect 2D–3D distillation quality, which we evaluate here.
We label 3D points by matching their distilled 3D VFM features with the 2D VFM encoding of prototype images.
We evaluate distillations of the best two VFMs in \cref{tab:2d_3d_features_abl}(a). We use the official models of ScaLR \cite{scalr}, that distills DINOv2, and OpenScene \cite{peng2023openscene}, that distills OpenSeg, staying in the CLIP image feature space \renaud{with improved dense visual features}.
Results in \cref{tab:2d_3d_features_abl}(b) show that the best performance of DINOv2 as a 2D VFM translates to point clouds after ScaLR distillation, although with logistic regression (LR) rather than nearest neighbor (NN). This prompted us to study \scalrplus, an improved ScaLR distillation scheme (\cref{sec:2d3dvfmdistill}), which ended up outperforming other distillations by a large margin \renaud{(see below).}

\simpletextparagraph{\scalrplus.}
\cref{tab:scalrplusablat,tab:img_src_abl} confirms the relevance of the changes brought to ScaLR to define {\scalrplus}. 
Notably, as shown in \cref{tab:2d_3d_features_abl}(b), the original ScaLR architecture, despite being a strong general-purpose backbone to construct 3D representations, performs below OpenScene under nearest neighbor matching. Its performance improves substantially only after our targeted \scalrplus modifications, which explicitly strengthen compatibility between the 2D and 3D feature spaces.

% \begin{table}[t]
\begin{wraptable}{r}{0.48\textwidth}
\centering
\vspace{-4mm}
\caption{\textbf{Impact of image source} on \ours (mIoU\%)
with NN or LR on NS val set. We evaluate \textsl{1 image per prompt}. 
}
\label{tab:img_src_abl}
\resizebox{1.0\linewidth}{!}{
\begin{tabular}{l@{~}r|c@{~~}c|c@{~~}c|c@{~~}c|c@{~~}c}
\toprule
\multirow{3}{*}{Distill.\,\rlap{method}~}
&& \multicolumn{2}{c|}{ChatGPT}  
& \multicolumn{2}{c|}{\nermin{Qwen}}
& \multicolumn{2}{c|}{Gemini}   
& \multicolumn{2}{c}{Web}

% & \multicolumn{2}{c}{Flux}
\\
\cmidrule(lr){3-10}
&& NN & LR & NN & LR & NN & LR & NN & LR  \\
\midrule
 ScaLR & \cite{scalr}  &  34.8 & 39.7 & 34.7 & 39.1  & 32.9 & 34.5 & 32.1 & 32.3  \\
\scalrplus & (ours) &  42.9 & \textbf{46.6}  & 38.9 & 43.5 & 42.2 & 42.3 & 38.4 & 37.3  \\  
\bottomrule
\end{tabular}}
% \vspace{-5mm}
\end{wraptable}

\simpletextparagraph{Sources of prototype images.}
\cref{tab:img_src_abl} evaluates our method on various sources of prototype images. 
ChatGPT provides significantly better prototypes. Gemini images are good too, though some can be slightly futuristic. Importantly, the open-source Qwen-Image model achieves 43.5\% mIoU, compared with 46.6\% for ChatGPT and 42.3\% for Gemini, demonstrating that IGLOSS does not rely on a particular closed-source generator and remains effective with a fully open-source alternative.
\renaud{In any case, \ours only requires generating one or a few images per class.}
These results also confirm that LR outperforms NN.
Last, it shows that \scalrplus consistently outperforms ScaLR across all settings, under both NN and LR. 

\begin{wraptable}{r}{0.47\textwidth}
\centering
\small
\vspace{-4mm}
\caption{\textbf{Ensembling prototypes from various sources.} Tested on NS val set. }
\label{tab:merging_templates}
\begin{tabular}{l@{~\,}c } %l@{~~\,}
\toprule
    Prototypes \renaud{per class}
    & \llap{(LR)~}mIoU\%
\\
\midrule
1 $\times$ ChatGPT &  46.6 \\
\midrule
\rowcolor{green!15}
2 $\times$ ChatGPT &  47.5 \\
ChatGPT + Qwen &  47.3 \\
ChatGPT + Gemini &  47.9 \\
ChatGPT + Web & 47.7 \\
\midrule
3 $\times$ ChatGPT & 47.2  \\
ChatGPT + Qwen + Web &  47.9 \\
ChatGPT + Gemini + Web &  48.0 \\
\midrule
4 $\times$ ChatGPT & 46.9  \\
ChatGPT + Qwen + Gemini + Web &   \textbf{48.2} \\

\bottomrule
\end{tabular}%
\vspace{-5mm}
\end{wraptable}

\simpletextparagraph{Ensembling prototypes from different sources.}
We study whether mixing image sources improves prototype diversity. Since ChatGPT provides the strongest prototypes (\cref{tab:img_src_abl}), we add prototypes from other sources on top of ChatGPT images, while scaling the number of ChatGPT prototypes. 
Results are shown in \cref{tab:merging_templates}.  
Although source combinations achieve the best results, we keep the final \ours model simple and use only two ChatGPT images \renaud{in the main result table (\cref{table:sota_semantic_ovss})}.

\simpletextparagraph{Impact of the number of prototypes.}  
We varied the number of templates from 1 to 8 and observed that performance plateaus at two. Across five runs with two randomly sampled prototypes per class, the \renaud{standard deviation} % variance 
was 0.08 \renaud{mIoU pt}. Detailed results are provided in \appref{sec:number_of_protos}.

\simpletextparagraph{Limitations.}
We show that, for 3D OVSS, IG from text is an effective alternative to the usual text-image alignment; we however do not show that it actually reduces the text-image modality gap.
Generating images currently takes between 3-5 seconds, which is longer than a feedfoward pass to obtain a text embedding. (LR fitting time is thus negligible.) Nevertheless, we show that a limited number of images are sufficient to obtain state-of-the-art results. Besides, very fast image generators have started to emerge \cite{liu2024instaflow}, which \ours can directly benefit from. In any case, IGLOSS is not intended for real time but, e.g., for auto-labeling or retrieval (\cref{sec:intro}). Failure cases are in \appref{sec:qualitative}.

\section{Conclusion}
\label{sec:conclusion}

\ours is a novel approach for 3D OVSS that evades the need for text and visual data to share the same feature space. It combines the use of IG with a well-crafted 2D-3D distillation method, \scalrplus, and defines a new SOTA for 3D OVSS and annotation-free close-set semantic segmentation. 
This joint use of IG and aligned 2D-3D VFMs opens perspectives in OV approaches. The simplicity of \ours makes it a strong baseline to build on. It also provides a bridge with few-shot methods.

\begin{ack}
We acknowledge EuroHPC Joint Undertaking for awarding the project ID EHPC-REG-2024R02-234 access to Karolina, Czech Republic. This work was supported by the European Union’s Horizon Europe research and innovation programme under grant agreement No 101214398 (ELLIOT). Gilles Puy contributed to this work while employed by Valeo; he accessed and processed all datasets on valeo.ai compute infrastructures; he is presently employed by Meta.
\end{ack}

\bibliography{main} 
\bibliographystyle{unsrtnat}

%%%%%%%%%%%%%%%%%%%%%%%%%%%%%%%%%%%%%%%%%%%%%%%%%%%%%%%%%%%%

\newpage

\appendix
\setcounter{secnumdepth}{5}

\title{\ours: Image Generation for Automotive Lidar Open-vocabulary Semantic Segmentation \\ --- Supplementary Material ---} 
\maketitle
\setcounter{figure}{5}
\setcounter{table}{6}

We supplement the main paper with more details and comparisons with related work (\cref{sec:moredetail}), classwise results of our method (\cref{sec:classwise-results}), \nermin{text prompts used to generate our prototype images (\cref{sec:prompts})},
\nermin{experiments with different number of templates and pairs (\cref{sec:number_of_protos})}, and qualitative results for the \ns and \sk datasets (\cref{sec:qualitative}). Finally, \cref{sec:licenses} provides details on compute resources and asset licenses.

\smallskip

\section{More details on related work}
\label{sec:moredetail}

Sections 1 and 2 of the main paper already discuss related work; in this section, we go into more details. We discuss the modality gap issue between image and text (\cref{sec:modgapissue}) and methods that try to address it (\cref{sec:modgapaddress}). We also analyze different levels of openness regarding vocabulary and semantic concepts (\cref{sec:levelov}). Last, we summarize the main ideas of 3D open-vocabulary semantic segmentation (OVSS) methods that apply to driving data (\cref{sec:detailmethods}) and we compare them to our method (\cref{sec:comparemethods}).

\subsection{Image-text modality gap in VLMs}
\label{sec:modgapissue}

Several authors have highlighted \cite{liang2022modalitygap, qian2023inmap, yamaguchi2025cliprefine, kang2025isclipideal}, via both theoretical analyses and empirical studies, an inherent modality gap between text and images in Visual Language Models (VLMs) like CLIP \cite{radford2021clip}, which try to directly align images and text. 

One of the main causes is the disparity of initializations induced by the need to train separate encoders of different nature for each modality \cite{liang2022modalitygap}. It tends to create a feature gap that contrastive learning, which many VLMs rely on, does not bridge when training \cite{liang2022modalitygap, qian2023inmap}. Other reasons include intrinsic goal contradictions for building joint representations of images and text \cite{kang2025isclipideal}. 

Besides, analyses of the joint feature space suggest that the uniformity of representations on the hypersphere is also a key aspect that has to be addressed while trying to minimize the gap between image and text features, especially since direct gap minimization may compromise the uniformity of these representations \cite{tongzhouw2020hypersphere, yamaguchi2025cliprefine}.

\subsection{Addressing the VLM modality gap}
\label{sec:modgapaddress}

A number of approaches have been proposed to address this modality gap issue, at different stages of the training and inference of the VLM: in the VLM pretraining stage (\cref{sec:modgappretraining}), as a refinement coming after VLM pretraining (\cref{sec:modgappostpretrain}), in a downstream finetuning stage (\cref{sec:modgapfinetuning}), or as a test-time tuning (\cref{sec:modgaptesttime}).

\subsubsection{VLM pretraining.}
\label{sec:modgappretraining}

When pretraining the text and image encoders, the feature space alignment can be improved with additional or alternative loss terms \cite{lee2022uniclip, goal2022cyclip}, textual negative engineering \cite{doveh2023svlc}, augmentation awareness \cite{lee2022uniclip}, or self-supervised learning training recipes \cite{li2022declip}. 

Another direction is to introduce some sharing at architecture level between image and text, e.g., sharing the weights of a single transformer after a first separate embedding stage \cite{you2022msclip}, or sharing tokens \cite{chen2023fdt}. 

Although these approaches contributes to a better feature alignment, a modality gap remains. Besides, these approaches involve full pretraining, which requires large-scale image-text pair datasets and heavy compute to reach good practical performance.

\subsubsection{VLM post-pretraining.}
\label{sec:modgappostpretrain}

A better alignment can also be obtained after finetuning with a loss that tries to match feature distributions rather than directly-paired features, while using the pretrained VLM as teacher to preserve past knowledge \cite{yamaguchi2025cliprefine}.

\subsubsection{VLM finetuning for a downstream task.}
\label{sec:modgapfinetuning}

Another line of work focuses on finetuning for a particular downstream task or dataset, possibly with learnable parameters on top of a VLM with frozen weights.

This includes textual prompt learning \cite{zhou2022coop, zhou2022cocoop}, visual prompt tuning \cite{jia2022vpt} or jointly learning prompts for both modalities \cite{khattak2023maple} or for multiple tasks \cite{shen2024mvlpt}. 

Still keeping the visual and textual backbones frozen, multi-modal adapters with shared projection layers, operating between encoder transformer blocks, have been proposed to preserve generalization while improving discrimination for specific tasks or data \cite{yang2024mma}. 

Observing that the poor alignment of textual and visual representations, even after finetuning, is partly due to separate embedding spaces, \cite{oh2023multimodalmixup} uses a mixup strategy to distribute multi-modal features more uniformly on the unit hypersphere when finetuning. 

\subsubsection{Test-time tuning.}
\label{sec:modgaptesttime}

Last, some approaches only operate at test time. 

A first category of method tries to optimize the textual prompts, e.g., tuning them at test time based on consistency under augmentations \cite{shu2022tpt} or using a Large Language Model (LLM) to elaborate class descriptions highlighting, at textual level, what characteristics a class is expected to have \cite{pratt2023cupl, menon2023visual}. 

A second kind of method operates at image level, using few shots and their VLM visual encoding to adjust classification predictions \cite{udandarao2023susx}. In fact, \cite{qian2023inmap} argues, based on theoretical justifications, that it is better for alignment to only operate in the vision space, and use text features only to get pseudo-labels from which vision prototype features can then be obtained. This is what \ours does, although very differently from \cite{qian2023inmap}.

\subsection{Levels of openness}
\label{sec:levelov}

3D OVSS methods are \emph{open-vocabulary} in that any text prompt can be used to define a semantic class to segment. The time when these text prompts are used and when the \emph{semantic concepts} are fixed, compared to the amount of processing that follows (training for distillation, inference, etc.), however widely varies.

There is in fact a hierarchy of openness w.r.t.\ vocabulary and semantic concepts: from closed-set models trained with open, prompt-based labels (marked \xmark\ in \cref{tab:relwork}), to models semantically biased by input text prompts but still OV (marked \mmmark), to OV models possibly semantically biased by the power of an underlying captioner (marked \mmark), to fully OV and open-world models (marked \cmark). \ours falls in that latter category (\intropt\openvoc). 

See however \cref{sec:distillbias} for an intrinsic remaining bias of all methods, originating from necessary 2D-3D distillation.

\subsubsection{3D training from labels.}

A first kind of 3D OVSS methods \cite{chen2023cns, kang2024hicl, zou2025adaco, sun2025afov, gebraad2025leap,samet2026losc} starts by predicting 2D semantic labels from text prompts via the 2D VLM, and trains the 3D network with these pseudo-labels. 

These \emph{image-labeling-time OV methods} are open-vocabulary before training time, but not anymore at test time, when they are closed-set as they can then only segment the previously-defined classes.

\subsubsection{3D training with the help of text prompts.}
\label{sec:trainingwtextprompts}

A second kind of methods \cite{clip2scene, wang2024ggsd, li2025sas} uses given text prompts of classes of interest early in the process too, but only to guide 2D-3D distillation, which is then performed on 2D VLM features rather than on semantic labels. 

These \emph{training-time OV methods} are open-vocabulary in the sense that the 3D network produces features that can be compared at test time to the VLM features of any text, not just of the text prompts provided earlier. However the training is biased towards the closed set of previously-defined classes. The method may thus not fully be open regarding the capacity to recognize semantic concepts beyond these known classes. 

\subsubsection{3D training with the help of a captioner.}

A third kind of methods \cite{jiang2024ov3d, wei20253davs} also exploits some text prompts early, but these texts are obtained automatically via an image-to-text model that describes with words images in the dataset, rather than being provided by the user. As is the case for the second category of methods above (\cref{sec:trainingwtextprompts}), these descriptions are then used as text prompts to guide 2D feature distillation onto 3D. Again, the 3D network then produces features that can be matched to the feature of any text prompt.

Such \emph{auto-vocabulary OV methods} are thus open-vocabulary at interface level. Yet, the training is somehow biased towards the kind of classes that could be identified automatically at image description time. These methods are more open than training-time 3D OVSS methods, but still possibly not entirely open to all semantic concepts present in the VLM (pre)training dataset.

\subsubsection{3D training from 2D features only.}

A fourth and last kind of method \cite{peng2023openscene, osep2024sal} makes no assumption regarding semantic classes that may appear in the distillation training set and at test time. 

Existing such \emph{test-time OV methods} directly distill 2D VLM features into a 3D network, which produces 3D features that can be compared to any text features. These methods are not only open-vocabulary at test time; they are also fully open in terms of semantic concepts, within the limits of the concepts appearing in the VLM pretraining dataset and in the 2D-3D distillation dataset (\cref{sec:distillbias}).

Our approach differs in that we only align visual features, i.e., 2D and 3D features. Free text prompts are then used to generate images from which 2D features are extracted and matched to 3D points. \ours is thus also a test-time 3D OVSS method in the above sense.

\subsubsection{The remaining distillation bias.}
\label{sec:distillbias}

Due to the lack of enough 3D data paired with text to directly train a decent 3D OV model, images have to be used as an intermediate modality to construct 3D OVSS, via 2D-3D knowledge distillation. It induces a somewhat neglected bias on the dataset used for this distillation, especially since the quantity of aligned 2D-3D data available is also small compared to the amount of data a 2D VLM is trained on. 

Indeed, resulting distilled 3D networks are not as open as the 2D VLM or VFM they are derived from: only semantic concepts that are present in the target dataset and identified by the 2D VLM (by producing corresponding features) are distilled onto 3D. For instance, while a car or a bookshelf can be easily identified by a 2D VLM, it is likely that the 3D point cloud of a car (resp.\ a bookshelf) could not be recognized as such by a 3D network distilled from that same 2D VLM if based on an indoor (resp.\ outdoor) dataset.

This is however not an issue in practice, as long as the training dataset is representative enough of the test data. \ours actually benefits from a 3D VFM that is distilled from several datasets \cite{scalr}, which improves its generalization capability.

On another note, a limitation also comes from the fact that the correspondence between 3D points and text is indirect, as it is constructed via 2D visual features. It is thus also sensitive to noise in 2D-3D correspondences, e.g., due to issues in sensor calibration, sensor synchronization or sensor parallax.

\subsection{3D OVSS methods applied to driving data}
\label{sec:detailmethods}

We describe in this section the main ideas of 3D OVSS papers that present applications to lidar data in driving settings. The presentation is organized according to their level of vocabulary and semantic concept openness (\cref{sec:levelov}).

\subsubsection{Image-labeling-time 3D OVSS methods.} 

Image-labeling-time 3D OVSS methods use text prompts to pseudo-label images before the 3D network is trained, and are thus closed-set at test time.

\textbf{CNS} \cite{chen2023cns} refines CLIP-based \cite{zhou2022maskclip} dense pseudo-labels on SAM~\cite{sam} masks, and uses them to train both a 2D and a 3D network. A second training stage for regularization includes self-training and cross-training (between 2D and 3D) as well as feature alignment with SAM.

\textbf{HICL} \cite{kang2024hicl} trains a 3D model from 2D pseudo-labels generated by MaskCLIP \cite{zhou2022maskclip} using a contrastive loss involving geometric consistency given by superpoints \cite{papon2013vccs} and correlation learning, both intra- and inter-scenes.

\textbf{AdaCo} \cite{zou2025adaco} obtains image masks from FastSAM \cite{zhao2023fastsam}, which are labeled automatically with text using SSA \cite{chen2023ssa}. These text labels are then matched with text prompts using word2vec \cite{church2017word2vec}. Unprojected 3D semantic labels are denoised by voxelwise voting across a short sequence of frames. Two training recipes reduce the impact of pseudo-labeling noise: one based on the learning curve, and another one on a clustering of instances using DBSCAN \cite{ester1996dbscan}. The whole pipeline is trained with a robust loss.

\textbf{LeAP} \cite{gebraad2025leap} combines Grounding DINO \cite{gdino} and SAM \cite{sam}, similarly to what GSAM \cite{gsam} does, to get image pseudo-labels from text prompts. After unprojection onto 3D, spatial and time consistency is enforced based on 2D class probabilities and on a 3D network trained on the most confident pseudo-labels.

\textbf{LOSC} \cite{samet2026losc} denoises 2D pseudo-labels from OpenSeeD \cite{openseed}, enforcing robustness under augmentations and consistency through time at voxel level after unprojection onto 3D. A 3D backbone is then self-trained using iterative finetuning from a pretrained ScaLR network \cite{scalr}.

\subsubsection{Training-time 3D OVSS methods.} 

Training-time 3D OVSS methods use given text prompts of classes of interest not only at test time but also to guide 2D-3D distillation. They are thus biased towards these classes despite being functionally open-vocabulary.

\textbf{CLIP2Scene}~\cite{clip2scene} trains a 3D network with a contrastive loss to align point features to CLIP features of text prompts defining classes of interest, via point-pixel correspondences and MaskCLIP 2D features \cite{zhou2022maskclip}.

\textbf{GGSD} \cite{wang2024ggsd} distills CLIP into 3D using correspondences of pixels with both points and unsupervised superpoints, maximizing cosine similarity. A second training stage adds a contrastive self-distillation loss term that exploits the predictions of the 3D network, based on given text prompts.

\textbf{AFOV} \cite{sun2025afov} pretrains its 3D network using image masks from SAN \cite{xu2023san}, which is CLIP-based, using a fixed class dictionary. AFOV turns these masks into superpixels and corresponding superpoints, and uses a contrastive loss on superpoints involving CLIP features of both superpixels and text. The network is then trained on SAN pseudo-labels. Additionally, AFOV uses a non-parametric module for label denoising. It also possibly exploits images at test time to slightly improve performance.

\textbf{SAS} \cite{li2025sas} ensembles several VLMs, possibly of different nature, e.g., LSeg \cite{li2022lseg}, which is CLIP-based, or OpenSeg \cite{openseg} and SEEM \cite{zou2023seem}, which are not. To that end, SAS uses TAP \cite{pan2024tap} as a captioner to generate descriptions for each mask of each VLM, to enrich the initial labels with more semantic information. A shared text encoder, like CLIP's, then aligns the visual information of different VLMs via the encoded rich descriptions. To ensemble information of the different VLMs, SAS relies on a pre-built vocabulary consisting of common classes in the scene (e.g., car, bicycle). For each of these classes, Stable Diffusion (SD) \cite{rombach2022sd} is used to generate several images. Masks of the object of the target class in each image are obtained by binarizing the cross-attention map of SD. These coarse masks are then refined by asking SAM \cite{sam} to generate masks from (2D) point prompts inside the coarse masks. The mIoU between VLM masks and SAM masks defines a VLM-specific score for each predefined classes. This class-wise score is then used for ensembling the information provided by the VLMs. Distillation onto 3D relies on superpoints as in GGSD \cite{wang2024ggsd} and is performed with an exponential moving average (EMA).

\subsubsection{Auto-vocabulary 3D OVSS methods.} 

Auto-vocabulary 3D OVSS methods leverage a separate VLM that provides text descriptions from images and use these descriptions to drive the 2D-3D distillation.

\textbf{OV3D} \cite{jiang2024ov3d} gets rich image descriptions using an Large Visual Language Model (LVLM), namely LLAVA \cite{liu2023llava}. These descriptions are then used to pseudo-label image masks from SEEM \cite{zou2023seem}, using SEEM as well for text-image alignment. The 2D pseudo-labels are then unprojected onto 3D for distillation into a 3D backbone on a target dataset. OV3D may also be combined with OpenScene-3D \cite{peng2023openscene} for a slightly better performance.

\textbf{3D-AVS} \cite{wei20253davs} focuses on text generation to automatically label segments. It uses existing CLIP-aligned encoders for text (CLIP \cite{radford2021clip}), image (OpenSeg \cite{openseg}) as well as points (OpenScene \cite{peng2023openscene} w/ OpenSeg \cite{openseg}), trained on the same target dataset. 3D-AVS first trains on the target dataset a point-cloud captioner from an image captioner (BLIP-3 \cite{xue2025xgenmmblip3}). Given a point cloud, this captioner generates text tags that are used for similarity-based 3D segmentation. At test time, these tags are mapped via LAVE \cite{ulger2025lave} into predefined classes using a Large Language Model (LLM), namely GPT-4o \cite{gpt4o}.

\subsubsection{Test-time 3D OVSS methods.} 

Test-time 3D OVSS methods preserve the openness of both the vocabulary and the corresponding semantic concepts until evaluation time.

\textbf{OpenScene}~\cite{peng2023openscene} distills a CLIP-like model (LSeg \cite{li2022lseg} or OpenSeg \cite{openseg}) into a 3D network on a target dataset, maximizing the similarity of 2D and 3D features of short sequences. While inference on pure 3D data is possible, the best (official) results are obtain by ensembling 2D and 3D alignment scores with text, requiring images at test time.

\textbf{SAL} \cite{osep2024sal} gets 3D masks from unprojected SAM \cite{sam} image masks, which are assigned MaskCLIP features \cite{ding2023maskclip} and refined via DBSCAN clustering. A 3D backbone is trained on a target dataset to predict MaskCLIP features as well as SAM masks.

\subsection{Comparing 3D OVSS methods for driving data}
\label{sec:comparemethods}

\cref{tab:relwork} lists 3D OVSS methods that apply to driving lidar data and summarizes their main characteristics.

\subsubsection{Image-text alignment.}

Almost all of these methods rely on a 2D VLM, mostly CLIP \cite{radford2021clip} or one of its dense variants \cite{zhou2022maskclip, ding2023maskclip}. This 2D VLM provides an alignment between image (\img) and text (\txt), via a shared feature (\feat) space: \quot{\ift}. However, this architecture restrains the alignment quality and requires a VLM trained on a large amount of curated data or extra supervision. \ours does not need such a VLM to bridge image and text (\intropts\easyitalign\nohugedata\ in \cref{sec:intro}).

Moreover, the training architecture of CLIP makes it difficult to obtain good segment boundaries \cite{zhou2022maskclip, li2022lseg, ding2023maskclip, dong2023maskclip}. An auxiliary foundation model for mask (\msk) segmentation (\quot{\igm}), such as SAM \cite{sam}, is thus often used to generate masks and learn with segments to improve the quality of final segmentation \cite{chen2023cns, zou2025adaco, gebraad2025leap, li2025sas, osep2024sal, zhang2025sal4d}. In contrast, \ours leverages a 2D Vision Foundation Model (VFM) good at dense tasks (\intropt\betterfeat), namely DINOv2 \cite{oquab2024dinov2}, which incidentally is trained with no supervision contrary to SAM. Besides, the VFM is used in a black box manner, which facilitates modularity and improvement of \ours when a better VFM becomes available (\intropt\bboxfm).

Only AdaCo \cite{zou2025adaco} and 3D-AVS \cite{wei20253davs} differ in global architecture. Rather than directly extracting VLM features of image masks (\quot{\igf}), AdaCo first generates textual descriptions of masks with a captioner (\quot{\igt}) \cite{chen2023ssa}, and compares these descriptions to the given text prompts via word2vec \cite{church2017word2vec} (\quot{\tgf} for both texts). As for 3D-AVS, it uses generated captions similarly, but from a trained point captioner and still relying on an underlying VLM. (Some other methods use captioners too, but for other purposes, such as text enrichment \cite{li2025sas} or auto vocabulary~\cite{jiang2024ov3d}.)

In contrast, \ours needs neither encoders aligned both on image and text (\quot{\ift}), nor captioners (\quot{\igt}). It bridges image and text ``in the opposite direction'': using an image generator conditioned on text (\quot{\tgi}). The alignment of points and text then just comes from the 2D-3D distillation (\quot{\ifp}) of a 2D VFM. At test time, no image is needed (\intropt\notrick); we only have to compare the features of the point clouds (\quot{\pgf}) to the features of the generated images (\quot{\tif}).

\subsubsection{Simplicity.}

Another difference of \ours with most other methods is that it is extremely simple (\intropt\simplicity) and does not require any training, beyond the direct 2D-3D feature distillation, for which there are proven recipes such as ScaLR \cite{scalr}, and even pretrained 3D network readily available \cite{scalr}. The image generator (\quot{\tgi}) is used as a complete black box (\intropt\bboxfm), contrary to SAS that uses Stable Diffusion only as a way to get cross-attention maps and extract intermediate masks. In contrast, many other methods use a heavy machinery relying on many modules and possibly involving several training stages. For instance, SAS \cite{li2025sas} and 3D-AVS \cite{wei20253davs} leverage simultaneously seven underlying external tools, 
and CNS \cite{chen2023cns}, CLIP2Scene \cite{clip2scene}, GGSD \cite{wang2024ggsd}, AdaCo \cite{zou2025adaco} and SAS train in two phases, while LOSC \cite{samet2026losc} performs several self-training iterations.

\subsubsection{Vocabulary and semantic concept openness.}
\label{sec:compareopenvoc}

As discussed above, there are different levels of vocabulary and semantic concept openness for 3D OVSS methods, which may come come with biases (\cref{sec:levelov}). An extra bias may also come from the 2D-3D distillation (\cref{sec:distillbias}). In fact, all 3D OVSS methods actually use the target (evaluation) dataset to train the distillation in the first place. 

In contrast (\intropt\openvoc), \ours is fully open-vocabulary and it uses a foundation model that is trained on several datasets \cite{scalr}, which reduces the bias, although it does not eliminate it entirely.

\subsubsection{Use of images or scan sequences at test time.}

Some 3D OVSS methods \cite{peng2023openscene, wei20253davs, sun2025afov} propose to also use images at test time to improve performance. The underlying 2D VLM then provides soft labels \cite{sun2025afov} or features \cite{peng2023openscene, wei20253davs} for ensembling with 3D information, after unprojection. It has however been argued \cite{wang2024ggsd} that 2D-3D ensembling at test time has high storage and computational costs, especially since a lidar frame is generally covered by several cameras (e.g., 6 in nuScenes).

Other methods exploit a sequence of scans \cite{peng2023openscene, wei20253davs, zou2025adaco} rather than a single frame, merging them to increase point density.

\ours needs neither images nor scan sequences at test time (\intropt\notrick) to get SOTA results (\cref{table:sota_semantic_ovss}).

\subsubsection{2D VLM ensembling.}

While OV3D \cite{jiang2024ov3d} proposes an integration
with OpenScene-3D \cite{peng2023openscene}, SAS\cite{li2025sas} is specifically designed to ensembles VLMs, possibly of different nature.

Ensembling is not required in \ours (\intropt\sota) to get SOTA results (\cref{table:sota_semantic_ovss}). It is however particularly easy to implement (see \cref{sec:ensembling}) and can further boost results by 3.6 to 4.6 mIoU pts.

\subsubsection{Datasets used for evaluation.}

As visible in \cref{table:sota_semantic_ovss}, all above methods (\cref{sec:detailmethods}) evaluate on the nuScenes dataset \cite{nuscenes}, which features a rotating lidar and six cameras around the ego-vehicle, covering together a 360\textdegree\ field of view.

However, only very few of them evaluate on SemanticKITTI \cite{semantickitti}. One of the reasons could be that this dataset also features a rotating lidar, but just one front camera. Only information from a single front view can thus be distilled into a whole 3D scan, which makes it more difficult for the distilled network to generalize to unseen areas on the side and on the back of the ego-vehicle. 

Again (\cref{sec:compareopenvoc}), \ours suffers less from this situation as it leverages a 3D VFM \cite{scalr} that is trained from several datasets at the same time \cite{nuscenes, semantickitti, xiao2021pandaset}. This multi-domain distillation is beneficial in general, compared to distilling on a single dataset, and in particular for SemanticKITTI where linear probing performance gets a +12.7 mIoU\% gain \cite{scalr}.

\begin{table*}[t]
\small
\ra{1.2}
\newcommand*\rotext{\multicolumn{1}{R{65}{1em}}}
\setlength{\tabcolsep}{3.5pt}
\centering
\caption{\textbf{Classwise semantic segmentation results (IoU\%) of \ours and its variants on \ns val set}.}
\label{tab:classwise-ns}
\resizebox{\linewidth}{!}{%
\tabcolsep 2pt
\begin{tabular}{l  c | c c c c c c c c c c c c c c c c}
\toprule 
    Method

        & \rotext{\bf mIoU\%}
        & \rotext{barrier}
        & \rotext{bicycle}
        & \rotext{bus}
        & \rotext{car}
        & \rotext{const. veh.}
        & \rotext{motorcycle}
        & \rotext{pedestrian}
        & \rotext{traffic cone}
        & \rotext{trailer}
        & \rotext{truck}
        & \rotext{driv. surf.}
        & \rotext{other flat}
        & \rotext{sidewalk}
        & \rotext{terrain}
        & \rotext{manmade}
        & \rotext{vegetation}
\\
\midrule
    \ours 
        % & \xmark
        & 47.5
        & 53.4
        & 21.7
        & 63.5
        & 67.4 
        & 39.9 
        & 60.9 
        & 52.2 
        & 37.0
        & 15.8 
        & 31.6 
        & 86.0 
        & 6.8
        & 45.3
        & 46.7
        & 68.4 
        & 63.7

\\

    \oursens 

        & 51.1
        & 42.9 
        & 18.5 
        & 74.1 
        & 73.4 
        & 32.6 
        & 71.0 
        & 58.1
        & 39.7 
        & 16.7 
        & 43.4 
        & 86.5 
        & 4.8
        & 46.9 
        & 55.5 
        & 75.6 
        & 78.5        
\\
\midrule
 \oursclosed 
        & 49.6
        & 56.7 
        & 23.6 
        & 64.6 
        & 69.0
        & 41.9 
        & 63.8 
        & 58.6 
        & 39.9 
        & 16.3 
        & 32.7 
        & 87.2 
        & 7.3
        & 46.6
        & 49.1 
        & 70.2 
        & 66.2
\\
\oursclosedens 
        & 54.1
        & 43.7 
        & 21.3 
        & 75.8 
        & 74.3 
        & 38.0 
        & 76.0 
        & 66.7 
        & 51.3 
        & 16.8 
        & 46.4 
        & 88.1 
        & 4.6
        & 49.6 
        & 56.8 
        & 75.8 
        & 79.5
\\
\bottomrule
\end{tabular}}
\end{table*}

\begin{table*}[t]
\small
\ra{1.2}
\newcommand*\rotext{\multicolumn{1}{R{65}{1em}}}
\setlength{\tabcolsep}{2.2pt}
\centering
\caption{\textbf{Classwise semantic segmentation results (IoU\%) of \ours and its closed-set variant on \sk val set}. (There is no result on \oursens\ and \oursclosedens as there is no OpenScene training on SemanticKITTI.)}
\label{tab:classwise-sk}
\resizebox{1.0\linewidth}{!}{%
\tabcolsep 2pt
\begin{tabular}{l c | c c c c c c c c c c c c c c c c c c c}
\toprule
    Method 
  
        & \rotext{\bf mIoU\%}
        & \rotext{car}
        & \rotext{bicycle}
        & \rotext{motorcycle}
        & \rotext{truck}
        & \rotext{other-vehicle}
        & \rotext{person}
        & \rotext{bicyclist}
        & \rotext{motorcyclist}
        & \rotext{road}
        & \rotext{parking}
        & \rotext{sidewalk}
        & \rotext{other-ground}
        & \rotext{building}
        & \rotext{fence}
        & \rotext{vegetation}
        & \rotext{trunk}
        & \rotext{terrain}
        & \rotext{pole}
        & \rotext{traffic-sign}
\\ 
\midrule
    \ours 
        & 34.3
        & 76.0
        & 14.4
        & 40.5 
        & 20.2
        & 10.5
        & 55.5 
        & 36.4 
        & 0.0
        & 83.4
        & 0.0 
        & 54.3
        & 0.6
        & 62.5 
        & 0.7 
        & 79.1
        & 33.0 
        & 64.9
        & 0.0
        & 19.8 
        
\\ 
    \oursclosed 
        & 39.4
        & 87.8
        & 21.6
        & 46.2
        & 19.8 
        & 11.2
        & 70.6
        & 53.2 
        & 0.0 
        & 89.6
        & 0.0
        &  64.9
        & 0.7
        & 68.3
        & 0.6
        &  84.6 
        & 39.3
        & 69.3 
        & 0.0
        & 20.7

\\ 

\bottomrule
\end{tabular}}
\end{table*}

\subsection{Test-time adaptation (TTA) for Open Vocabulary tasks} 

Test-time adaptation (TTA) for OV tasks has emerged in the literature as a prominent general strategy to bridge the gap between pretrained models and flexible downstream tasks. 
It includes the exploitation of side knowledge, e.g., dictionary-based contextual prompts \cite{wysoczanska2025testtimecontrastiveconcepts} or external more or less heavy models \cite{liu2025test, wysoczanska2025testtimecontrastiveconcepts, clipdiy} to adjust prompts or segments. It also includes light optimization stages for test-time prompt tuning (TPT) \cite{shu2022tpt, feng2023difftpt}, feature \cite{noori2025testtimeadaptation} or proposal reweighting \cite{belal2025vlod}, learning on-the-fly a classifier for novel classes \cite{pham2024lp}, or more involved test-time training (TTT) \cite{zhou2026zeroshotovhumanmotion}.

These works, recognized as OV, demonstrate that adapting at test time to the target sample (or even distribution \cite{karmanov2024efficienttta}) is a increasingly common mechanism for enhancing the performance of open-vocabulary methods. It justifies our in-place replacement of the classical nearest-neighbor (NN) classification by a logistic regression (LR) fitted at test time, which is in any case very fast ($\sim$\,70ms).

\section{Classwise results}
\label{sec:classwise-results}

\cref{tab:classwise-ns,tab:classwise-sk} respectively present classwise semantic segmentation results on \ns \cite{nuscenes} and \sk \cite{semantickitti} validation sets as complementary results to \cref{table:sota_semantic_ovss} in the main paper.

\section{Text prompts}
\label{sec:prompts}
    \def\notp#1{\textcolor{blue}{\textsl{#1}}}
    \def\notextprompt{\hspace*{5em}No text prompting; we use the same prototype images as for nuScenes}
    \def\notextprompt{\hspace*{1.5em}\sl No prompting, same images as nuScenes}
    \def\notextprompt{\hspace*{1em} \notp{nuScenes prompts/images for}}
    \newcommand\Tobj{$T_\text{obj}$}
    \newcommand\Tstuff{$T_\text{stuff}$}
    \newcommand\Tdrive{$T_\text{drive}$}
    \newcommand\Tobjdrive{$T_\text{dr-obj}$}
    \newcommand\Tstuffdrive{$T_\text{dr-stuff}$}
    \newcommand\pTobj[1]{\Tobj(~#1~)}
    \newcommand\pTstuff[1]{\Tstuff(~#1~)}
    \newcommand\pTdrive[1]{\Tdrive(~#1~)}
    \newcommand\pTobjdrive[1]{\Tobjdrive(~#1~)}%{\Tdrive{\Tobj{#1}}}
    \newcommand\pTstuffdrive[1]{\Tstuffdrive(~#1~)}%{\Tdrive{\Tstuff{#1}}}

\begin{table*}[p]
    \centering
    \small
    \caption{\textbf{Our context prompt templates used to generate prototype images.}}
    \label{tab:text_prompts} 
    \setlength\extrarowheight{4pt}
    \begin{tabular}{l|l}
    \toprule
    Template & Text prompt \\
    \midrule
    \pTobj{$c$}   & {generate an image of $c$ with white background} \\
    \pTstuff{$c$} & {generate an image of $c$ covering the whole image} \\
    \pTdrive{$p$}   & {$p$, similar to what you see along roadsides and in cities} \\
    \pTobjdrive{$c$} & {\pTdrive{\pTobj{$c$}}} \\
    \pTstuffdrive{$c$} & {\pTdrive{\pTstuff{$c$}}} \\
    \bottomrule
    \end{tabular} 
    \vspace{4ex}

    \caption{\textbf{Our classwise text prompts used to generate prototype images for nuScenes.} Templates \Tobj, \Tstuff, \Tobjdrive\ and \Tstuffdrive\ are defined in \cref{tab:text_prompts}.}
    \label{tab:text_prompts_ns} 
    \begin{tabular}{l|l|l}
    \toprule
    Class & Templ. & Text prompts \\
    \midrule
    \text{pedestrian} & \Tobj & {pedestrian} \\
    \text{bicycle} & \Tobj & {bicycle} \\
    \text{bus} & \Tobj & {bus} \\
    
    \text{car}  & \Tobj & {car} \\[-1mm]
                & \Tobj & {van}  \\
    
    \text{constr.\ vehicle} & \Tobj & {construction vehicle}  \\											
    \text{motorcycle} & \Tobj & {motorcycle} \\
    \text{trailer}  & \Tobj & {trailer} \\
    \text{truck}    & \Tobj & {truck} \\[-1mm]
                    & \Tobj & {lorry with open cargo cab} \\[-1mm]
                    & \Tobj & {lorry with closed cargo cab} \\[-1mm]
                    & \Tobj & {lorry with open high cargo cab} \\
    
    \text{barrier}  & \Tobj & {concrete barrier}  \\
    \text{traffic cone} & \Tobj & {traffic cone} \\
    \text{driveable surface} & \Tstuff & {road}\\ 															
    \text{other flat} & \Tobj & {traffic island}   \\
    \text{sidewalk} & \Tstuff & {sidewalk without objects on it} \\
    
    \text{terrain}  & \Tstuffdrive & {green terrain} \\[-1mm]
                    & \Tstuffdrive & {less green and soil terrain} \\[-1mm]
                    & \Tstuffdrive & {soil terrain}  \\

    \text{manmade}  & \Tstuff    & {wall} \\[-1mm]
                    & \Tobjdrive & {concrete stairs} \\[-1mm]
                    & \Tobjdrive & {traffic light} \\[-1mm]
                    & \Tobjdrive & {traffic sign} \\[-1mm]
                    & \Tobjdrive & {pole} \\[-1mm]
                    & \Tobjdrive & {fire hydrant} \\[-1mm]
                    & \Tobjdrive & {2-3 skyscrapers close to each other} \\[-1mm]
                    & \Tobjdrive & {house} \\[-1mm]
                    & \Tobjdrive & {apartments}  \\
    
    \text{vegetation} & \Tobjdrive & {bush} \\[-1mm]
                    & \Tobjdrive & {shrub} \\[-1mm]
                    & \Tobjdrive & {horizontal vegetation that includes shrub and bushes} \\[-1mm]
                    & \Tobjdrive & {woods} \\[-1mm]
                    & \Tobj & {tree trunk} \\
\bottomrule
\end{tabular}
\end{table*}
    
\begin{table*}[t]
    \centering
    \small
    \caption{\textbf{Our classwise text prompts used to generate prototype images for SemanticKITTI.} Templates \Tobj\ and \Tstuffdrive\ are defined in \cref{tab:text_prompts}. Only a few prompts are specific to SemanticKITTI; for other classes, the prompts are the same as in nuScenes for similar classes. In practice, we do not generate new images; we just reuse the same prototype images already generated for nuScenes.}
    \setlength\extrarowheight{4pt}
    \begin{tabular}{l|l|l}
    \toprule
    Class & Templ. & Text prompts \\
    \midrule

    \text{car} & & \notextprompt\ \textrm{car} \\
    \text{bicycle} & & \notextprompt\ \textrm{bicycle} \\
    \text{motorcycle} & & \notextprompt\ \textrm{motorcycle}  \\
    \text{truck} & & \notextprompt\ \textrm{truck}  \\
    \text{other vehicle} & & \notextprompt\ \textrm{bus}, \textrm{trailer}, \textrm{construction vehicle} \\
    
    \text{person} & &\notextprompt\ \textrm{pedestrian} \\
    \text{bicyclist} & \Tobj & {bicyclist sitting on a bicycle} \\
    \text{motorcyclist} & \Tobj & {motorcyclist sitting on a motorcycle} \\
    \text{road} & & \notextprompt\ \textrm{driveable surface}\\
    
    \text{parking} & \Tstuffdrive & {parking}    \\
    \text{sidewalk} & & \notextprompt\ \textrm{sidewalk} \\											
    \text{other ground} & & \notextprompt\ \textrm{other flat} \\																		
    \text{building} & & \notextprompt\ \text{skyscrapers}, \textrm{house}, \textrm{apartment} \\ %\textsl{in} \rlap{\textrm{manmade}}  \\    
    \text{fence} &  \Tobj & {fence}    \\    
    \text{vegetation} & & \notextprompt\ \textrm{vegetation} \notp{except} \textrm{tree trunk} \\
    \text{trunk} &  & \notextprompt\ \textrm{tree trunk} \\ % \textsl{in} \textrm{vegetation} \\				
    \text{terrain} & & \notextprompt\ \textrm{terrain}  \\
    \text{pole} & & \notextprompt\ \textrm{pole} \\ % in \textrm{manmade} \\
    \text{traffic sign} & & \notextprompt\ \textrm{traffic sign} % \\  \textsl{in} \textrm{manmade}
    \raisebox{-5pt}{}\\
    \bottomrule
    \end{tabular}
    \label{tab:text_prompts_sk} 
\end{table*}

\cref{tab:text_prompts_ns,tab:text_prompts_sk} lists the text prompts used to generate prototype images for \ns and \sk.

As explained in \cref{sec:protogen}, we use a (single) context prompt per class, depending on whether it is an object or a stuff class, using respectively templates \Tobj\ and \Tstuff\ defined in \cref{tab:text_prompts}. For some classes, the context prompt makes explicit the driving environment, using \Tobjdrive\ or \Tstuffdrive\ (see \cref{tab:text_prompts}), which actually could have been done for all classes but was useful only for a fraction of them. This single context prompt is to be compared to the possibly many context prompts (up to 85 templates) used in other approaches (see \cref{tab:rwprompts}).

\begin{table*}[t!]
\centering\tabcolsep 3pt
\caption{\textbf{Number of prompt templates and subclass prompts used by 3D OVSS methods for nuScenes.} \ul{\emph{Code/checkpoint available:}} \codeavail~= yes for both nuScenes (NS) and SemanticKITTI (SK); \codeavailnoSK~= for NS, not for SK; \codeavailissues~= partial or with long-standing open issues; \nocodeavail~= no; $^\dagger$: upon publication. \ul{\emph{No templ./subcl.:}} number of context prompt templates per subclass name. \ul{\emph{No subcl.:}} number of subclass names used to make the 16 classes of nuScenes explicit. }
\resizebox{0.99\linewidth}{!}{%
\def\sqc{~~~}
\def\spc{~~}
\begin{tabular}{l@{~}r|c|r|r|l}  
\toprule
        && Code/ & \multicolumn{1}{c|}{No.} & \\
\multicolumn{2}{l|}{3D OVSS}  
        & chkpt & \multicolumn{1}{c|}{templ.} & \multicolumn{1}{c|}{No.} & \\
\multicolumn{2}{l|}{method}  
        & avail. & \multicolumn{1}{c|}{\,/subcl.} & \multicolumn{1}{c|}{\,subcl.} & Remark\\
\midrule
CNS & \cite{chen2023cns}    & \nocodeavail & 85 \sqc & - \spc & \\
HICL & \cite{kang2024hicl}  & \nocodeavail & 85 \sqc & - \spc & \\
AdaCo & \cite{zou2025adaco} & \nocodeavail &  - \sqc & 55 \spc \\
AFOV & \cite{sun2025afov}   & \codeavailnoSK & 14 \sqc & 175 \spc \\
LeAP & \cite{gebraad2025leap}& \nocodeavail &  - \sqc &  - \spc & car $\Rightarrow$ automobile, van, etc.\\
LOSC & \cite{samet2026losc} & \nocodeavail & 81 \sqc & 44 \spc & Templates are via OpenSeeD \cite{openseed}\\
\midrule
\rowcolor{black!20}
CLIP2Scene & \cite{clip2scene}  & \codeavailissuesnoSK & 85 \sqc & 16 \spc & 16 raw class names, but poor results \\
GGSD & \cite{wang2024ggsd}      & \codeavailissuesnoSK &  \textbf{1} \sqc &  - \spc & Template is ``a $C$ in a scene''\\ 
SAS & \cite{li2025sas}          & \codeavailissuesnoSK &  \textbf{1} \sqc & 43 \spc & Template is ``a $C$ in a scene''\\ 
\midrule
OV3D & \cite{jiang2024ov3d}     & \nocodeavail         &  81 \sqc & - \spc & Templates are via SEEM \cite{zou2023seem} \\
\multicolumn{2}{l|}{OV3D\,w/\,OpenScene} & \nocodeavail & 82 \sqc & 43 \spc & Template is ``a $C$ in a scene'' \\
3D-AVS & \cite{wei20253davs}    & \codeavailissuesnoSK &  \textbf{1} \sqc & 43 \spc & Template is ``a $C$ in a scene''\\
\midrule
OpenScene & \cite{peng2023openscene} & \codeavailnoSK  &  \textbf{1} \sqc & 43 \spc & Template is ``a $C$ in a scene''\\
SAL & \cite{osep2024sal}             & \nocodeavail    & 80 \sqc & 47 \spc & \\
\rowcolor{orange!18}
%\cellcolor{orange!18} 
\ours & \cellcolor{orange!18} (ours)  & \codeavail\rlap{$^\dagger$}          &  \textbf{1} \sqc & \textbf{34} \spc & See \cref{tab:text_prompts_ns}\\
\bottomrule
\end{tabular}
}
\label{tab:rwprompts}
\end{table*}

\begin{table}[!t]
\centering
\caption{\textbf{Increasing the number of generated ChatGPT images per prompt.} Performance (on NS val set) plateaus at two.}
\label{tab:number_of_templates}
\begin{tabular}{cc } 
\toprule
    Prototype Number & (LR) mIoU\%
\\
\midrule
1 $\times$ ChatGPT &  46.6 \\
2 $\times$ ChatGPT &  47.3 \\
3 $\times$ ChatGPT & 47.2 \\
4 $\times$ ChatGPT & 46.9  \\
5 $\times$ ChatGPT & 46.5  \\
8 $\times$ ChatGPT & 47.4  \\
\bottomrule
\end{tabular}
\end{table}
\begin{table}[!t]
\centering
\caption{\textbf{Randomly selecting two prototypes for each class/prompt}. The variance is very small between the selected random sets which shows the robustness of our method against generated templates.}
\label{tab:variance_of_template_pairs}
\begin{tabular}{cc } 
\toprule
    Random pair id & (LR) mIoU\%
\\
\midrule
v1 &  47.3 \\
v2 &  46.8 \\
v3 & 47.5 \\
v4 & 47.0  \\
v5 & 47.0  \\ 
\midrule
Variance & 0.08 \\ 
Standard Deviation & 0.28 \\ 
\bottomrule
\end{tabular}
\end{table}

As also explained in \cref{sec:protogen}, for ``composite'' classes such as \textit{manmade} or \textit{vegetation} (being different from \textit{terrain}), as well as for ``negative'' classes such as \textit{other flat} or \textit{other ground}, we have to introduce a few explicit classes, as described in the annotator instructions \cite{nuscenesannotinstruct}. To make the 16 classes of nuScenes explicit enough, we use 34 subclass names (see \cref{tab:text_prompts_ns}). In comparison, apart from CLIP2Scene that uses the 16 raw class names (and thus has a poor performance), all other methods use at least 43 subclasses, and up to 175 (see \cref{tab:rwprompts}). 

Note that we do not generate different images (nor use different text prompts) for classes that are identical in both \ns and \sk, such as \textit{car} or \textit{truck}. In fact, when segmenting SemanticKITTI, we reuse as much as possible the images generated for nuScenes; only new classes require extra prompts and images (see \cref{tab:text_prompts_sk}).

\section{Impact of the number of prototypes}
\label{sec:number_of_protos}

In \Cref{tab:number_of_templates}, we use an increasing number of randomly generated ChatGPT images per prompt and report performances on \ns val set. We observe that performance plateaus at two. 

\Cref{tab:variance_of_template_pairs} shows the robustness of our method to different % random 
generations. We randomly select two prototypes for each class/prompt and show the variance is very small between these random sets.

\section{Qualitative results}
\label{sec:qualitative}

\begin{figure*}
\centering
\begin{tabular}{ccc}
& Ground truth & \ours \\
%
% nuScenes
%
\toprule
    \rotatebox{90}{\hspace{1.5cm}\ns}
&    \includegraphics[width=0.43\linewidth,keepaspectratio] {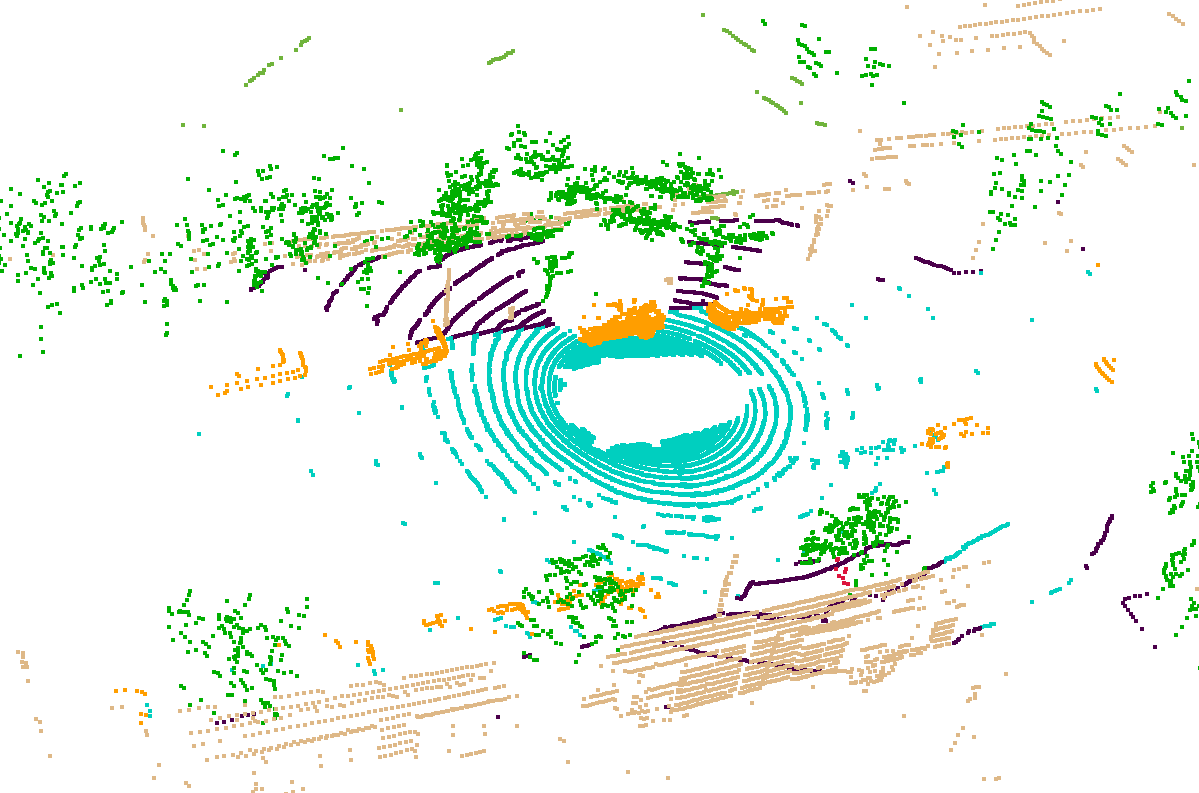}
&    \includegraphics[width=0.43\linewidth,keepaspectratio] {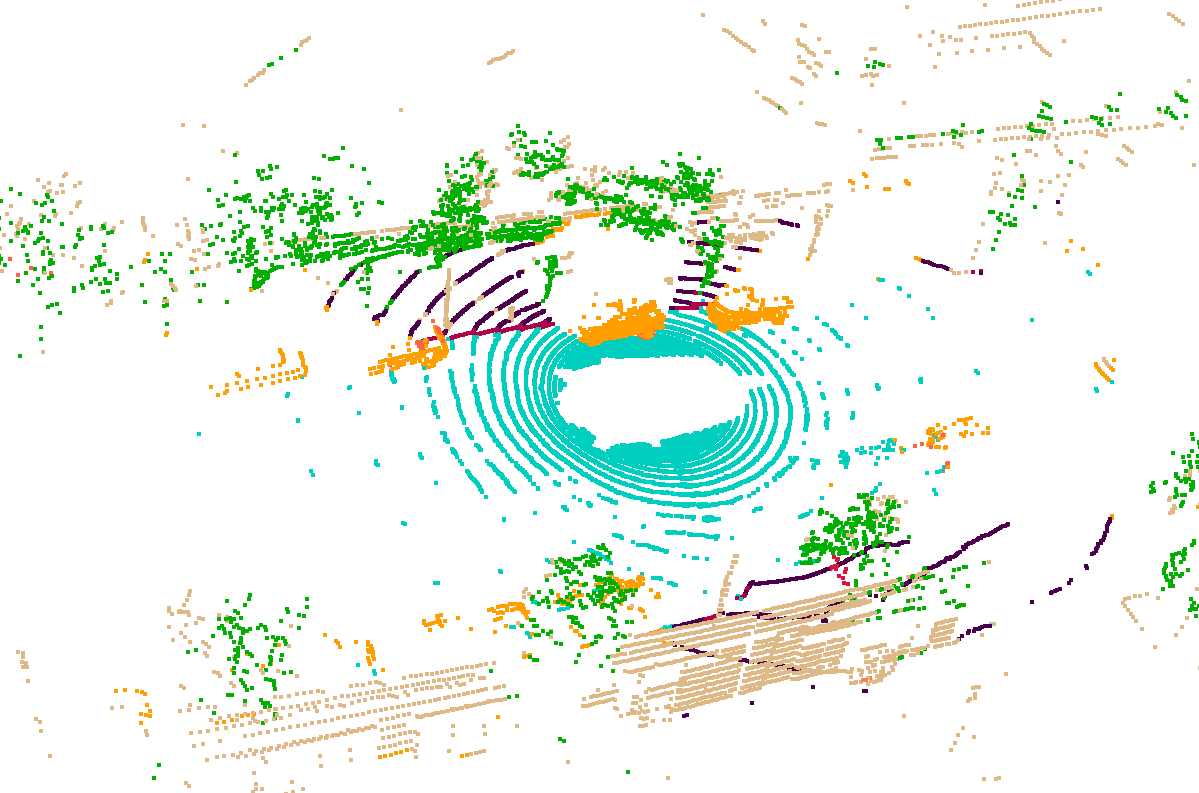}
\\ \midrule %\hline
    \rotatebox{90}{\hspace{1.5cm}\ns}
&    \includegraphics[width=0.43\linewidth,keepaspectratio] {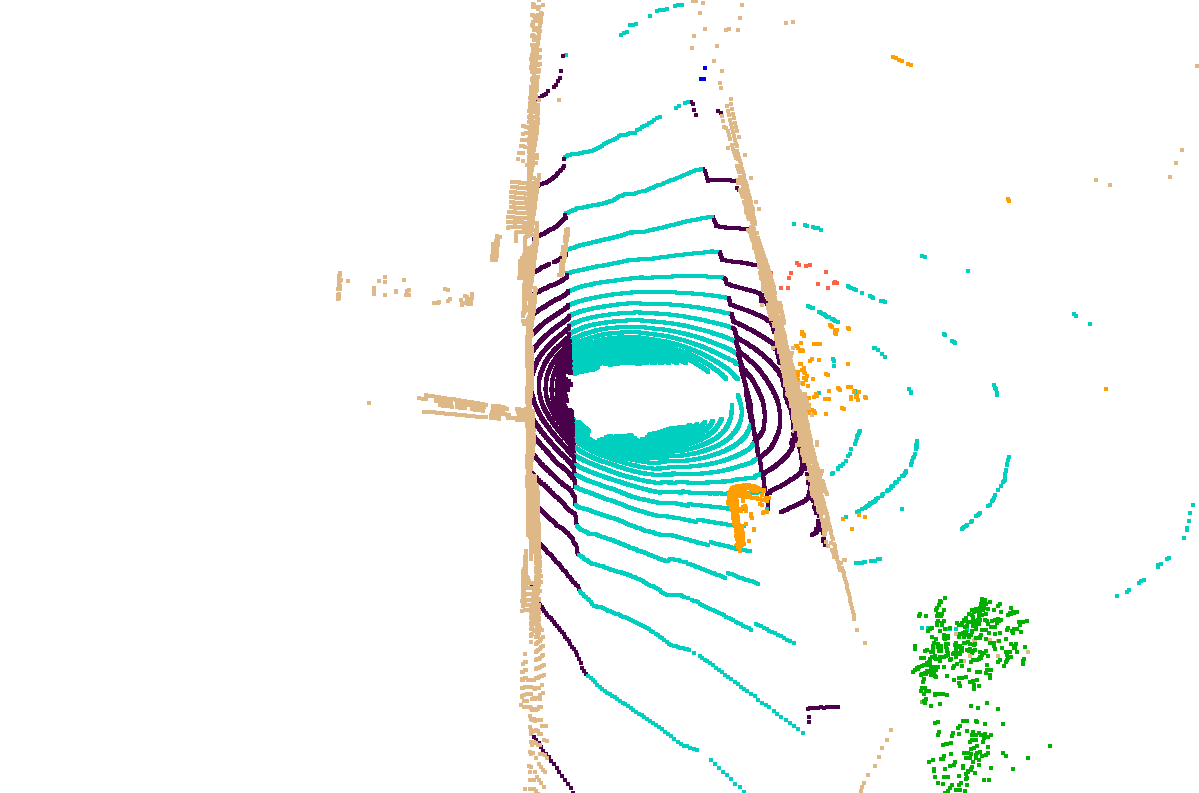} 
&    \includegraphics[width=0.43\linewidth,keepaspectratio] {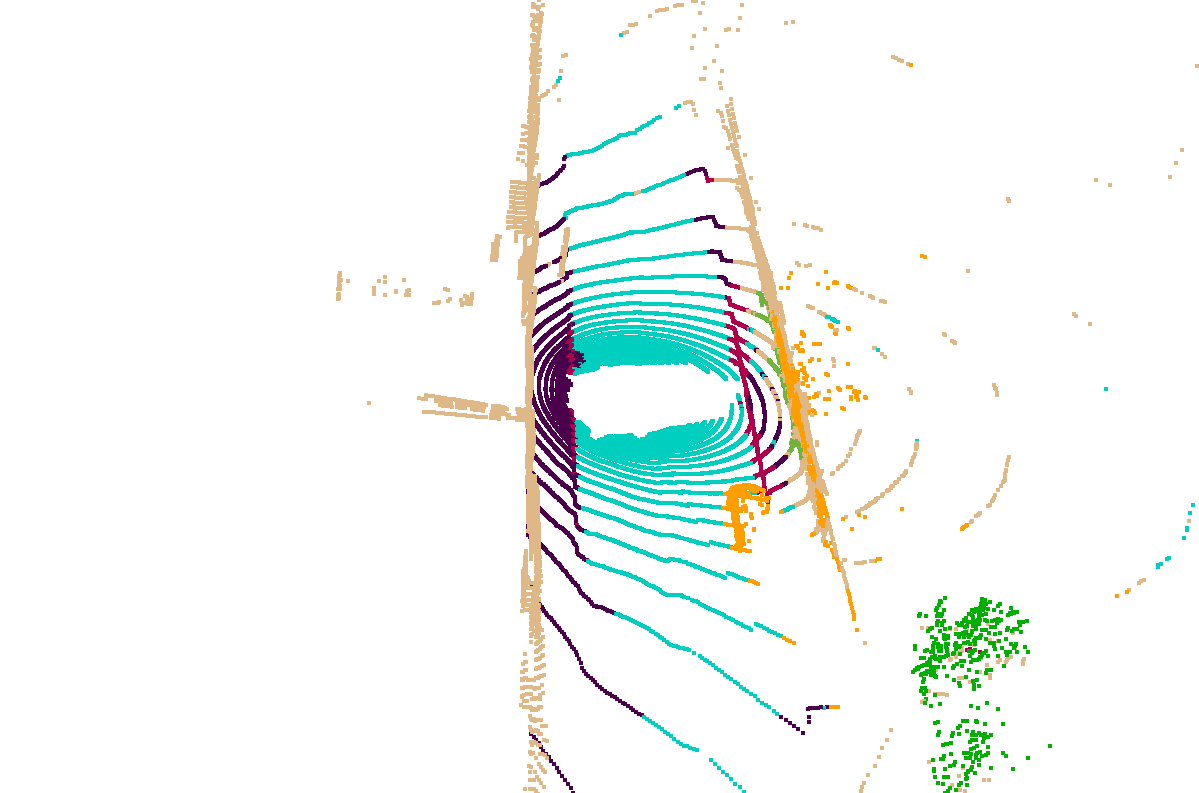}
\\
\toprule
%
% SemanticKITTI
%
    \rotatebox{90}{\hspace{1cm}\sk}
&  \includegraphics[width=0.43\linewidth,keepaspectratio] {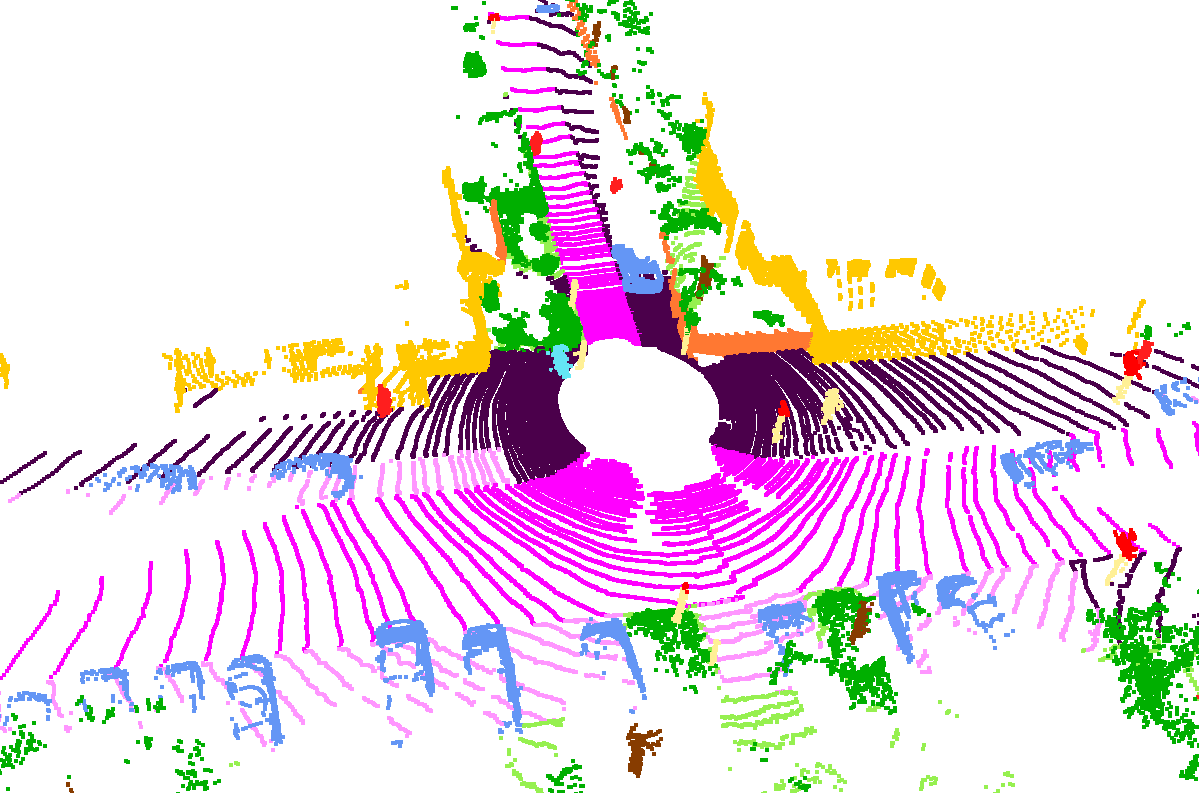}
&
\includegraphics[width=0.43\linewidth,keepaspectratio] {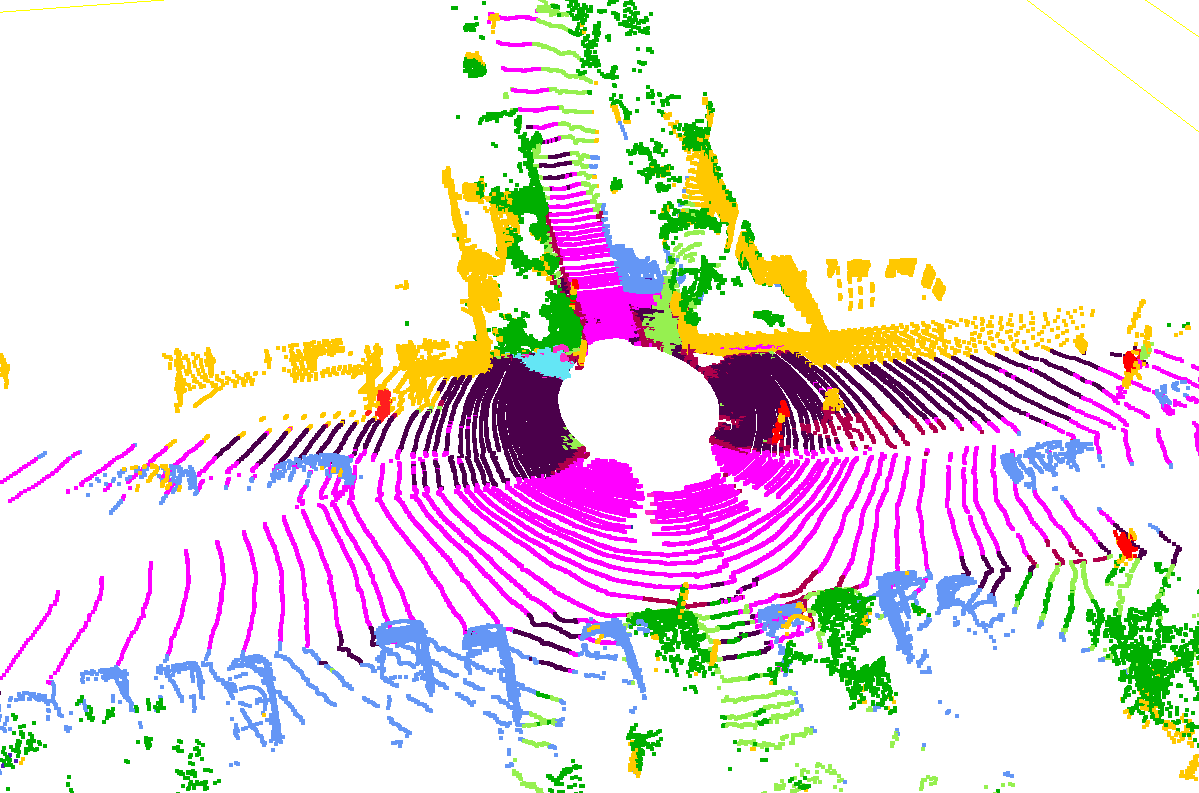}
\\ \midrule %\hline
    \rotatebox{90}{\hspace{1cm}\sk}
&    \includegraphics[width=0.43\linewidth,keepaspectratio] {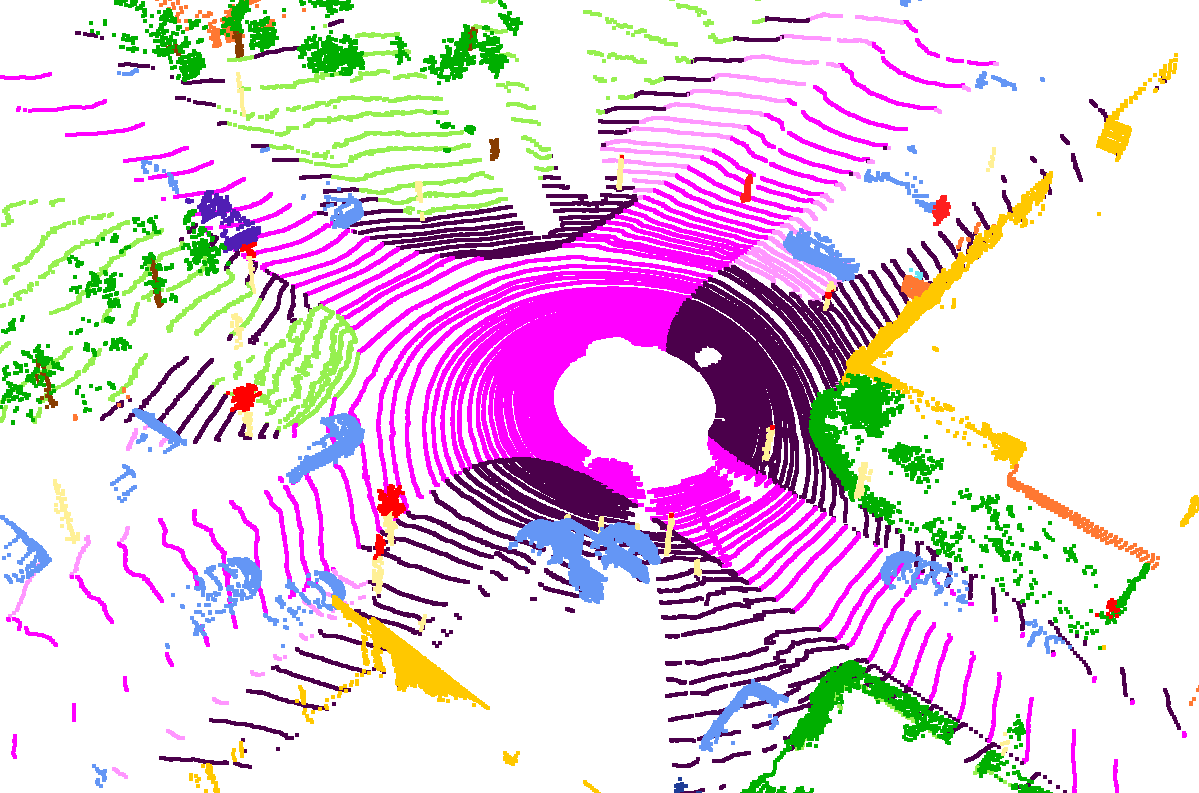}
&    \includegraphics[width=0.43\linewidth,keepaspectratio] {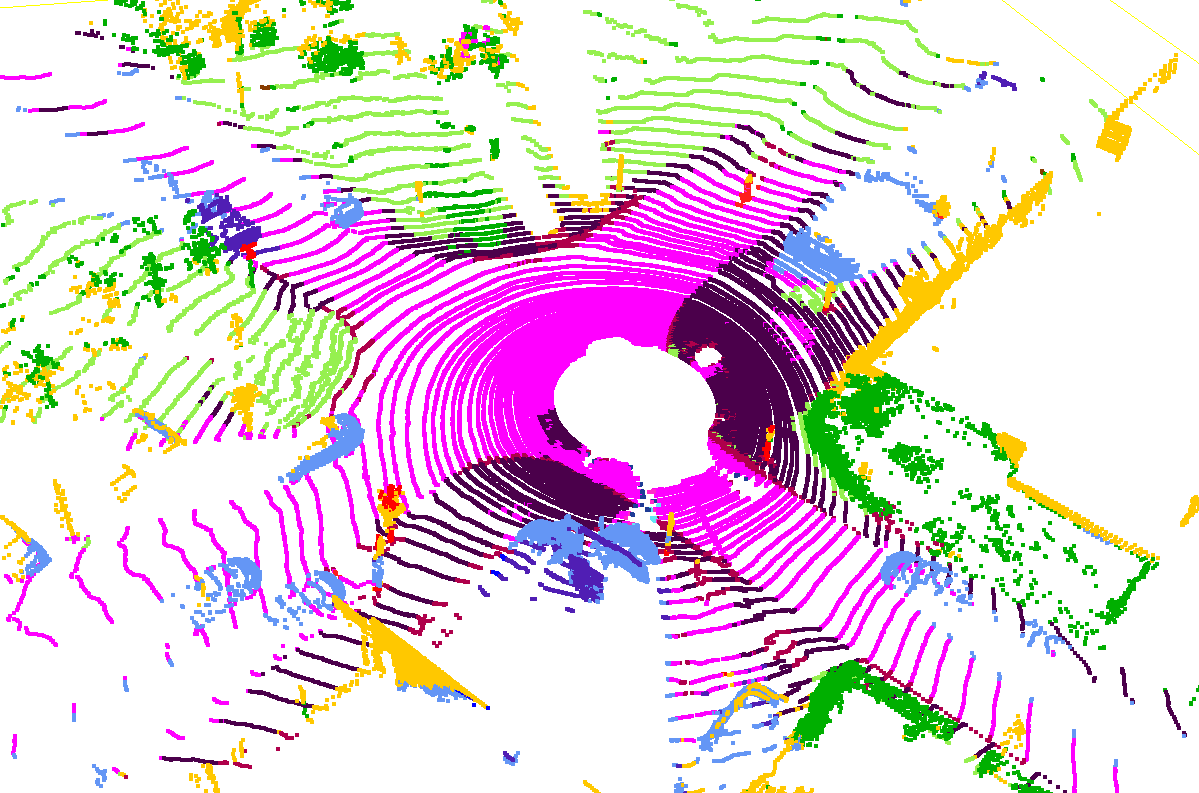}
\\
\bottomrule
\end{tabular}
\caption{Qualitative results of semantic segmentation from the validation sets of \ns and \sk. The color code used to represent each class is provided in \cref{tab:colorcodes}.}
\label{fig:qual}
\end{figure*}

In \cref{fig:qual}, we present qualitative results on the \ns and \sk validation sets. 

\textbf{Failure cases.} A common error of \ours on both datasets is confusion between flat surface classes, such as \emph{road/driveable surface}, \emph{sidewalk}, and \emph{terrain}. In \sk, \emph{trunk} points are also often classified as \emph{vegetation} instead of being separated as an individual class.

\begin{figure*}[t]
\centering
\begin{tabular}{ccc}
OpenScene \cite{peng2023openscene} & \ours & Ground truth \\
%
% nuScenes
%
\toprule
    % \rotatebox{90}{\hspace{1.5cm}\ns}
\includegraphics[width=0.3\textwidth,keepaspectratio] {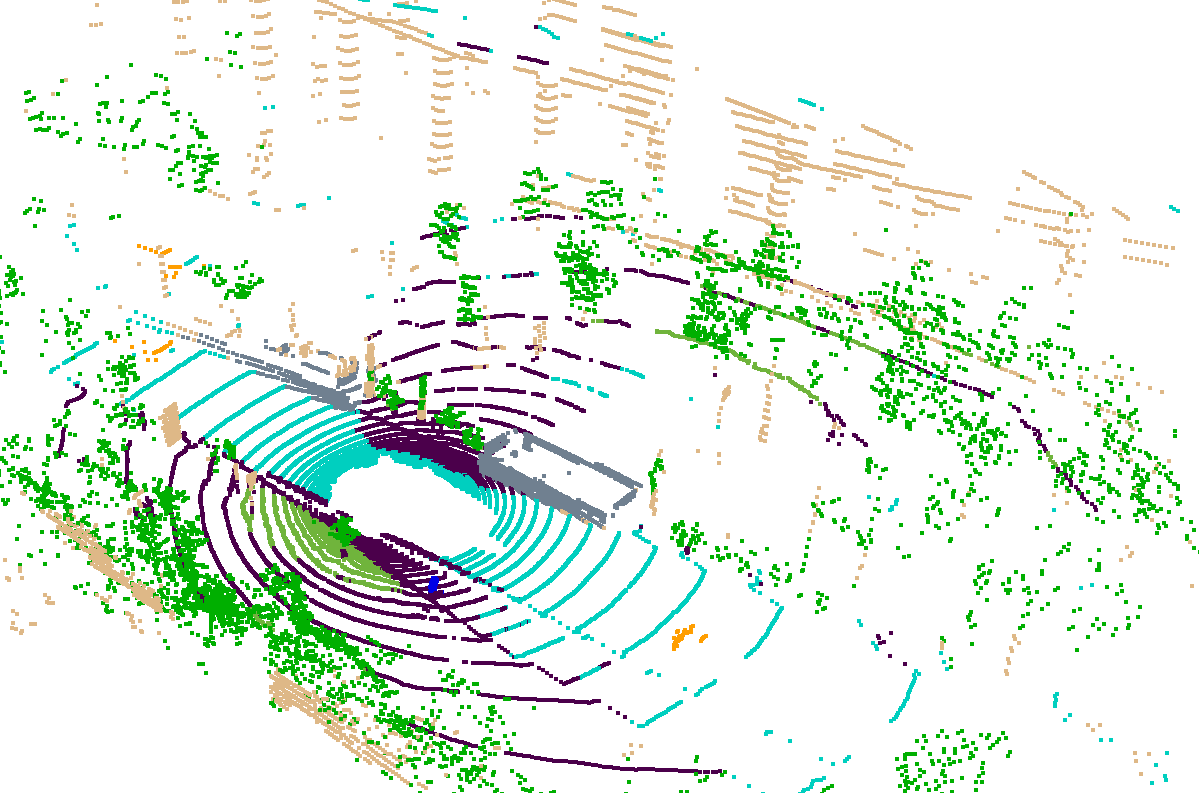}
&    \includegraphics[width=0.3\textwidth,keepaspectratio] {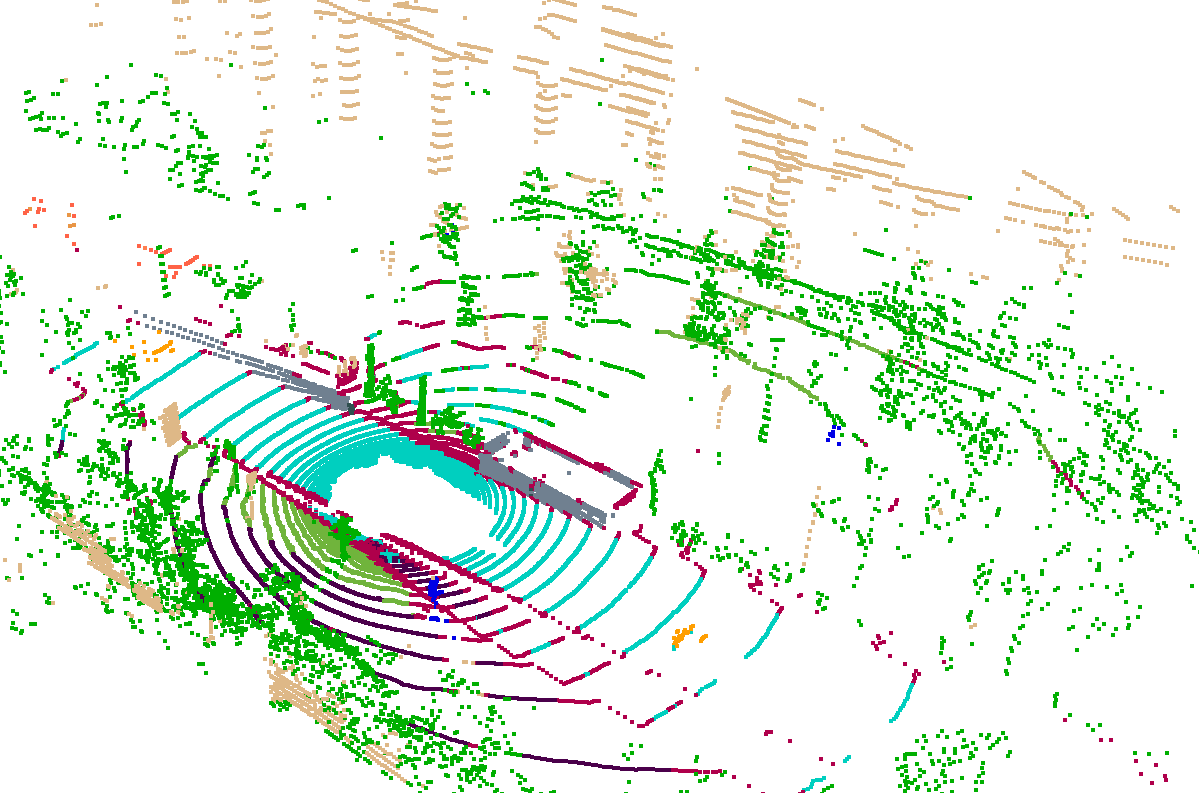}
&    \includegraphics[width=0.3\textwidth,keepaspectratio] {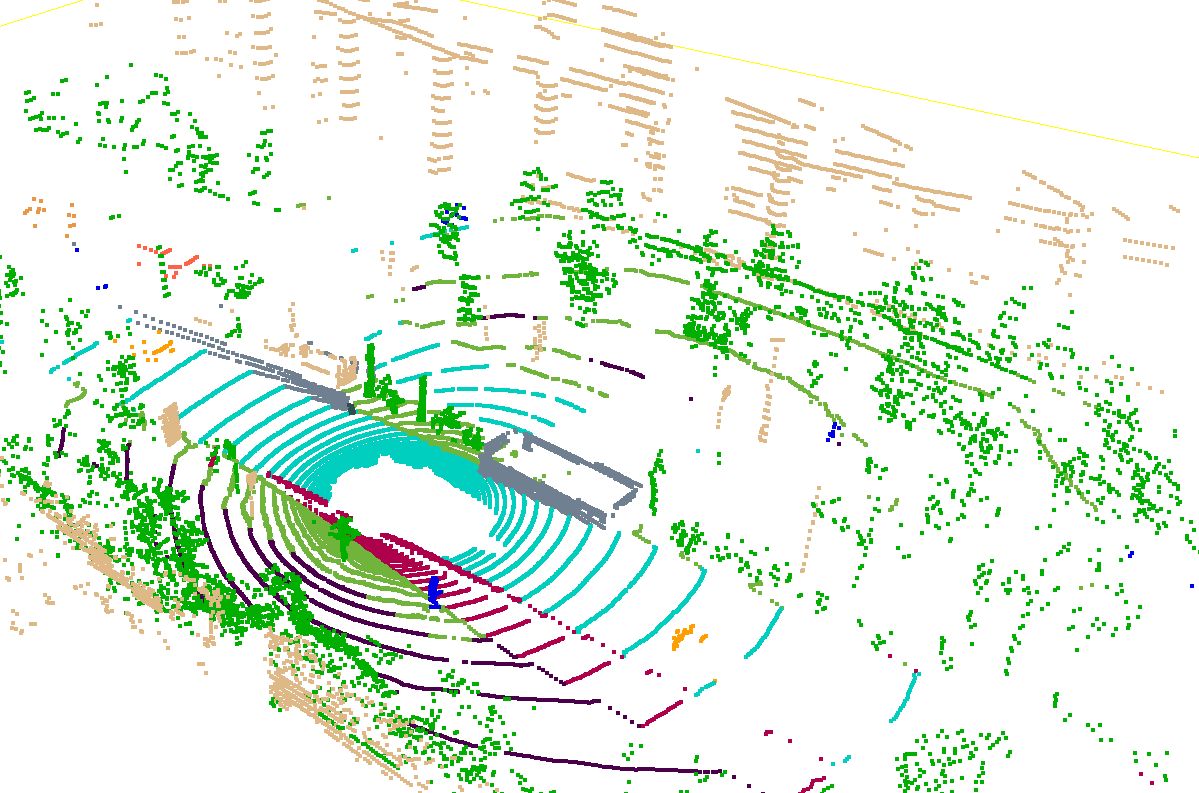} \\
\midrule % \hline
\includegraphics[width=0.3\textwidth,keepaspectratio] {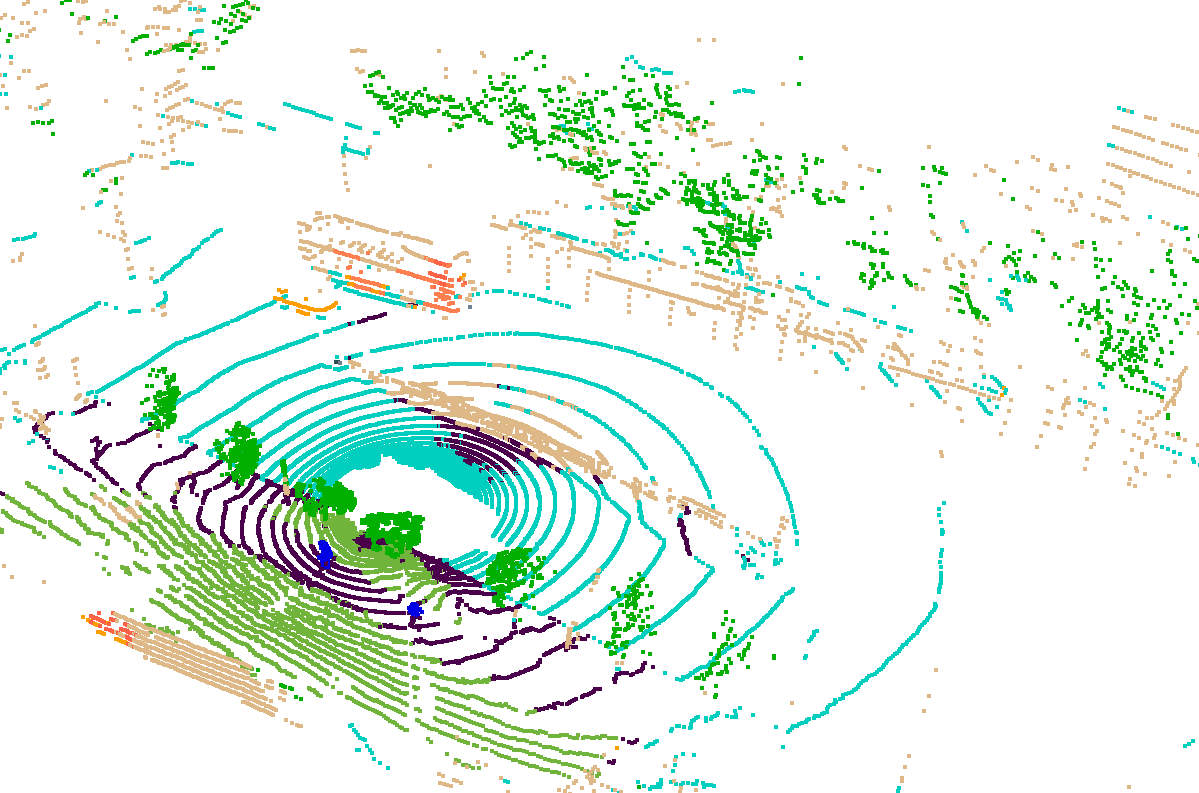}
&    \includegraphics[width=0.3\textwidth,keepaspectratio] {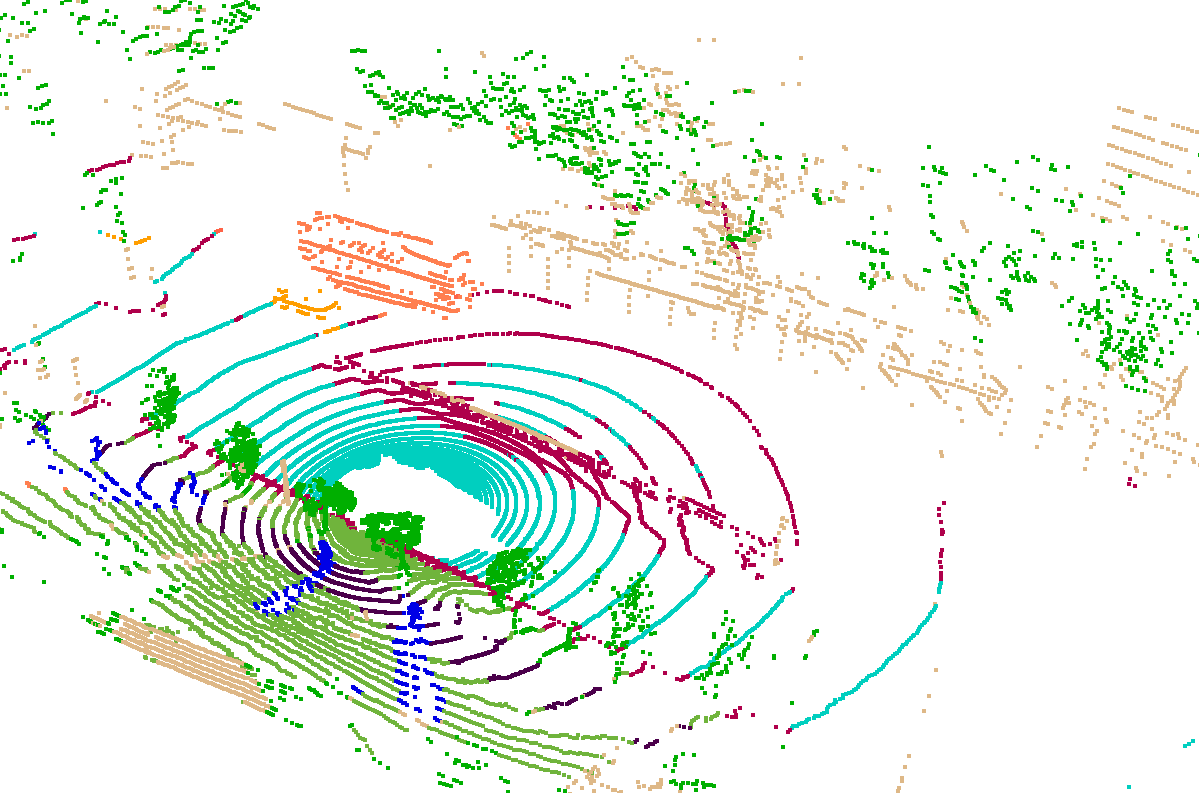}
&    \includegraphics[width=0.3\textwidth,keepaspectratio] {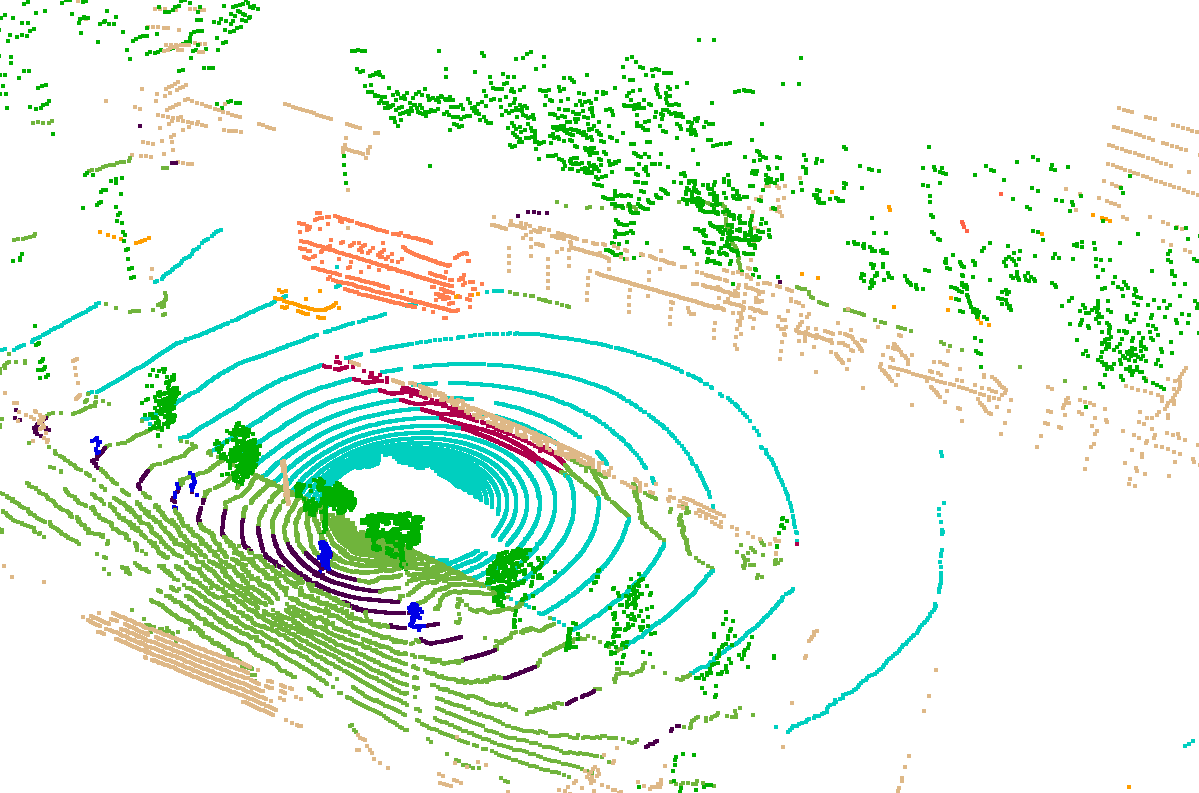} \\
\midrule % \hline
\includegraphics[width=0.3\textwidth,keepaspectratio] {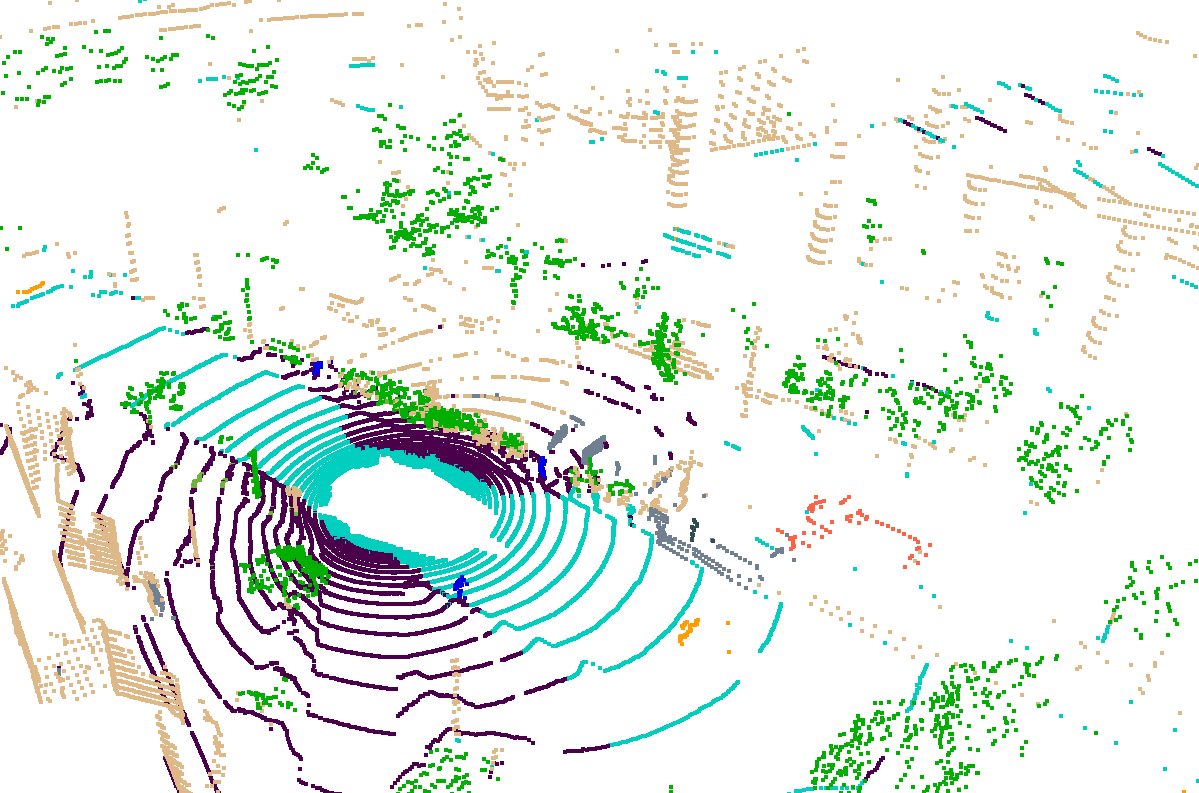}
&    \includegraphics[width=0.3\textwidth,keepaspectratio] {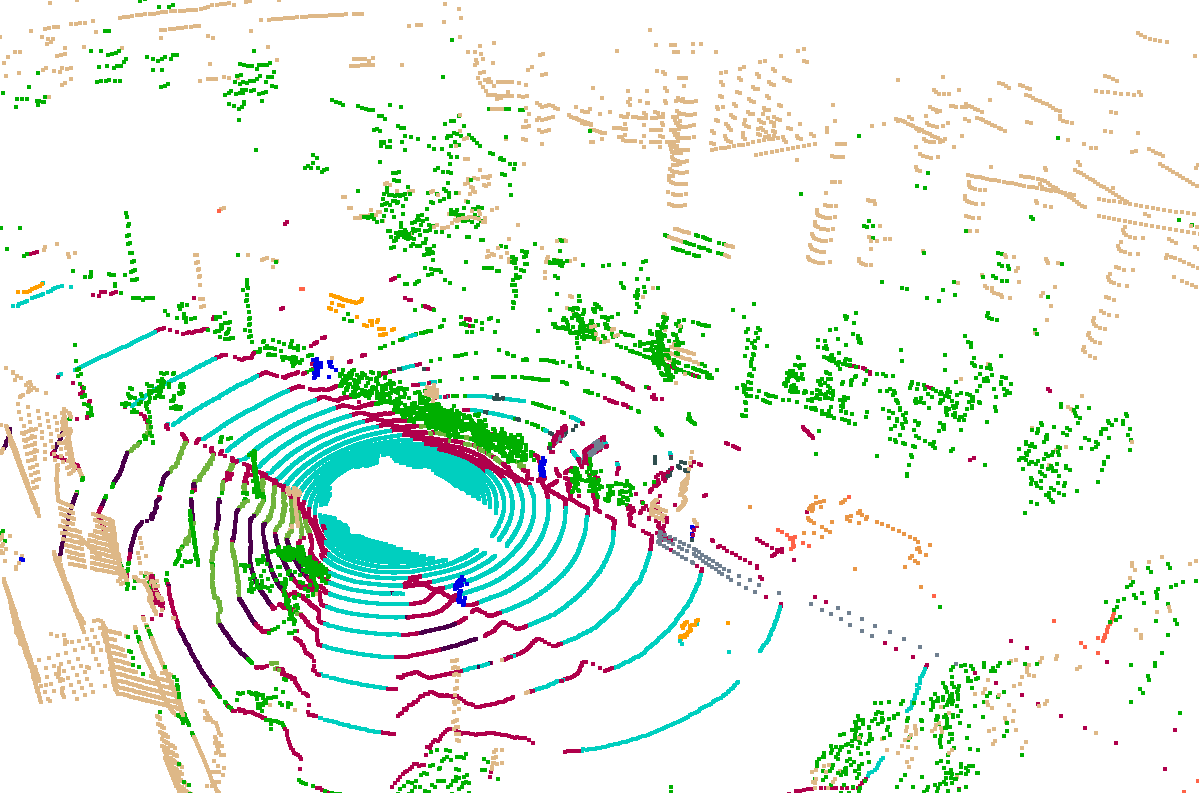}
&    \includegraphics[width=0.3\textwidth,keepaspectratio] {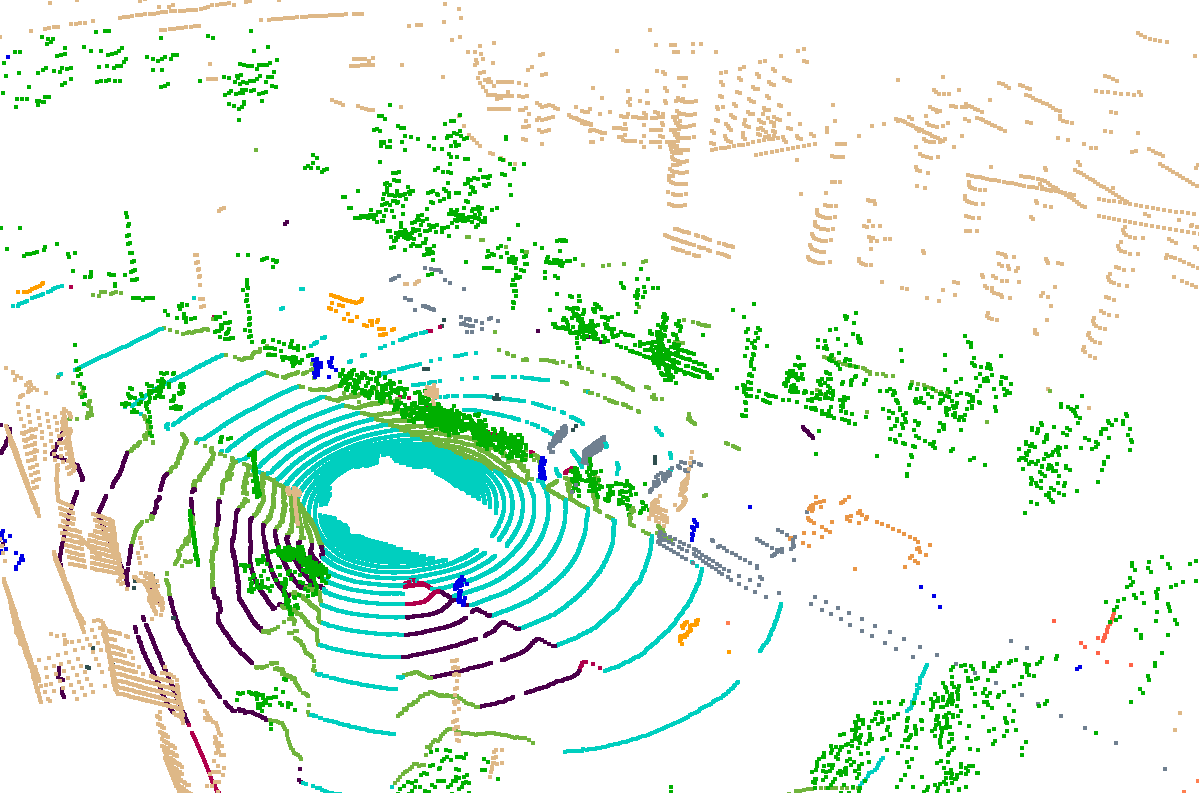} \\
\bottomrule
\end{tabular}
\caption{Qualitative comparison of semantic segmentation on the \ns validation set using our method and OpenScene \cite{peng2023openscene}. The color code used to represent each class is provided in \cref{tab:colorcodes}. Our method produces more faithful results by detecting other flat in the first row, segmenting trailer in the second row, and segmenting road and pedestrians in the third row.}
\label{fig:openscene_comparison}
\end{figure*}

We also include a qualitative comparison with OpenScene \cite{peng2023openscene} in \cref{fig:openscene_comparison}, as it is open-sourced.

\begin{table}[ht]
\newcommand\colrect[1]{\textcolor[RGB]{#1}{\rule{5ex}{1.3ex}}}
\caption{\textbf{color codes} used to represent each class of nuScenes or SemanticKITTI in \cref{fig:qual} and \cref{fig:openscene_comparison}.}
\label{tab:colorcodes}
%\vspace*{-3ex}
\centering
\resizebox{\linewidth}{!}{%
\begin{tabular}[t]{c@{~~}lc@{~~}l}
    % \multicolumn{4}{c}{\textbf{nuScenes}} \\
         Color & \rlap{\textbf{nuScenes class}} \\
\midrule
         \colrect{112,128,144} & barrier            & \colrect{255,140,000} & trailer            \\
         \colrect{220,020,060} & bicycle            & \colrect{255,099,071} & truck              \\
         \colrect{255,127,080} & bus                & \colrect{000,207,191} & driveable surface  \\
         \colrect{255,158,000} & car                & \colrect{222,184,135} & manmade            \\
         \colrect{233,150,070} & construction vehicle   & \colrect{175,000,075} & other flat         \\
         \colrect{255,061,099} & motorcycle         & \colrect{075,000,075} & sidewalk           \\
         \colrect{000,000,230} & pedestrian         & \colrect{112,180,060} & terrain            \\
         \colrect{047,079,079} & traffic cone       & \colrect{000,175,000} & vegetation         \\
    \end{tabular}
    ~~
    \begin{tabular}[t]{c@{~~}lc@{~~}l}
    %\rlap{\hspace*{-1ex}\textbf{SemanticKITTI}} \\
         Color &  \rlap{\textbf{SemanticKITTI class}} \\
\midrule
         \colrect{100, 230, 245} & bicycle      & \colrect{175, 0, 75} & other ground    \\
         \colrect{255, 40, 200} & bicyclist     & \colrect{75, 0, 75} & sidewalk         \\
         \colrect{10, 150, 245} & car           & \colrect{255, 200, 0} & building       \\
         \colrect{0, 0, 255} & other vehicle    & \colrect{255, 120, 50} & fence         \\
         \colrect{30, 60, 150} & motorcycle     & \colrect{255, 240, 150} & pole         \\
         \colrect{150, 30, 90} & motorcyclist   & \colrect{255, 0, 0} & traffic sign     \\
         \colrect{255, 30, 30} & person         & \colrect{135, 60, 0} & trunk           \\
         \colrect{80, 30, 180} & truck          & \colrect{150, 240, 80} & terrain       \\
         \colrect{255, 0, 255} & road           & \colrect{0, 175, 0} & vegetation       \\
         \colrect{255, 150, 255} & parking      \\ 
    \end{tabular}
    }
\end{table}

\section{Compute report and licenses}
\label{sec:licenses}

\textbf{Compute resources}. We train ScaLR+ on 8 NVIDIA A100 GPUs with 40 GB of memory each, and 64 GB of CPU RAM. We run our OVSS experiments on a single A100 GPU.

\subsection{Licenses} 
\label{app:licenses}

All datasets and code used in this work are utilized in accordance with their respective licenses and terms of use. No data is used in violation of applicable licensing restrictions.

\textbf{SemanticKITTI license.} The SemanticKITTI dataset is released under the Creative Commons Attribution–NonCommercial–ShareAlike 4.0 International License (CC BY-NC-SA 4.0). This license permits use, distribution, and modification of the data for non-commercial purposes, provided that appropriate credit is given and any derivative works are distributed under the same license. Commercial use is not permitted without additional authorization from the dataset providers.

\textbf{nuScenes license.}
The nuScenes dataset is distributed under the nuScenes Terms of Use provided by Motional. The dataset is available for research and non-commercial use subject to acceptance of these terms. Redistribution and commercial use are restricted and require a separate agreement with the dataset providers. Users must comply with all conditions specified in the official nuScenes license.

\textbf{ScaLR repository license.}
The ScaLR codebase is released under the Apache License 2.0. This is a permissive open-source license that allows use, modification, and distribution of the code for both research and commercial purposes, provided that the original copyright notice and license terms are preserved.

\textbf{WaffleIron repository license.}
The WaffleIron codebase is released under the Apache License 2.0. This permissive open-source license allows use, modification, and distribution of the code for both research and commercial purposes, provided that the original copyright notice and license terms are preserved. 

\textbf{OpenScene repository license.}
The OpenScene codebase is released under the Apache License 2.0. This permissive open-source license allows use, modification, and distribution of the code for both research and commercial purposes, provided that the original copyright notice and license terms are preserved.

\end{document}